\documentclass{aicom2e}
\usepackage[active]{srcltx}
\usepackage{ amsthm, amssymb,amsfonts}
\usepackage{graphicx}
\usepackage{epsfig}
\usepackage{epstopdf}
\usepackage{latexsym}
\usepackage{colortbl}
\usepackage{times}
\usepackage{helvet}
\usepackage{courier}
\usepackage{caption}
\usepackage{float}
\usepackage{xcolor}
\usepackage{pslatex}
\usepackage{eso-pic}
\usepackage[bottom]{footmisc}
\usepackage{url}
\usepackage{etex}
\usepackage{algorithm}
\usepackage{algorithmic}
\usepackage{lscape}
\usepackage{rotating}
\usepackage{subcaption}
\usepackage{color, colortbl}

\begin{document}

\begin{frontmatter}                           
%
\title{Learning in Unlabeled Networks ---\\ An Active Learning and Inference Approach}
\maketitle
\author[]{\fnms{Tomasz} \snm{Kajdanowicz\textsuperscript{1}}, \fnms{Rados{\l}aw} \snm{Michalski\textsuperscript{1},}
}
\author[]{\fnms{Katarzyna} \snm{Musial{\textsuperscript{2}}}, \fnms{Przemys{\l}aw} \snm{Kazienko\textsuperscript{1}}
}
\address{\textsuperscript{1}Wroc{\l}aw University of Technology\\
Department of Computational Intelligence\\
Wroc{\l}aw, Poland\\
E-mail: tomasz.kajdanowicz@pwr.edu.pl,\\
radoslaw.michalski@pwr.edu.pl,\\
kazienko@pwr.edu.pl\\
-\\
\textsuperscript{2}Bournemouth University\\
Faculty of Science and Technology\\
Bournemouth, UK\\
E-mail: kmusialgabrys@bournemouth.ac.uk}

\begin{abstract}
The task of   determining  labels of all network nodes based on the knowledge about network structure and labels of some training subset of nodes is called the within--network classification. It may happen that none of the labels of the nodes is known and additionally there is no information about number of classes (types of labels)  to which nodes can be assigned. In such a case a subset of nodes has to be selected for initial label acquisition. The question that arises is: "labels of which nodes should be collected and used for learning in order to provide the best classification accuracy for the whole network?".
Active learning and inference is a practical framework to study this problem.

In this paper, set of methods for active learning and inference for within network classification is proposed and validated. The utility score calculation for each node based on network structure is the first step in the entire process. The scores enable to rank the nodes.  Based on the created ranking, a set of nodes, for which the labels are acquired, is selected (e.g. by taking top or bottom $N$ from the ranking). The new measure--neighbour methods proposed in the paper suggest not obtaining labels of nodes from the ranking but rather acquiring labels of their neighbours. The paper examines 29 distinct formulations of utility score and selection methods reporting their impact on the results of two collective classification algorithms: Iterative Classification Algorithm (ICA) and Loopy Belief Propagation (LBP). 

We advocate that the accuracy of presented methods depends on the structural properties of the examined network. We claim that measure--neighbour methods will work better than the regular methods for networks with higher clustering coefficient and worse than regular methods for networks with low clustering coefficient. According to our hypothesis, based on clustering coefficient of a network we are able to recommend appropriate active learning and inference method.

Experimental studies were carried out on six real--world networks. In order to investigate our hypothesis, all analysed networks were categorized based on their structural characteristics into three groups. In addition, the  representativeness of initial set of nodes for which the labels are obtained and its influence on classification accuracy was examined. 
\end{abstract}

\end{frontmatter}

\section{Introduction}

In  many real--world networks, a set of nodes and connections between them are known but the information about their characteristics can be fragmentary and not coherent. On many occasions, however, the information about nodes'  labels is essential, e.g. knowing users' preferences or demographic profile is needed in the process of personalised recommendation of products or services. Of course, all labels can be obtained by asking everybody about them but, due to the scale of some networks and anonymity of many users in the online world, it may be a very time--consuming, costly and ineffective process. In order to reduce the resources required for manual acquisition of labels for all nodes, more sophisticated method, which enables to uncover labels of only limited number of nodes and based on this knowledge to perform the automatic classification for the rest of nodes, is needed. The research presented in this paper aims at addressing this issue by proposing an effective method for selection of starting nodes and acquisition of their labels that will serve as a training set for within network classification task. 

We also claim that there is no method that will work accurately in all cases and we show in our experiments that the accuracy of different methods strongly depend on some of the structural characteristics of a network.

For the illustration purposes we present an example of marketing campaign that will help to understand presented research. Let us assume that a~marketing campaign has to be addressed to a given community of customers. The knowledge about relationships between customers is available (e.g. derived from their monitored interactions), hence, we can create a customer social network. The main purpose of the campaign is to propose new product only to those who are likely to buy it within the next year. However, the top management allocates to the campaign only a fixed  amount of resources, which is not sufficient to target all community members. Thus, the question is which customers should be initially targeted in order to optimise the return on investment (ROI) of the entire campaign. ROI strongly depends on the quality of classification of community members into two classes: (i) customers and (ii) non--customers as we save resources by not sending the offer to non--customers.

One of the approaches to model the campaign is to use collective classification. 
In order to perform collective classification task, it is required to retrieve class labels for an initial population of nodes and next to use it in the inference process. Before classifying the whole network, some selected nodes need to be provided with the offer and their positive and negative responses together with the response rate should be collected. Afterwards, based on theirs behaviour as well as relationships between social network nodes, collective classification could model responses for the remaining nodes. The main issue is  to determine  which nodes should be selected to acquire their labels in order to maximise the performance of classification. An intuitive answer is that we should start with the nodes estimating the whole network most accurately. The solutions of the problem of which nodes' labels should be obtained in order to perform the collective classification are  called \textit {active learning} or \textit {active inference approaches} because they actively, not randomly, support selection of the learning set. 

The problem of selecting appropriate nodes in order to start collective classification process is studied in this paper. First, the literature review in the areas of (i) collective classification, as well as (ii) active learning and inference methods is described in Section \ref{sec:related}. Section \ref{sec:realtaionalActiveLearning} presents the method for selection of initial nodes for classification purposes proposed in this paper and in Section \ref{sec:Algorithms} there are revoked basic collective classification algorithms. The experiments using the proposed method and the datasets used in the process are described in Section \ref{sec:experiments} and discussed in Section \ref{sec:discussion}. Finally, Section \ref{sec:conclusions} summarises the main contribution of the paper. 


\section{Related work\label{sec:related}}
The research area that is in the focus of this paper is active learning and inference methods for within network classification and the literature that relates to this topic is presented below.  However, first the problem of collective classification is discussed to give a general background of the field and facilitate the understanding for non--expert readers.    

\subsection{Collective classification}

\begin{figure}
\centering
\includegraphics[width=0.55\columnwidth]{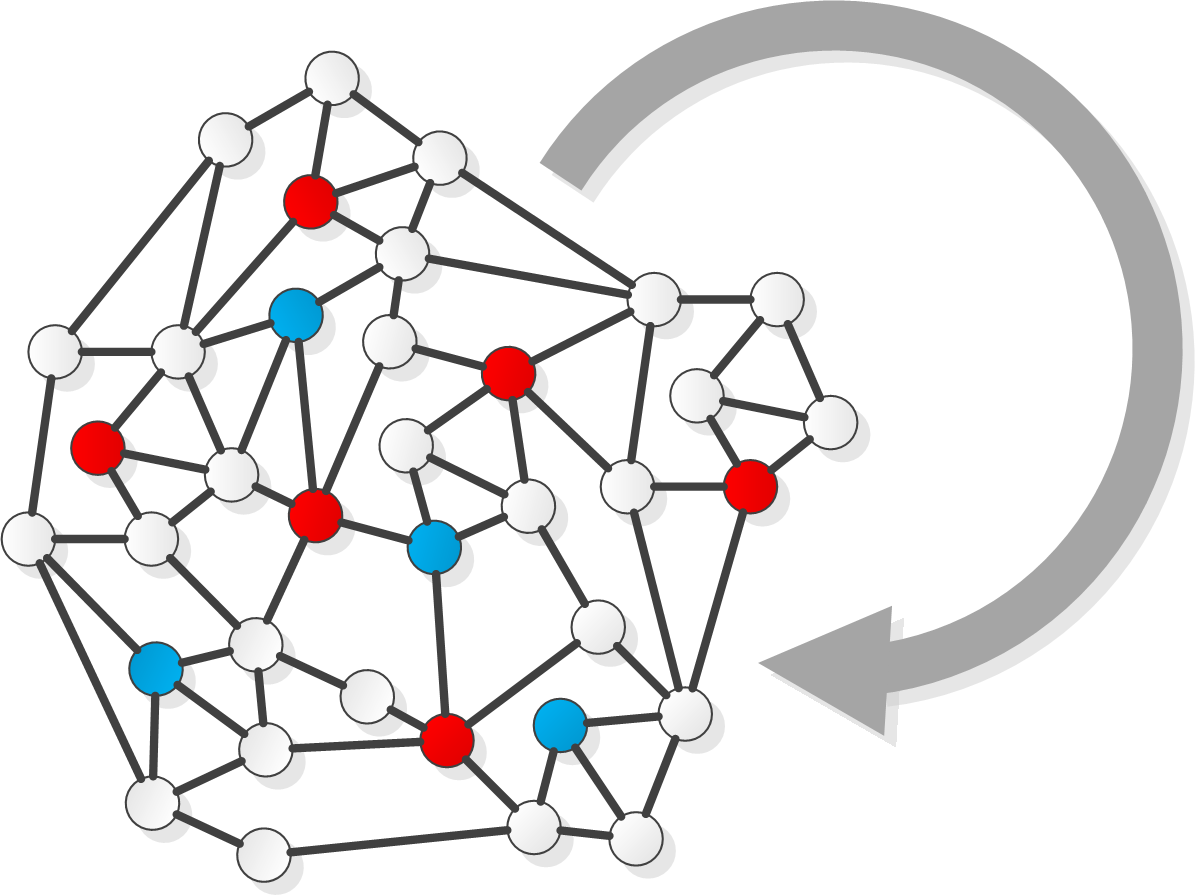}
\caption{An illustration of within-network classification task.\label{fig:withinClassification}}
\label{fig1}
\end{figure}

\begin{figure}
\centering
\includegraphics[width=1\columnwidth]{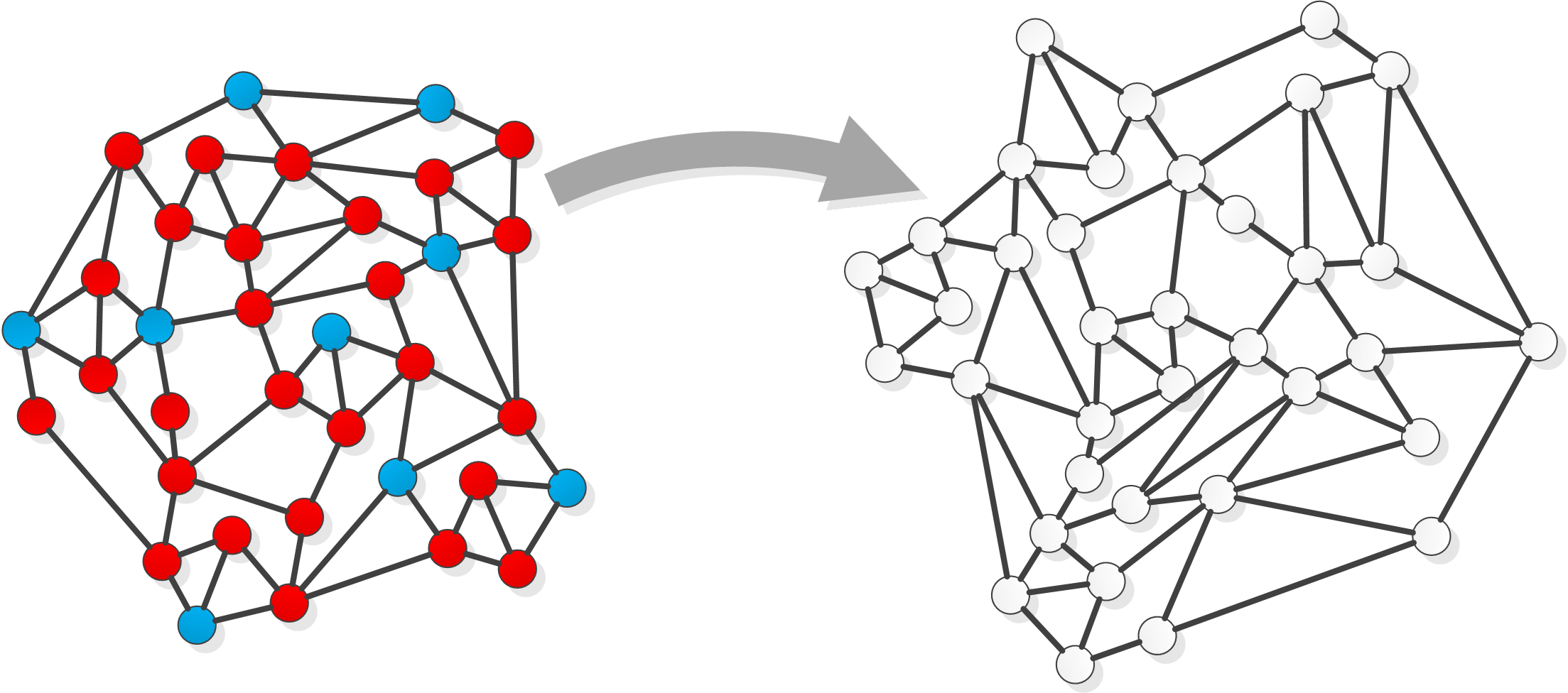}
\caption{An illustration of across-network classification task.\label{fig:accrossClassification}}
\label{fig2}
\end{figure}

Although problem of classification in traditional machine learning is not new, together with the explosion of Web--based social networks \cite{muka13}, the new branch in this area called collective (relational) classification has emerged. 
The main difference between collective classification and traditional approach to classification problem is that the former one allows data to be dependent  whereas the latter one assumes independent and identically distributed data (i.i.d.). Collective approach allows to consider both characteristics of nodes and topology of the network in the process of assigning node to a specific class. It means, that not only features of a node to be classified are taken into account but also the attributes and labels of related nodes (e.g. direct neighbours) can be considered \cite{kaka13}. 
 Two approaches can be distinguished for classification of nodes in the network (i) within-network (Figure \ref{fig1}) and (ii) across-network inference (Figure \ref{fig2}). In the within-network classification \cite{deka09} training nodes are directly connected to nodes whose labels are to be assigned in the classification process. In the across-network classification \cite{luge03} models learnt from one network are applied to another similar network.

One of the issues in collective classification is to determine set of features that should be used in order to maximize the classification accuracy.  Recent research in this area shows that the values of new attributes derived from structural properties of the graph, such as betweenness centrality, may  improve the accuracy of the classification task \cite{gael08}. Another confirmation of that fact and some more discussion about it comes from other research \cite{kaka12}. 
Another element that should be considered in the collective classification is the order of visiting nodes in the graph to perform re--labelling. The order of visiting the nodes influences the values of input values derived from the structure. To address that problem variety algorithms have been proposed. Random ordering \cite{knde01} is a simple strategy used with iterative classification algorithms and can be quite robust. Another and most popular examples of collective classification methods are: Iterative Classification Algorithm (ICA) and Gibbs Sampling Algorithm (GSA), introduced by Geman \& Geman in the image processing context \cite{gege84}. Both of them belong to so called approximate local inference algorithms that are based on local conditional classifiers \cite{sena08}. Next technique is a Loopy Belief Propagation (LBP) \cite{pear88} that is a global approximate inference method used for collective classification. But, according to recent studies \cite{kaka12a}, above mentioned methods are not robust enough to work efficiently and accurately in sparsely labelled and large-scale environments. This is a very important conclusion as majority of Web-- and technology--based social networks are very sparse and large.    Another drawback is that they cannot be easily deployed within multi-dimensional networks \cite{kamu11}, what creates space for developing new, robust collective classifiers.

According to \cite{elne12}, a very promising area of research, which may contribute to solve the issues identified above, is building compound ensemble collective classifiers. As it has been already shown, despite the fact that ensemble methods are performing accurately for i.i.d. data, there is a lack of similar analysis for relational data. For instance, in \cite{brei96} it was  presented that bagging reduces total classification error in i.i.d. data by reducing the error due to variance. The extension of i.i.d. ensembles to improve classification accuracy for relational domains has been shown in \cite{elne11}. That includes a method for constructing ensembles while accounting for the increased variance of network data and an ensemble method for reducing variance in the inference process. Another promising work \cite{faje08} showed that different ensemble method -- stacking -- improves collective classification by reducing inference bias.

\subsection{Active learning and inference in networks}

As mentioned above active learning and inference are methods used to select nodes for which labels should be acquired in order to perform collective classification \cite{angl88},  \cite{sett09}, \cite{sett11}, \cite{atpr11}, \cite{rama07}.  The main goal of these methods is to improve classification accuracy by choosing nodes in a non-random way. In contrast to passive methods where all labels are obtained once, active methods perform this task iteratively. Research results show, that in order to achieve similar accuracy, in some cases number of nodes to be queried for labels is logarithmic when comparing to passive methods \cite{beda09}. It should also be emphasized that the goal of  active inference and learning methods is different than e.g. seeding strategies where the most influential nodes are selected. Active inference and learning in the context of collective classification aims at selecting nodes that enable to minimize the classification error for the whole network.
The main drawback of active learning and inference methods is that the set of queried labels is losing its i.i.d. characteristics, what may lead to spending querying budget on bias sampling \cite{sugi06} or propagating the information in areas, where surrounding nodes may cover the inner "islands" that are labeled differently \cite{bige08}.

To overcome some of the discussed limitations active inference and learning offer different approaches when dealing with separable and non-separable data, e.g. agnostic active learning \cite{babe06} or query by committee \cite{seop92}.

One of the approaches to active learning in relational data is Reflect and Correct (RAC) method introduced in \cite{bige08} and further developed in \cite{bige10}. It is based on a simple intuition that the misclassified nodes  are gathered close to each other; they are clustered together. Thus, misclassifying one node might cause wrong labelling of neighbours. Therefore, it is reasonable to acquire the actual label of representative nodes from such clusters and use it in the inference. In order to find these clusters a label utility function can be applied \cite{bige10}. Authors introduce three types of features -- local, relational and global -- which are used as a learner of the classifier. These features measure three different aspects of misclassification. Local feature focuses on the attributes of a node, the relational takes into consideration its neighbourhood and the global feature measures the difference between the prior belief about the class distributions and posterior distributions based on the predictions. Having all these features available, authors of \cite{bige10} use a training graph and the predictions of a collective model on this graph to learn the distribution of labels. This  approach presents a reasonable assumption that we are having some budget to spend for acquiring labels. Authors compared the results of the introduced RAC method against two other approaches including their viral marketing approach and greedy acquisition showing that RAC method outperforms the others.

Another approach was proposed in \cite{nalo12}. The paper introduces a technique for node selection in an active learning framework. It selects a set of nodes that should be used in the collective classification based on a given limited budget. Using the idea of smoothness (similar distributions of independent attributes as well as relational features between nodes) it decides which nodes to select. The smoothness idea incorporates high utility from nodes that are close to each of the queried nodes. It is also proposed how to compute utility for each non--surveyed node and how to sample within the budget. However, in this method, authors assumed that network structure may not be available a priori, so the queries may reflect the labels and the neighbourhood of the node. A similar approach has been introduced in relational active learning proposal in \cite{kune11}. The key idea behind this approach was to select these nodes to acquire the label, whose predictions are potentially most certain. It is worth to emphasize that this is inconsistent with many conventional utility metrics used in i.i.d. settings, which favour labelling nodes with high uncertainty.

\section{Relational active learning and inference\label{sec:realtaionalActiveLearning}}

\subsection{The method for active learning and inference in within--network classification based on node selection}

\begin{figure}
\centering
\includegraphics[width=1.0\columnwidth]{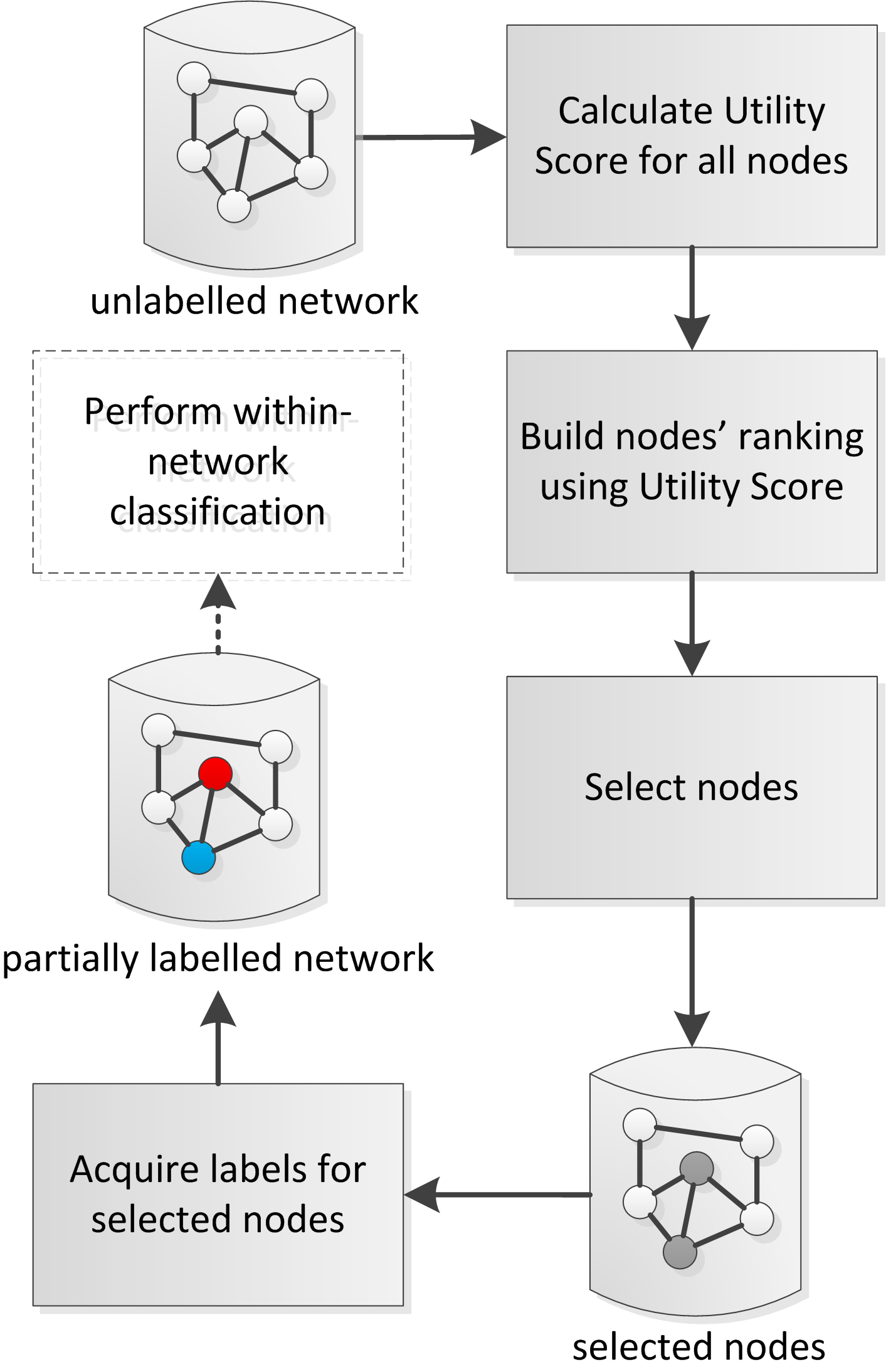}
\caption{Major steps of the active learning and inference method for within-network classification.\label{fig:method}}
\end{figure}

The proposed method for active learning and inference in within-network classification task consists of five main steps, see Figure \ref{fig:method}. First, for a given unlabelled network, the utility scores for each network node are obtained by calculating the node's structural measures such as degree centrality, closeness centrality, etc. In general, the utility score should reflect the usefulness of the node's label in the process of within--network classification. Further discussion on considered utility scores is provided in Section \ref{sec:utilityScores}. In addition, new 'measure'--neighbour approaches for assessment of the nodes' utility have been proposed in Section \ref{sec:newUtilityScores}.

Afterwards, the previously obtained utility scores are sorted in the ascending or descending order; it enables to form  nodes' ranking. Depending on the type of the utility score, the most useful nodes are these with the highest or smallest score value.  In the next step, the method selects nodes for which the label will be queried based on top $N$ items in the ranking. Once the process of label acquisition is completed, the appropriate relational classification algorithm can be applied to perform within--network classification. Note, that this last step is not the main contribution of the paper and many different classification algorithms can be applied. In this research two of them were selected as the most representative and widely used, i.e. Iterative Classification Algorithm (ICA) and Loopy Belief Propagation (LBP).

\subsection{Utility scores\label{sec:utilityScores}}

Utility scores reflect the usefulness of the node's label in the process of within--network classification. It is expected that learning the relational classification model using some previously acquired labels will result with small classification error. The process of label acquisition in active learning should be expressed as simple optimization problem of nodes selection, but in the within--network classification setting it is not possible. The general requirement of the selection mechanism is to pick those nodes' labels from label set $L$ whose usage in relational inference will result in the smallest possible misclassification error $\mathcal{E}$. 

In general terms, the expected misclassification error for all unlabelled nodes $v_i \in V^{UK}$ on their labels $Y_i^{UK}$ with $x_i$ attributes depends on abilities of relational classification algorithm $\Phi$ that is learnt on previously acquired labels $Y^{K}$ for the initial node set: 

\begin{equation}
\sum_{Y_i\setminus Y^{K}} \mathcal{E}(Y_i|X=x_i, Y^{K}, \Phi(Y^{K}))
\end{equation}
Therefore, the main problem is to obtain unknown values of $Y^{K}$ for which classification algorithm $\Phi$ will provide proper results. Then, the aggregated error must reflect an expectation over all possible values of $Y^{K}$:

\begin{equation}
\sum_{Y_i\setminus Y^{K}} \sum_{l\in L}  P(Y^{K}=l)\mathcal{E}(Y_i|X=x_i, Y^{K}=l, \Phi(Y^{K}=l))
\end{equation}

where $P(Y^{K}=l)$ is the chance that $Y^{K}$ takes a value of label $l$. Although the presented error is in general suitable for across--network classification, it does not comply with within--network classification. It is impossible to assess the correctness of classification for all nodes related to $Y_i\setminus Y^{K}$ due to the lack of these labels. Thus, it is impossible to propose any utility score that will directly lead to classification error minimization.

Nevertheless, it is still possible to make use of other utility scores that reflect structural properties of nodes, relying on the assumption that classification error depends on these measures. Depending on the characteristics of the underlying network, a proper measure can be chosen from the vast variety of nodes structural measures \cite{Musial09} such as indegree centrality, outdegree centrality, betweeness centrality, clustering coeficient, hubnes, authority, page rank.  

\subsection{A new 'measure'--neighbour approach to utility score\label{sec:newUtilityScores}}

To extend the typical structural measures approach, enumerated in Section~\ref{sec:utilityScores}, authors proposed and evaluated another method. Assuming that some nodes with the highest measures' values may actually be located on the border of 'classes', it may be useful to pick not this node itself, but its neighbours. The intuition suggests that it may hold especially for the betweenness or page rank measures, since nodes with high betweenness and page rank may be located at the border of clusters or groups or may even play the role of bridges. In this case, it may be worth acquiring the label of the neighbour instead of the bridge label itself. For example, in the case of betweenness, we identify nodes with the highest value of betweenness and select their neighbours for label acquisition. By analogy, we can create indegree--neighbour utility score, page rank--neighbour utility score, etc. All of them focus on neighbours of nodes with a given property. 

In order to confirm or reject the above concept, the authors performed  set of experiments comparing the results of this approach with the typical measure--based methods, see e.g. Table \ref{tab:ica_neighbour_results} and \ref{tab:lbp_neighbour_results}.

The neighbour selection heavily depends on the structure of the network. It may happen that in particular cases some nodes selected from rankings do not have neighbours. Therefore actual number of neighbouring nodes may be smaller than the size of sampled ranking. It was assumed that for each selected node from ranking it is selected only one neighbour. Thus, for instance if it is selected 10\% of nodes from the network according to particular ranking, we may end up with smaller than 10\% population of network constituted by neighbours (see Figure~\ref{fig:discovered}). Moreover, for the propagation algorithm (LBP) when a node from the training set has no neighbours, the information about the label during the classification process will not be propagated. On the contrary the ICA method does not depend on network structure in the sense that even if the labelled node has no neighbours, the label may be assigned to nodes in other, even disconnected, components.

However, in general, this did not have adverse effect on the obtained results. The exception was one dataset (CS\_PHD) where the network was highly disconnected and almost no nodes were labelled in the neighbour algorithm. In other cases, despite the fact that less nodes were used as an input to classification algorithm,  LBP--neighbour method outperformed classical LBP in both approaches -- top and bottom (i.e. when initial nodes were selected from top and bottom of the rankings created based on the utility scores respectively).

\section{Within-network classification algorithms}
\label{sec:Algorithms}

There exist several algorithms for within-network classification. Two of them were utilized in experiments in  Section \ref{sec:experiments}. 

The first algorithm is  the Iterative Classification Algorithm (ICA). The basic idea behind ICA is quite simple. Considering a node $v_i \in V^{UK}$, where $V^{UK}$ is a set of nodes with unknown labels, $V^{UK} \subset V$ and $V$ is the set of all nodes in the network, we aim at discovering $v_i$'s label $l_i$. Having labels of $v_i$'s neighbourhood known, ICA utilizes a local classifier $\Phi$ that takes the attribute values of nodes with known labels ($V^K$) and returns the most appropriate label value $l_i$ for $v_i$ from the class label set $L$, i.e. $l_i \in L$. If the knowledge of the neighbouring labels is only partial, the classification process needs to be repeated iteratively. In each iteration, labelling of each node $v_i$ is done using current best estimates of local classifier $\Phi$ and continues until the label assignments stabilize. A local classifier might be any function that is able to accomplish the classification task. It can range from simple to complex models like Naive Bayes, decision tree or SVM. 

Algorithm \ref{alg:ica} depicts the ICA algorithm as a pseudo-code where the local classifier is trained using the initially labelled nodes $V^K$. It can be observed that the attributes utilized in classification depend on the current label assignment (lines 8 and 9 in Algorithm \ref{alg:ica}). Thus, the repetition of classification phase needs to be performed until labels stabilize or the maximum number of iteration is reached. Computation of nodes' attributes (line 2 and 8) is the calculation of various nodes' structural measures describing profile of each node, including label--dependent and/or label--independent features \cite{kaka12}. Note that optimization of the model (line 4) is based on the local knowledge, since $x_i$ attribute of $v_i$ reflects only local information with $v_i$'s perspective.

\begin{algorithm}
\caption{Iterative Classification Algorithm (ICA)\label{alg:ica}, the idea based on \cite{sena08}}
\begin{algorithmic}[1]
\FOR{each node $v_i \in V^{UK}$}
\STATE compute $x_i$, i.e. $v_i$'s attributes using the observed (known) nodes from $V^K$
\ENDFOR
\STATE train classifier $\Phi$ by $\Theta$ optimization using the attributes of $V^K$ nodes
\REPEAT
\STATE generate ordering $O$ over nodes in $V^{UK}$
\FOR{each node $v_i \in O$}
\STATE compute $x_i$, i.e. $v_i$'s attributes using current assignments
\STATE $l_i \leftarrow \Phi(x_i,\Theta)$
\ENDFOR
\UNTIL{label stabilization or the maximum number of iterations is reached}
\end{algorithmic}
\end{algorithm}

Another method applied in experiments was the Loopy Belief Propagation algorithm (LBP). It is an alternative to ICA approach to perform collective classification. The main difference is that it defines a global objective function to be optimized, instead of performing local classifier optimization (ICA).

LBP is an iterative message--passing algorithm.  The messages are transferred between all connected nodes $v_i$ and $v_j$; where $v_i, v_j \in V$, $(v_i,v_j)\in E$, $E$ is the set of network edges. These messages might be interpreted as belief to what extent $v_j$ label should be based on $v_i$ label.

The global objective function, which is optimized in LBP, is derived from the idea of pairwise Markov Random Field (pairwise MRF) \cite{Taskar:2002}. In order to calculate the message for propagation, the calculation presented in Eq. \ref{eq:message} is performed. 

\begin{equation}
\label{eq:message}
m_{i\rightarrow j}(l_j) = \alpha \sum_{l_i \in L}\Psi_{ij}(l_i,l_j)\phi(l_i)
\prod_{v_k \in V^{UK}\setminus v_j} m_{k\rightarrow i}(l_i)
\end{equation}
where $m_{i\rightarrow j}(l_j)$ denotes a message to be sent from node $v_i$ to $v_j$, $\alpha$ is the normalization constant that ensures each message to sum to 1, $\Psi$ and $\phi$ denote the clique potentials. For further details please see \cite{sena08}.

The calculation of believe $b(l_i)$ for node $v_i$ can be expressed as in Eq. \ref{eq:belief}:

\begin{equation}
\label{eq:belief}
b(l_i) = \alpha \phi(l_i) \prod_{v_j \in V^{UK}} m_{j\rightarrow i}(l_i)
\end{equation}

The LBP algorithm consists of two main phases: message passing that is repeated until the messages are stabilized and believe computation, see Algorithm \ref{alg:lbp}.

\begin{algorithm}
\caption{Loopy Belief Propagation (LBP)\label{alg:lbp}, the idea based on \cite{sena08}}
\begin{algorithmic}[1]
\FOR{each edge $(v_i,v_j) \in E, v_i,v_j \in V^{UK}$}
\FOR{each class label $l_i \in L$}
\STATE $m_{i\rightarrow j}(l) \leftarrow 1$
\ENDFOR
\ENDFOR
\STATE //perform message passing
\REPEAT 
\FOR{each edge $(v_i,v_j) \in E, v_i,v_j \in V^{UK}$}
\FOR{each class label $l_i \in L$}
\STATE $m_{i\rightarrow j}(l_j) \leftarrow \alpha \sum_{l_i \in L}\Psi_{ij}(l_i,l_j)\phi(l_i)$\\
\STATE $\quad\quad\quad\quad\quad\prod_{v_k \in V^{UK}\setminus v_j} m_{k\rightarrow i}(l_i)$
\ENDFOR
\ENDFOR
\UNTIL{stop condition}
\STATE //compute beliefs
\FORALL{$v_i \in V^{UK}$}
\FORALL{$l_i \in L$}
\STATE $b_i(l_i)\leftarrow \alpha\phi(l_i)\prod_{v_j \in V^{UK}}m_{j\rightarrow i}(l_i)$
\ENDFOR
\ENDFOR
\end{algorithmic}
\end{algorithm}

\section{Experiments}
\label{sec:experiments}

\subsection{Experimental set--up}

In order to evaluate the proposed method for active learning and inference in terms of classification accuracy, the Iterative Classification (ICA) and Loopy Belief Propagation (LBP) algorithms were tested with various utility scores. The experimental scenario aims at examining the following structural measures used as utility scores:  

\begin{itemize}
\item indegree centrality,
\item outdegree centrality,
\item betweenness centrality,
\item clustering coefficient,
\item hubness,
\item authority,
\item page rank.
\end{itemize}

All of them were applied in two selection methods: nodes with the top (the greatest) and bottom (the smallest) values of individual scores. Independently, another new 'measure'--neighbour method proposed in Section \ref{sec:newUtilityScores} was also evaluated. Its idea is to chose the neighbours of the node with the greatest/smallest value of a given utility score.

In total, 29 selection methods were tested: 14 for original structural measures (7 measures; 'top' or 'bottom' for each), 14 for 'measure'--neighbour methods and a random selection. The random selection was repeated 14 times and the average error was taken as its final validation result.

The experiments were carried out on original dataset with labels acquired according to particular setting of selection method and utility score. Thanks to that, each dataset was split into known and unknown node sets. The models were learnt on acquired labels in nine distinct proportions (from 10\% to 90\% of known labels) and tested on the remaining part. In order to evaluate the quality of classification, the classification error was recorded. According to previously gathered experience on the configuration of the classification algorithms \cite{kaka12a} the implementation of ICA was based on Random Forest base classifier \cite{breiman2001random} and it we used $50$ iterations or $0.01$ as relative change of labels in the LBP as the stop condition. 'Measure'--neighbour version of training set selection (Section \ref{sec:newUtilityScores}) used a draw with the uniform distribution from adjacent nodes.

\subsection{Datasets} \label{sec_datasets}

The experiments were carried out on six datasets. The AMD\_NETWORK graph presents attendance at the conference seminars. The dataset was a result of the project that took place during "The Last HOPE" Conference held in July 18-20, 2008 in New York City, USA. The Radio Frequency Identification devices were distributed among participants of the conference that allowed to identify them uniquely and to track what sessions they attended. The dataset was built from the information about descriptions of participants' interests, their interactions via instant messages, as well as their location over the course of the conference. Location tracking allowed to extract a list of attendances for each conference talk. In general, the most interesting data from the experiment point of view are: information about conference participants, conference talks and presence on the talks. 

Another genealogy dataset CS\_PHD is the network that contains ties between PhD students and their advisors in theoretical computer science field where the arcs lead from advisors to students \cite{Nooy:2004}. 

The third dataset NET\_SCIENCE contains a co-authorship network of scientists working in the area of  network science \cite{Newman:2006}. It was extracted from the bibliographies of two review articles on networks. 

Another biological dataset YEAST is a protein-protein interaction network \cite{Sun:2003}.

The PAIRS\_FSG dataset is a dictionary from the University of South Florida with word association, rhyme and word fragment norms. Its graph reflects correlations between nouns, verbs and adjectives. In the experiments we use the original PAIRS\_FSG data as well as its reduced version PAIRS\_FSG\_SMALL.

\begin{table*}[h]
\centering
\caption{Basic properties of datasets utilized in experiments.}
\label{tab:datasets}
\begin{tabular}{p{2.05cm}p{.4cm}p{.5cm}p{0.5cm}p{0.7cm}p{0.6cm}p{.8cm}p{.8cm}p{1.2cm}p{1cm}p{0.9cm}p{1.1cm}p{1.5cm}}
\hline\noalign{\smallskip}
Dataset           & Group & Nodes & Edges & Directed & Classes (labels) & Avg. node degree & Avg. path length & No. of connected components & Modularity & Graph density & Network Diameter & Avg. nodes clustering coeff.\\
\hline\noalign{\smallskip}
AMD\_NETWORK      & 1     & 319  & 34385 & no  & 16      & 215.58       & 1.322         & 1          & 0.102      & 0.678         & 2        & 0.824                                                                  \\
NET\_SCIENCE      & 1     & 1588 & 2742  & yes & 26      & 1.727        & 1.997         & 395        & 0.955      & 0.001         & 7        & 0.319                                                                                                                                    \\
PAIRS\_FSG        & 2     & 4931 & 61449 & yes & 3       & 12.462       & 4.278         & 1          & 0.594      & 0.003         & 10       & 0.122                                                                  \\
PAIRS\_FSG\_SMALL & 2     & 1972 & 12213 & yes & 3       & 6.193        & 5.358         & 13         & 0.688      & 0.003         & 14       & 0.127                                                                  \\
YEAST             & 2     & 2361 & 7182  & yes & 13      & 3.042        & 4.648         & 59         & 0.59       & 0.001         & 16       & 0.065\\
CS\_PHD           & 3     & 1451 & 924   & yes & 16      & 0.636        & 2.265         & 531        & 0.967      & 0             & 10       & 0.001                                                                 \\
\noalign{\smallskip}\hline\noalign{\smallskip}
\end{tabular}
\end{table*}

The profiles of all datasets were shortly depicted in Table \ref{tab:datasets}. In order to investigate our hypothesis that the accuracy of classification depends on network characteristics, the datasets were divided into three groups (see column 'Group' in Table~\ref{tab:datasets} as well as description of groups in Table \ref{tab:datasets_groups}).  It was done based on the commonly used network's characteristics, such as average node degree, average path length,  modularity, graph density, network diameter, and average clustering coefficient of nodes. Based on those characteristics, networks that belong to Group 1 are small--world networks, those that belong to Group 2 are non--small--world modular networks and those in Group 3 are random networks that are very sparse (at the edge of phase transition condition for random networks).

  The graphs that belong to the first group\ have short average path length, the smallest network diameter out of all analysed networks, and  clustering coefficient larger than 0.3. With relatively large clustering coefficient and short average path length  networks in group 1 exhibit characteristics of small--world networks. The second group contains networks with  moderate average path length, modularity from the range $(0.5 ; 0.9)$, graph density from the range $(0.01;0.1)$ and clustering coefficient smaller than $0.15$. Those characteristics indicate that networks in this group are  non--small--world modular networks.   Group 3 contains only one dataset which is highly disconnected,  with many isolated nodes, density and clustering coefficient approaching zero, and over 400 nodes with node degree equal to 0. This group represents random networks with very low probability that the link will exist between two randomly selected nodes, which means that the giant component is not present and such networks consist of many small connected components. Further detailed characteristics of the networks followed by information about number of distinct classes are available in the Supplement Material in Sections 1 and~2.

\begin{table*}[h]
\centering
\caption{The description of groups of datasets and profiles of their collective classification results.}
\label{tab:datasets_groups}
\begin{tabular}{p{0.6cm}p{2cm}p{5cm}p{7cm}}

Group & Datasets & Network profile & Results profile\\
\hline\noalign{\smallskip}
1 & AMD\_NETWORK, NET\_SCIENCE &  small--world profile-like, short avg. path length; the smallest network diameter; clustering coeff.~\textgreater 0.3 & high error level decreasing with for increasing \% of training set; good performance of 'measure'-neighbour methods; LBP neighbour (bottom page rank) outperforms the others; most of 'measure'-neighbour methods outperform random\\
2 & YEAST, PAIRS\_FSG, PAIRS\_FSG\_SMALL &  non--small--world modular networks, moderate avg. path length; modularity $ \in (0.5;0.9) $; graph density $ \in~(0.01;0.1) $; clustering coeff. \textless 0.15 & relatively small variance of results; the more classes, the worst results; for smaller modularity and density, and greater clustering coeff. LBP-neighbour outperforms the others\\
3 & CS\_PHD & random--like network with very low probability of edges between nodes, highly disconnected; many isolated nodes (avg. node degree \textless 0.7); density 0; clustering coeff. close to 0 & LBP error \textgreater 0.8; ICA better than LBP, but still poor; 'measure'-neighbour methods worse than original and random
\end{tabular}
\end{table*}

\begin{figure*}
        \begin{subfigure}[b]{0.45\textwidth}
                \includegraphics[width=\textwidth]{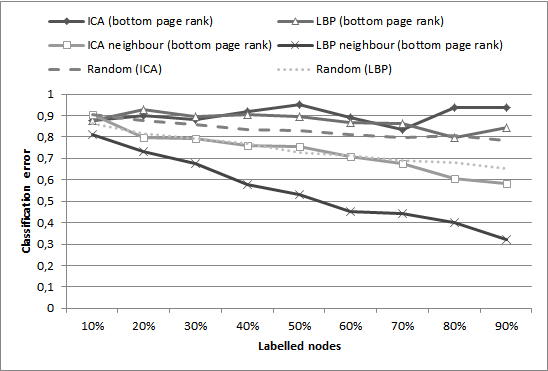}
                \caption{AMD dataset (group 1)}
                \label{fig:best_results_amd}
        \end{subfigure}
        \begin{subfigure}[b]{0.45\textwidth}
                \includegraphics[width=\textwidth]{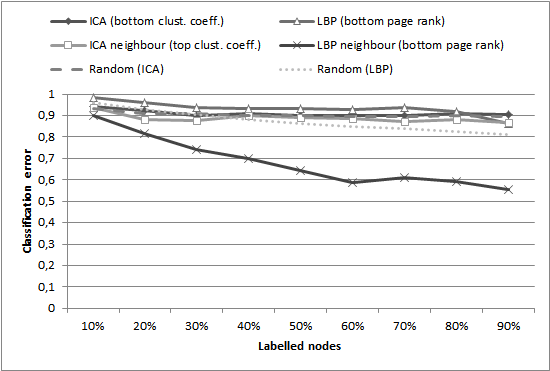}
                \caption{NET\_SCIENCE dataset (group 1)}
                \label{fig:best_results_net_science}
        \end{subfigure}
         \begin{subfigure}[b]{0.45\textwidth}
                \includegraphics[width=\textwidth]{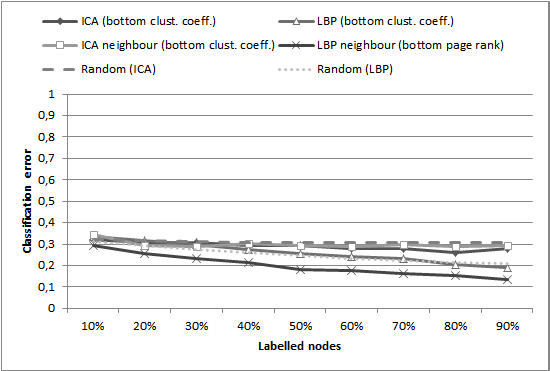}
                \caption{PAIRS\_FSG dataset (group 2)}
                \label{fig:best_results_pair_fsg}
        \end{subfigure}
        \begin{subfigure}[b]{0.45\textwidth}
                \includegraphics[width=\textwidth]{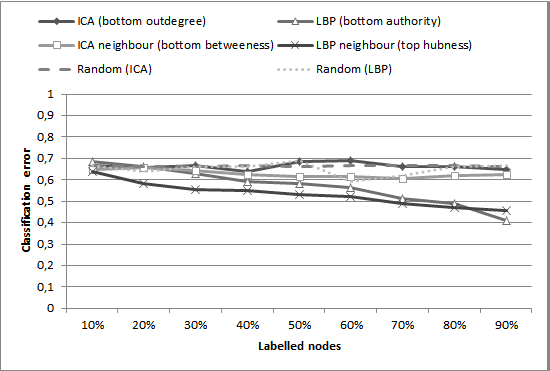}
                \caption{PAIRS\_FSG\_SMALL dataset (group 2)}
                \label{fig:best_results_pairs_fsg_small}
        \end{subfigure}
        
        \begin{subfigure}[b]{0.45\textwidth}
                \includegraphics[width=\textwidth]{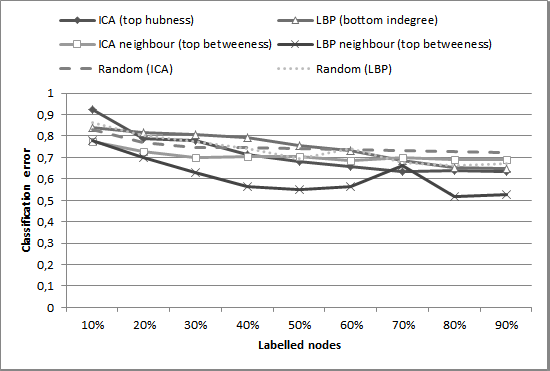}
                \caption{YEAST dataset (group 2)}
                \label{fig:best_results_yeast}
        \end{subfigure}
        \begin{subfigure}[b]{0.45\textwidth}
                \includegraphics[width=\textwidth]{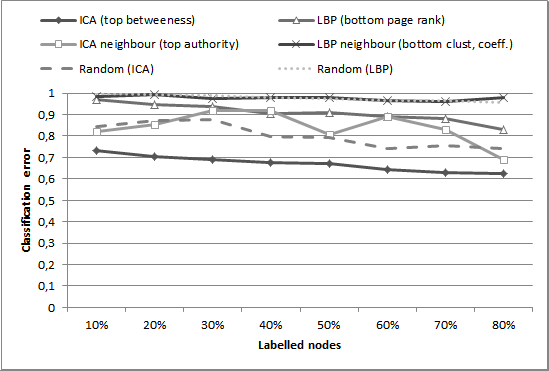}
                \caption{CS\_PHD dataset (group 3)}
                \label{fig:best_results_cs_phd}
        \end{subfigure}
        \caption{Comparison of classification error for ICA and LBP; both only for their most efficient utility scores for original utility scores and 'measure'-neighbour variants.}\label{fig:best_results}
\end{figure*}

%
%

\begin{table*}[h]
\centering
\caption{Classification error in active learning based on ICA for distinct selection strategy; initial nodes taken directly from the ranking. Note: the last column represents how many of the non--random selection strategies outperformed the random case.\label{tab:ica_results}}
\begin{tiny}
\begin{tabular}{p{1.2cm}p{0.5cm}p{0.5cm}p{0.5cm}p{0.5cm}p{0.5cm}p{0.5cm}p{0.5cm}p{0.5cm}p{0.5cm}p{0.5cm}p{0.5cm}p{0.5cm}p{0.5cm}p{0.5cm}p{0.5cm}p{0.5cm}p{0.5cm}p{0.5cm}}
Dataset & Lab. nodes & Top indegree & Top outdegree & Top betweeness & Top clust. coeff. & Top hubness & Top authority & Top page rank & Down indegree & Down outdegree & Down betweeness & Down clust. coeff. & Down hubness & Down authority & Down page rank & Random & \# better \\\hline
& 10\% & 0.983 & 0.983 & 0.91 & 0.879 & 0.976 & 0.976 & 0.91 & 0.875 & 0.875 & 0.896 & 0.91 & 0.875 & 0.875 & 0.875 & 0.906 & 7 \\
& 20\% & 0.883 & 0.883 & 0.902 & 0.981 & 0.883 & 0.883 & 0.883 & 0.898 & 0.898 & 0.906 & 0.898 & 0.938 & 0.938 & 0.898 & 0.878 & 0 \\
& 30\% & 0.897 & 0.897 & 0.906 & 0.875 & 0.911 & 0.911 & 0.933 & 0.884 & 0.884 & 0.853 & 0.897 & 0.911 & 0.911 & 0.884 & 0.858 & 1 \\
& 40\% & 0.943 & 0.943 & 0.906 & 0.938 & 0.885 & 0.891 & 0.885 & 0.912 & 0.912 & 0.927 & 0.943 & 0.912 & 0.912 & 0.917 & 0.837 & 0 \\
AMD & 50\% & 0.881 & 0.881 & 0.888 & 0.919 & 0.881 & 0.969 & 0.931 & 0.95 & 0.95 & 0.9 & 0.906 & 0.95 & 0.95 & 0.95 & 0.831 & 0 \\
& 60\% & 0.961 & 0.961 & 0.914 & 0.906 & 0.875 & 0.875 & 0.961 & 0.914 & 0.914 & 0.914 & 0.898 & 0.891 & 0.891 & 0.891 & 0.811 & 0 \\
& 70\% & 0.969 & 0.969 & 0.906 & 0.938 & 0.948 & 0.948 & 0.906 & 0.833 & 0.833 & 0.948 & 0.896 & 0.833 & 0.958 & 0.833 & 0.798 & 0 \\
& 80\% & 0.938 & 0.938 & 0.922 & 0.969 & 0.953 & 0.953 & 0.938 & 0.938 & 0.938 & 0.969 & 0.953 & 0.938 & 0.938 & 0.938 & 0.805 & 0 \\
& 90\% & 0.938 & 0.938 & 0.938 & 0.875 & 0.906 & 0.906 & 0.938 & 0.938 & 0.938 & 0.875 & 0.906 & 0.906 & 0.906 & 0.938 & 0.782 & 0 \\\hline
& 10\% & 0.967 & 0.969 & 0.941 & 0.944 & 0.954 & 0.935 & 0.934 & 0.973 & 0.91 & 0.922 & 0.919 & 0.936 & 0.935 & 0.94 & 0.935 & 4 \\
& 20\% & 0.935 & 0.925 & 0.933 & 0.926 & 0.916 & 0.918 & 0.92 & 0.911 & 0.909 & 0.922 & 0.914 & 0.942 & 0.915 & 0.923 & 0.911 & 2 \\
& 30\% & 0.906 & 0.904 & 0.933 & 0.902 & 0.902 & 0.922 & 0.912 & 0.914 & 0.911 & 0.915 & 0.921 & 0.925 & 0.918 & 0.916 & 0.908 & 4 \\
& 40\% & 0.932 & 0.909 & 0.923 & 0.911 & 0.903 & 0.91 & 0.915 & 0.93 & 0.909 & 0.912 & 0.923 & 0.903 & 0.925 & 0.92 & 0.902 & 0 \\
NET\_SCIENCE& 50\% & 0.914 & 0.904 & 0.912 & 0.899 & 0.915 & 0.918 & 0.922 & 0.91 & 0.899 & 0.922 & 0.91 & 0.926 & 0.914 & 0.912 & 0.899 & 2 \\
& 60\% & 0.906 & 0.911 & 0.925 & 0.901 & 0.923 & 0.913 & 0.923 & 0.925 & 0.954 & 0.911 & 0.916 & 0.923 & 0.927 & 0.904 & 0.898 & 0 \\
& 70\% & 0.916 & 0.916 & 0.92 & 0.902 & 0.911 & 0.916 & 0.923 & 0.925 & 0.923 & 0.913 & 0.907 & 0.902 & 0.907 & 0.913 & 0.898 & 0 \\
& 80\% & 0.891 & 0.908 & 0.922 & 0.911 & 0.922 & 0.911 & 0.922 & 0.922 & 0.98 & 0.932 & 0.918 & 0.898 & 0.915 & 0.928 & 0.898 & 2 \\
& 90\% & 0.884 & 0.898 & 0.898 & 0.905 & 0.932 & 0.918 & 0.925 & 0.905 & 0.925 & 0.939 & 0.905 & 0.918 & 0.918 & 0.905 & 0.896 & 1 \\\hline
& 10\% & 0.855 & 0.984 & 0.845 & 0.845 & 0.761 & 0.887 & 0.839 & 0.815 & 0.807 & 0.85 & 0.85 & 0.882 & 0.85 & 0.815 & 0.824 & 4 \\
& 20\% & 0.816 & 0.889 & 0.88 & 0.949 & 0.855 & 0.913 & 0.834 & 0.883 & 0.837 & 0.84 & 0.955 & 0.946 & 0.955 & 0.56 & 0.777 & 1 \\
& 30\% & 0.928 & 0.938 & 0.959 & 0.993 & 0.828 & 1 & 0.841 & 0.607 & 0.945 & 0.89 & 0.993 & 0.938 & 0.976 & 0.607 & 0.66 & 2 \\
& 40\% & 0.892 & 0.864 & 0.988 & 0.92 & 0.912 & 0.948 & 0.9 & 0.61 & 0.936 & 0.679 & 0.928 & 0.888 & 0.908 & 0.61 & 0.643 & 2 \\
PAIRS\_FSG& 50\% & 0.291 & 0.295 & 0.302 & 0.335 & 0.32 & 0.29 & 0.349 & 0.329 & 0.319 & 0.337 & 0.292 & 0.333 & 0.339 & 0.334 & 0.308 & 5 \\
& 60\% & 0.288 & 0.295 & 0.295 & 0.335 & 0.325 & 0.284 & 0.293 & 0.328 & 0.325 & 0.331 & 0.28 & 0.312 & 0.443 & 0.528 & 0.307 & 6 \\
& 70\% & 0.279 & 0.297 & 0.293 & 0.332 & 0.33 & 0.28 & 0.291 & 0.335 & 0.313 & 0.341 & 0.279 & 0.313 & 0.528 & 0.346 & 0.306 & 6 \\
& 80\% & 0.27 & 0.293 & 0.259 & 0.333 & 0.339 & 0.271 & 0.276 & 0.344 & 0.308 & 0.35 & 0.26 & 0.297 & 0.439 & 0.36 & 0.309 & 8 \\
& 90\% & 0.257 & 0.296 & 0.237 & 0.326 & 0.342 & 0.263 & 0.263 & 0.362 & 0.3 & 0.37 & 0.281 & 0.281 & 0.577 & 0.383 & 0.305 & 8 \\\hline
& 10\% & 0.671 & 0.668 & 0.683 & 0.668 & 0.689 & 0.695 & 0.687 & 0.661 & 0.669 & 0.701 & 0.704 & 0.709 & 0.7 & 0.699 & 0.665 & 1 \\
& 20\% & 0.654 & 0.678 & 0.695 & 0.695 & 0.715 & 0.715 & 0.697 & 0.671 & 0.657 & 0.713 & 0.737 & 0.732 & 0.734 & 0.706 & 0.661 & 2 \\
& 30\% & 0.668 & 0.673 & 0.715 & 0.709 & 0.735 & 0.736 & 0.701 & 0.68 & 0.669 & 0.738 & 0.739 & 0.757 & 0.737 & 0.723 & 0.664 & 0 \\
& 40\% & 0.663 & 0.689 & 0.726 & 0.698 & 0.757 & 0.76 & 0.716 & 0.676 & 0.641 & 0.728 & 0.737 & 0.754 & 0.755 & 0.727 & 0.667 & 2 \\
PAIRS\_FSG\_SM& 50\% & 0.663 & 0.663 & 0.725 & 0.721 & 0.756 & 0.78 & 0.741 & 0.695 & 0.686 & 0.74 & 0.714 & 0.762 & 0.769 & 0.728 & 0.665 & 2 \\
& 60\% & 0.679 & 0.696 & 0.761 & 0.747 & 0.778 & 0.804 & 0.752 & 0.689 & 0.689 & 0.738 & 0.703 & 0.762 & 0.768 & 0.731 & 0.667 & 0 \\
& 70\% & 0.664 & 0.652 & 0.776 & 0.786 & 0.813 & 0.827 & 0.769 & 0.674 & 0.664 & 0.759 & 0.725 & 0.778 & 0.774 & 0.723 & 0.665 & 3 \\
& 80\% & 0.654 & 0.662 & 0.809 & 0.835 & 0.83 & 0.84 & 0.794 & 0.702 & 0.664 & 0.761 & 0.705 & 0.756 & 0.779 & 0.748 & 0.669 & 3 \\
& 90\% & 0.701 & 0.665 & 0.767 & 0.868 & 0.893 & 0.843 & 0.797 & 0.614 & 0.65 & 0.812 & 0.695 & 0.756 & 0.787 & 0.782 & 0.661 & 2 \\\hline
& 10\% & 0.965 & 0.95 & 0.93 & 0.751 & 0.922 & 0.934 & 0.873 & 0.928 & 0.881 & 0.94 & 0.801 & 0.811 & 0.774 & 0.846 & 0.832 & 4 \\
& 20\% & 0.903 & 0.901 & 0.78 & 0.776 & 0.788 & 0.768 & 0.765 & 0.75 & 0.766 & 0.816 & 0.778 & 0.816 & 0.833 & 0.796 & 0.768 & 4 \\
& 30\% & 0.917 & 0.92 & 0.77 & 0.762 & 0.78 & 0.751 & 0.771 & 0.754 & 0.79 & 0.804 & 0.777 & 0.831 & 0.811 & 0.784 & 0.748 & 0 \\
& 40\% & 0.926 & 0.713 & 0.704 & 0.739 & 0.714 & 0.741 & 0.768 & 0.771 & 0.917 & 0.834 & 0.788 & 0.828 & 0.812 & 0.773 & 0.744 & 5 \\
YEAST & 50\% & 0.713 & 0.695 & 0.88 & 0.723 & 0.681 & 0.744 & 0.761 & 0.813 & 0.811 & 0.873 & 0.819 & 0.861 & 0.8 & 0.781 & 0.74 & 4 \\
& 60\% & 0.924 & 0.654 & 0.661 & 0.697 & 0.658 & 0.741 & 0.753 & 0.802 & 0.935 & 0.827 & 0.843 & 0.87 & 0.8 & 0.777 & 0.737 & 4 \\
& 70\% & 0.647 & 0.64 & 0.636 & 0.687 & 0.633 & 0.749 & 0.742 & 0.791 & 0.877 & 0.831 & 0.922 & 0.891 & 0.804 & 0.767 & 0.73 & 5 \\
& 80\% & 0.62 & 0.628 & 0.628 & 0.679 & 0.639 & 0.761 & 0.738 & 0.799 & 0.899 & 0.844 & 0.875 & 0.913 & 0.812 & 0.793 & 0.729 & 5 \\
& 90\% & 0.612 & 0.629 & 0.629 & 0.637 & 0.633 & 0.7 & 0.738 & 0.831 & 0.882 & 0.827 & 0.831 & 0.873 & 0.785 & 0.793 & 0.724 & 6 \\\hline
& 10\% & 0.927 & 0.939 & 0.851 & 0.754 & 0.755 & 0.774 & 0.941 & 0.94 & 0.777 & 0.829 & 0.82 & 0.751 & 0.75 & 0.973 & 0.835 & 8 \\
& 20\% & 0.954 & 0.919 & 0.733 & 0.912 & 0.892 & 0.926 & 0.686 & 0.936 & 0.919 & 0.925 & 0.882 & 0.908 & 0.909 & 0.915 & 0.844 & 2 \\
& 30\% & 0.802 & 0.929 & 0.704 & 0.797 & 0.797 & 0.927 & 0.668 & 0.911 & 0.918 & 0.939 & 0.884 & 0.797 & 0.794 & 0.9 & 0.873 & 7 \\
& 40\% & 0.812 & 0.641 & 0.689 & 0.843 & 0.841 & 0.787 & 0.653 & 0.771 & 0.761 & 0.794 & 0.841 & 0.843 & 0.849 & 0.896 & 0.876 & 13 \\
CS\_PHD& 50\% & 0.676 & 0.71 & 0.678 & 0.706 & 0.71 & 0.646 & 0.691 & 0.87 & 0.912 & 0.782 & 0.708 & 0.708 & 0.659 & 0.718 & 0.797 & 12 \\
& 60\% & 0.689 & 0.617 & 0.673 & 0.689 & 0.689 & 0.635 & 0.706 & 0.739 & 0.692 & 0.647 & 0.692 & 0.692 & 0.697 & 0.715 & 0.792 & 14 \\
& 70\% & 0.636 & 0.649 & 0.646 & 0.699 & 0.696 & 0.643 & 0.696 & 0.73 & 0.687 & 0.743 & 0.699 & 0.699 & 0.712 & 0.705 & 0.741 & 13 \\
& 80\% & 0.648 & 0.61 & 0.629 & 0.606 & 0.61 & 0.615 & 0.732 & 0.714 & 0.676 & 0.695 & 0.606 & 0.61 & 0.61 & 0.653 & 0.756 & 14 \\
& 90\% & 0.794 & 0.561 & 0.626 & 0.626 & 0.589 & 0.589 & 0.776 & 0.738 & 0.636 & 0.71 & 0.617 & 0.589 & 0.589 & 0.664 & 0.742 & 12
\end{tabular}
\end{tiny}
\end{table*}

\begin{table*}[h]
\centering
\caption{Classification error in active learning based on 'measure'-neighbour version of distinct selection strategy with ICA. Note: the last column represents how many of the non--random selection strategies outperformed the random case.\label{tab:ica_neighbour_results}}
\begin{tiny}
\begin{tabular}{p{1.2cm}p{0.5cm}p{0.5cm}p{0.5cm}p{0.5cm}p{0.5cm}p{0.5cm}p{0.5cm}p{0.5cm}p{0.5cm}p{0.5cm}p{0.5cm}p{0.5cm}p{0.5cm}p{0.5cm}p{0.5cm}p{0.5cm}p{0.5cm}}
Dataset & Lab. nodes & Top indegree & Top outdegree & Top betweeness & Top clust. coeff. & Top hubness & Top authority & Top page rank & Down indegree & Down outdegree & Down betweeness & Down clust. coeff. & Down hubness & Down authority & Down page rank & Random & \# better\\\hline

& 10\% & 0.869 & 0.927 & 0.907 & 0.883 & 0.886 & 0.89 & 0.855 & 0.892 & 0.889 & 0.865 & 0.903 & 0.907 & 0.896 & 0.917 & 0.906 & 10 \\
& 20\% & 0.851 & 0.87 & 0.799 & 0.865 & 0.837 & 0.844 & 0.883 & 0.894 & 0.898 & 0.822 & 0.821 & 0.821 & 0.897 & 0.83 & 0.878 & 10 \\
& 30\% & 0.83 & 0.742 & 0.795 & 0.781 & 0.805 & 0.822 & 0.846 & 0.8 & 0.789 & 0.777 & 0.807 & 0.812 & 0.794 & 0.828 & 0.858 & 14 \\
& 40\% & 0.749 & 0.75 & 0.763 & 0.816 & 0.726 & 0.78 & 0.728 & 0.749 & 0.772 & 0.779 & 0.786 & 0.772 & 0.776 & 0.766 & 0.837 & 14 \\
AMD & 50\% & 0.738 & 0.798 & 0.757 & 0.768 & 0.719 & 0.704 & 0.742 & 0.706 & 0.746 & 0.762 & 0.722 & 0.683 & 0.757 & 0.788 & 0.831 & 14 \\
& 60\% & 0.751 & 0.762 & 0.707 & 0.722 & 0.673 & 0.708 & 0.71 & 0.768 & 0.694 & 0.794 & 0.711 & 0.785 & 0.706 & 0.738 & 0.811 & 14 \\
& 70\% & 0.669 & 0.752 & 0.677 & 0.736 & 0.689 & 0.69 & 0.652 & 0.7 & 0.739 & 0.671 & 0.706 & 0.691 & 0.727 & 0.703 & 0.798 & 14 \\
& 80\% & 0.678 & 0.671 & 0.604 & 0.651 & 0.662 & 0.676 & 0.653 & 0.694 & 0.729 & 0.747 & 0.673 & 0.712 & 0.722 & 0.724 & 0.805 & 14 \\
& 90\% & 0.614 & 0.662 & 0.583 & 0.613 & 0.635 & 0.629 & 0.593 & 0.671 & 0.689 & 0.669 & 0.669 & 0.63 & 0.709 & 0.672 & 0.782 & 14 \\\hline
& 10\% & 0.937 & 0.914 & 0.928 & 0.937 & 0.943 & 0.939 & 0.932 & 0.937 & 0.935 & 0.964 & 0.935 & 0.953 & 0.933 & 0.94 & 0.935 & 5 \\
& 20\% & 0.904 & 0.905 & 0.901 & 0.884 & 0.932 & 0.938 & 0.927 & 0.922 & 0.94 & 0.952 & 0.935 & 0.961 & 0.936 & 0.91 & 0.911 & 5 \\
& 30\% & 0.901 & 0.902 & 0.916 & 0.876 & 0.914 & 0.893 & 0.911 & 0.903 & 0.893 & 0.899 & 0.899 & 0.928 & 0.915 & 0.885 & 0.908 & 9 \\
& 40\% & 0.901 & 0.905 & 0.893 & 0.901 & 0.897 & 0.889 & 0.916 & 0.895 & 0.891 & 0.896 & 0.903 & 0.915 & 0.933 & 0.893 & 0.902 & 9 \\
NET\_SCIENCE& 50\% & 0.886 & 0.885 & 0.89 & 0.89 & 0.902 & 0.91 & 0.894 & 0.883 & 0.884 & 0.905 & 0.942 & 0.917 & 0.925 & 0.887 & 0.899 & 8 \\
& 60\% & 0.902 & 0.891 & 0.881 & 0.885 & 0.904 & 0.898 & 0.897 & 0.882 & 0.891 & 0.884 & 0.897 & 0.906 & 0.903 & 0.89 & 0.898 & 9 \\
& 70\% & 0.897 & 0.887 & 0.89 & 0.87 & 0.892 & 0.902 & 0.888 & 0.882 & 0.912 & 0.879 & 0.892 & 0.917 & 0.918 & 0.867 & 0.898 & 10 \\
& 80\% & 0.879 & 0.885 & 0.899 & 0.88 & 0.888 & 0.893 & 0.897 & 0.876 & 0.871 & 0.882 & 0.9 & 0.926 & 0.928 & 0.861 & 0.898 & 10 \\
& 90\% & 0.892 & 0.87 & 0.899 & 0.869 & 0.872 & 0.907 & 0.881 & 0.889 & 0.877 & 0.884 & 0.88 & 0.917 & 0.916 & 0.873 & 0.896 & 10 \\\hline
& 10\% & 0.796 & 0.965 & 0.807 & 0.83 & 0.767 & 0.98 & 0.977 & 0.801 & 1 & 0.803 & 0.824 & 0.844 & 0.826 & 0.979 & 0.824 & 6 \\
& 20\% & 0.756 & 0.834 & 0.824 & 0.859 & 0.73 & 0.81 & 0.809 & 0.802 & 0.827 & 0.806 & 0.85 & 0.842 & 0.874 & 0.793 & 0.777 & 2 \\
& 30\% & 0.839 & 0.854 & 0.838 & 0.89 & 0.787 & 0.835 & 0.828 & 0.787 & 0.817 & 0.828 & 0.877 & 0.882 & 0.888 & 0.78 & 0.66 & 0 \\
& 40\% & 0.834 & 0.808 & 0.799 & 0.902 & 0.776 & 0.861 & 0.829 & 0.777 & 0.852 & 0.82 & 0.908 & 0.904 & 0.901 & 0.781 & 0.643 & 0 \\
PAIRS\_FSG& 50\% & 0.293 & 0.291 & 0.301 & 0.305 & 0.31 & 0.299 & 0.302 & 0.303 & 0.295 & 0.303 & 0.293 & 0.295 & 0.301 & 0.297 & 0.308 & 13 \\
& 60\% & 0.293 & 0.307 & 0.298 & 0.3 & 0.306 & 0.301 & 0.303 & 0.301 & 0.302 & 0.293 & 0.287 & 0.297 & 0.302 & 0.3 & 0.307 & 14 \\
& 70\% & 0.294 & 0.295 & 0.306 & 0.306 & 0.304 & 0.294 & 0.298 & 0.293 & 0.301 & 0.295 & 0.296 & 0.296 & 0.295 & 0.297 & 0.306 & 14 \\
& 80\% & 0.287 & 0.294 & 0.301 & 0.307 & 0.299 & 0.294 & 0.294 & 0.296 & 0.305 & 0.297 & 0.287 & 0.298 & 0.301 & 0.294 & 0.309 & 14 \\
& 90\% & 0.29 & 0.286 & 0.302 & 0.301 & 0.31 & 0.304 & 0.305 & 0.299 & 0.291 & 0.297 & 0.291 & 0.294 & 0.291 & 0.291 & 0.305 & 13 \\\hline
& 10\% & 0.636 & 0.668 & 0.662 & 0.658 & 0.662 & 0.669 & 0.666 & 0.665 & 0.661 & 0.647 & 0.646 & 0.663 & 0.664 & 0.647 & 0.665 & 10 \\
& 20\% & 0.66 & 0.663 & 0.658 & 0.655 & 0.671 & 0.673 & 0.671 & 0.662 & 0.647 & 0.659 & 0.641 & 0.653 & 0.643 & 0.65 & 0.661 & 9 \\
& 30\% & 0.66 & 0.658 & 0.673 & 0.65 & 0.676 & 0.694 & 0.675 & 0.652 & 0.648 & 0.643 & 0.651 & 0.639 & 0.647 & 0.63 & 0.664 & 10 \\
& 40\% & 0.651 & 0.657 & 0.682 & 0.654 & 0.668 & 0.678 & 0.652 & 0.644 & 0.65 & 0.627 & 0.649 & 0.642 & 0.646 & 0.624 & 0.667 & 11 \\
PAIRS\_FSG\_SM& 50\% & 0.663 & 0.647 & 0.665 & 0.662 & 0.666 & 0.7 & 0.68 & 0.652 & 0.65 & 0.617 & 0.635 & 0.636 & 0.648 & 0.617 & 0.665 & 10 \\
& 60\% & 0.657 & 0.656 & 0.664 & 0.657 & 0.68 & 0.699 & 0.681 & 0.647 & 0.655 & 0.617 & 0.628 & 0.631 & 0.629 & 0.632 & 0.667 & 11 \\
& 70\% & 0.666 & 0.654 & 0.682 & 0.661 & 0.671 & 0.681 & 0.676 & 0.623 & 0.662 & 0.606 & 0.624 & 0.637 & 0.656 & 0.617 & 0.665 & 9 \\
& 80\% & 0.646 & 0.644 & 0.662 & 0.66 & 0.678 & 0.702 & 0.682 & 0.643 & 0.658 & 0.622 & 0.609 & 0.623 & 0.598 & 0.614 & 0.669 & 11 \\
& 90\% & 0.66 & 0.656 & 0.679 & 0.674 & 0.677 & 0.698 & 0.673 & 0.64 & 0.669 & 0.624 & 0.622 & 0.598 & 0.599 & 0.636 & 0.661 & 8 \\\hline
& 10\% & 0.756 & 0.89 & 0.776 & 0.909 & 0.925 & 0.784 & 0.889 & 0.892 & 1 & 0.925 & 0.9 & 0.941 & 0.93 & 0.941 & 0.832 & 3 \\
& 20\% & 0.711 & 0.724 & 0.726 & 0.715 & 0.763 & 0.721 & 0.738 & 0.912 & 1 & 0.917 & 0.746 & 0.934 & 0.777 & 0.951 & 0.768 & 8 \\
& 30\% & 0.706 & 0.697 & 0.702 & 0.705 & 0.695 & 0.703 & 0.723 & 0.74 & 1 & 0.855 & 0.845 & 0.776 & 0.796 & 0.723 & 0.748 & 9 \\
& 40\% & 0.709 & 0.705 & 0.702 & 0.703 & 0.69 & 0.687 & 0.728 & 0.732 & 0.947 & 0.756 & 0.725 & 0.777 & 0.748 & 0.727 & 0.744 & 10 \\
YEAST & 50\% & 0.705 & 0.705 & 0.703 & 0.698 & 0.674 & 0.695 & 0.711 & 0.749 & 0.935 & 0.765 & 0.716 & 0.772 & 0.714 & 0.712 & 0.74 & 10 \\
& 60\% & 0.724 & 0.685 & 0.686 & 0.695 & 0.692 & 0.695 & 0.715 & 0.709 & 0.729 & 0.72 & 0.722 & 0.73 & 0.696 & 0.699 & 0.737 & 14 \\
& 70\% & 0.687 & 0.689 & 0.699 & 0.686 & 0.683 & 0.688 & 0.707 & 0.698 & 0.719 & 0.692 & 0.702 & 0.714 & 0.718 & 0.687 & 0.73 & 14 \\
& 80\% & 0.68 & 0.701 & 0.689 & 0.683 & 0.681 & 0.696 & 0.687 & 0.687 & 0.704 & 0.692 & 0.7 & 0.706 & 0.684 & 0.691 & 0.729 & 14 \\
& 90\% & 0.699 & 0.693 & 0.692 & 0.687 & 0.68 & 0.711 & 0.704 & 0.684 & 0.683 & 0.683 & 0.698 & 0.71 & 0.685 & 0.687 & 0.724 & 14 \\\hline
& 10\% & 0.934 & 0.953 & 0.884 & 0.696 & 0.83 & 0.692 & 0.892 & 0.951 & 0.781 & 0.748 & 0.762 & 0.78 & 0.944 & 0.919 & 0.835 & 7 \\
& 20\% & 0.975 & 0.76 & 0.652 & 0.721 & 0.715 & 0.82 & 0.897 & 0.928 & 0.963 & 0.914 & 0.854 & 0.679 & 0.901 & 0.93 & 0.844 & 6 \\
& 30\% & 0.891 & 0.85 & 0.93 & 0.814 & 0.917 & 0.855 & 0.883 & 0.936 & 0.848 & 0.767 & 0.782 & 0.792 & 0.946 & 0.922 & 0.873 & 7 \\
& 40\% & 0.937 & 0.908 & 0.741 & 0.95 & 0.949 & 0.921 & 0.964 & 0.875 & 0.779 & 0.843 & 0.979 & 0.896 & 0.931 & 0.923 & 0.876 & 4 \\
CS\_PHD& 50\% & 0.878 & 0.938 & 0.819 & 0.714 & 0.658 & 0.921 & 0.939 & 0.902 & 0.922 & 0.956 & 0.902 & 0.938 & 0.819 & 0.956 & 0.797 & 2 \\
& 60\% & 0.923 & 0.828 & 0.939 & 0.801 & 0.818 & 0.807 & 0.96 & 0.91 & 0.803 & 0.911 & 0.774 & 0.816 & 0.951 & 0.789 & 0.792 & 2 \\
& 70\% & 0.775 & 0.701 & 0.878 & 0.958 & 0.765 & 0.889 & 0.914 & 0.86 & 0.791 & 0.854 & 0.891 & 0.93 & 0.872 & 0.917 & 0.741 & 1 \\
& 80\% & 0.68 & 0.958 & 0.921 & 0.942 & 0.925 & 0.829 & 0.803 & 0.969 & 0.968 & 0.932 & 0.839 & 0.922 & 0.93 & 0.848 & 0.756 & 1 \\
& 90\% & 0.814 & 0.831 & 0.922 & 0.947 & 0.951 & 0.689 & 0.929 & 0.913 & 0.9 & 0.645 & 0.886 & 0.816 & 0.956 & 0.934 & 0.742 & 2
\end{tabular}
\end{tiny}
\end{table*}

\begin{table*}[h]
\centering
\caption{Classification error in active inference based on LBP for distinct selection strategy; initial nodes taken directly from the ranking. Note: the last column represents how many of the non--random selection strategies outperformed the random case.\label{tab:lbp_results}}
\begin{tiny}
\begin{tabular}{p{1.2cm}p{0.5cm}p{0.5cm}p{0.5cm}p{0.5cm}p{0.5cm}p{0.5cm}p{0.5cm}p{0.5cm}p{0.5cm}p{0.5cm}p{0.5cm}p{0.5cm}p{0.5cm}p{0.5cm}p{0.5cm}p{0.5cm}p{0.5cm}p{0.5cm}}
Dataset & Lab. nodes & Top indegree & Top outdegree & Top betweeness & Top clust. coeff. & Top hubness & Top authority & Top page rank & Down indegree & Down outdegree & Down betweeness & Down clust. coeff. & Down hubness & Down authority & Down page rank & Random & \# better\\\hline
& 10\% & 0.931 & 0.931 & 0.882 & 0.868 & 0.931 & 0.931 & 0.927 & 0.913 & 0.913 & 0.896 & 0.879 & 0.917 & 0.917 & 0.913 & 0.863 & 0 \\
& 20\% & 0.938 & 0.938 & 0.934 & 0.891 & 0.938 & 0.938 & 0.938 & 0.895 & 0.895 & 0.898 & 0.93 & 0.898 & 0.898 & 0.895 & 0.817 & 0 \\
& 30\% & 0.897 & 0.897 & 0.893 & 0.888 & 0.897 & 0.897 & 0.893 & 0.884 & 0.884 & 0.884 & 0.897 & 0.866 & 0.866 & 0.884 & 0.793 & 0 \\
& 40\% & 0.901 & 0.901 & 0.912 & 0.917 & 0.901 & 0.901 & 0.901 & 0.896 & 0.896 & 0.917 & 0.906 & 0.896 & 0.896 & 0.917 & 0.77 & 0 \\
AMD & 50\% & 0.894 & 0.894 & 0.888 & 0.9 & 0.894 & 0.894 & 0.894 & 0.925 & 0.925 & 0.919 & 0.894 & 0.925 & 0.925 & 0.944 & 0.729 & 0 \\
& 60\% & 0.891 & 0.891 & 0.852 & 0.906 & 0.891 & 0.891 & 0.891 & 0.883 & 0.883 & 0.922 & 0.867 & 0.914 & 0.914 & 0.922 & 0.714 & 0 \\
& 70\% & 0.833 & 0.833 & 0.865 & 0.917 & 0.833 & 0.833 & 0.833 & 0.948 & 0.948 & 0.938 & 0.865 & 0.948 & 0.948 & 0.948 & 0.692 & 0 \\
& 80\% & 0.828 & 0.828 & 0.844 & 0.984 & 0.828 & 0.828 & 0.828 & 0.953 & 0.953 & 0.953 & 0.797 & 0.953 & 0.953 & 0.953 & 0.681 & 0 \\
& 90\% & 0.875 & 0.875 & 0.875 & 0.906 & 0.906 & 0.906 & 0.875 & 0.938 & 0.938 & 0.969 & 0.844 & 0.938 & 0.938 & 0.938 & 0.651 & 0 \\\hline
& 10\% & 0.995 & 0.986 & 0.992 & 0.995 & 0.999 & 0.999 & 0.998 & 0.986 & 0.997 & 0.998 & 0.999 & 0.999 & 0.999 & 0.984 & 0.959 & 0 \\
& 20\% & 0.986 & 0.986 & 0.992 & 0.986 & 0.999 & 0.999 & 0.997 & 0.977 & 0.987 & 0.993 & 0.997 & 0.999 & 0.999 & 0.962 & 0.929 & 0 \\
& 30\% & 0.981 & 0.984 & 0.992 & 0.979 & 0.999 & 0.999 & 0.998 & 0.965 & 0.974 & 0.988 & 0.997 & 0.999 & 0.999 & 0.937 & 0.904 & 0 \\
& 40\% & 0.981 & 0.97 & 0.993 & 0.967 & 0.999 & 0.999 & 0.998 & 0.966 & 0.97 & 0.982 & 0.995 & 0.999 & 0.999 & 0.932 & 0.883 & 0 \\
NET\_SCIENCE& 50\% & 0.982 & 0.967 & 0.999 & 0.96 & 0.999 & 0.999 & 0.999 & 0.96 & 0.963 & 0.981 & 0.996 & 0.999 & 0.999 & 0.932 & 0.861 & 0 \\
& 60\% & 0.983 & 0.966 & 0.998 & 0.956 & 0.998 & 0.998 & 0.998 & 0.957 & 0.968 & 0.974 & 0.99 & 0.998 & 0.998 & 0.928 & 0.848 & 0 \\
& 70\% & 0.975 & 0.961 & 0.998 & 0.957 & 0.998 & 0.998 & 0.998 & 0.964 & 0.975 & 0.964 & 0.995 & 0.998 & 0.998 & 0.939 & 0.84 & 0 \\
& 80\% & 0.963 & 0.952 & 0.997 & 0.959 & 0.997 & 0.997 & 0.997 & 0.956 & 0.973 & 0.959 & 0.997 & 0.997 & 0.997 & 0.918 & 0.827 & 0 \\
& 90\% & 0.959 & 0.939 & 0.993 & 0.959 & 0.993 & 0.993 & 0.993 & 0.939 & 0.973 & 0.946 & 0.993 & 0.993 & 0.993 & 0.864 & 0.813 & 0 \\\hline
& 10\% & 0.965 & 0.893 & 0.957 & 0.936 & 0.995 & 0.941 & 0.968 & 0.92 & 1 & 0.938 & 0.941 & 0.936 & 0.941 & 0.92 & 0.908 & 1 \\
& 20\% & 0.946 & 0.877 & 0.919 & 0.976 & 0.943 & 0.925 & 0.994 & 0.889 & 0.946 & 0.934 & 0.982 & 0.979 & 0.973 & 0.889 & 0.86 & 0 \\
& 30\% & 0.986 & 0.876 & 0.893 & 0.993 & 0.921 & 0.969 & 0.997 & 0.848 & 0.972 & 0.938 & 0.997 & 0.972 & 0.976 & 0.848 & 0.844 & 0 \\
& 40\% & 0.972 & 0.859 & 0.932 & 0.992 & 0.98 & 0.96 & 1 & 0.823 & 0.964 & 0.912 & 0.996 & 0.976 & 0.972 & 0.823 & 0.823 & 0 \\
PAIRS\_FSG& 50\% & 0.29 & 0.272 & 0.278 & 0.305 & 0.316 & 0.285 & 0.295 & 0.274 & 0.295 & 0.259 & 0.258 & 0.275 & 0.268 & 0.276 & 0.245 & 0 \\
& 60\% & 0.284 & 0.261 & 0.276 & 0.315 & 0.286 & 0.289 & 0.3 & 0.259 & 0.282 & 0.24 & 0.241 & 0.273 & 0.262 & 0.265 & 0.234 & 0 \\
& 70\% & 0.309 & 0.257 & 0.287 & 0.298 & 0.28 & 0.309 & 0.318 & 0.257 & 0.257 & 0.247 & 0.232 & 0.276 & 0.257 & 0.27 & 0.223 & 0 \\
& 80\% & 0.33 & 0.257 & 0.309 & 0.276 & 0.243 & 0.326 & 0.347 & 0.248 & 0.25 & 0.237 & 0.207 & 0.263 & 0.255 & 0.273 & 0.213 & 1 \\
& 90\% & 0.458 & 0.253 & 0.451 & 0.294 & 0.245 & 0.456 & 0.476 & 0.267 & 0.231 & 0.235 & 0.19 & 0.235 & 0.259 & 0.292 & 0.208 & 1 \\\hline
& 10\% & 0.667 & 0.722 & 0.678 & 0.681 & 0.726 & 0.681 & 0.665 & 0.712 & 0.671 & 0.716 & 0.727 & 0.709 & 0.685 & 0.684 & 0.666 & 1 \\
& 20\% & 0.666 & 0.63 & 0.679 & 0.669 & 0.692 & 0.673 & 0.679 & 0.665 & 0.664 & 0.653 & 0.648 & 0.745 & 0.66 & 0.639 & 0.639 & 1 \\
& 30\% & 0.68 & 0.641 & 0.696 & 0.656 & 0.682 & 0.696 & 0.702 & 0.693 & 0.644 & 0.623 & 0.623 & 0.64 & 0.632 & 0.603 & 0.661 & 8 \\
& 40\% & 0.683 & 0.65 & 0.711 & 0.715 & 0.743 & 0.706 & 0.72 & 0.62 & 0.643 & 0.578 & 0.673 & 0.635 & 0.593 & 0.591 & 0.664 & 7 \\
PAIRS\_FSG\_SM& 50\% & 0.688 & 0.636 & 0.739 & 0.654 & 0.694 & 0.751 & 0.758 & 0.59 & 0.64 & 0.57 & 0.618 & 0.609 & 0.585 & 0.566 & 0.69 & 10 \\
& 60\% & 0.685 & 0.657 & 0.768 & 0.676 & 0.705 & 0.781 & 0.796 & 0.567 & 0.641 & 0.548 & 0.601 & 0.605 & 0.563 & 0.59 & 0.591 & 4 \\
& 70\% & 0.664 & 0.638 & 0.815 & 0.688 & 0.705 & 0.837 & 0.842 & 0.516 & 0.63 & 0.525 & 0.589 & 0.62 & 0.511 & 0.576 & 0.621 & 6 \\
& 80\% & 0.725 & 0.636 & 0.891 & 0.728 & 0.743 & 0.906 & 0.898 & 0.751 & 0.618 & 0.489 & 0.618 & 0.588 & 0.491 & 0.537 & 0.66 & 7 \\
& 90\% & 0.721 & 0.665 & 0.975 & 0.853 & 0.772 & 0.99 & 0.995 & 0.432 & 0.589 & 0.437 & 0.589 & 0.553 & 0.411 & 0.457 & 0.669 & 8 \\\hline
& 10\% & 0.865 & 0.796 & 0.825 & 0.869 & 0.845 & 0.873 & 0.948 & 0.84 & 1 & 0.889 & 0.93 & 0.953 & 0.952 & 0.848 & 0.863 & 5 \\
& 20\% & 0.83 & 0.746 & 0.795 & 0.812 & 0.822 & 0.84 & 0.868 & 0.817 & 1 & 0.884 & 0.903 & 0.95 & 0.92 & 0.822 & 0.803 & 2 \\
& 30\% & 0.819 & 0.737 & 0.776 & 0.737 & 0.795 & 0.832 & 0.845 & 0.806 & 1 & 0.873 & 0.895 & 0.946 & 0.903 & 0.8 & 0.775 & 2 \\
& 40\% & 0.812 & 0.729 & 0.777 & 0.748 & 0.804 & 0.831 & 0.835 & 0.791 & 0.992 & 0.863 & 0.896 & 0.946 & 0.886 & 0.764 & 0.743 & 1 \\
YEAST & 50\% & 0.812 & 0.737 & 0.776 & 0.775 & 0.767 & 0.814 & 0.843 & 0.755 & 0.935 & 0.852 & 0.889 & 0.931 & 0.815 & 0.734 & 0.696 & 0 \\
& 60\% & 0.797 & 0.711 & 0.775 & 0.814 & 0.773 & 0.825 & 0.822 & 0.73 & 0.947 & 0.866 & 0.92 & 0.922 & 0.805 & 0.735 & 0.74 & 3 \\
& 70\% & 0.815 & 0.748 & 0.812 & 0.811 & 0.8 & 0.835 & 0.808 & 0.686 & 0.866 & 0.849 & 0.889 & 0.921 & 0.766 & 0.735 & 0.678 & 0 \\
& 80\% & 0.837 & 0.789 & 0.803 & 0.806 & 0.795 & 0.837 & 0.82 & 0.651 & 0.896 & 0.795 & 0.765 & 0.915 & 0.738 & 0.774 & 0.663 & 1 \\
& 90\% & 0.814 & 0.764 & 0.781 & 0.776 & 0.76 & 0.886 & 0.776 & 0.65 & 0.869 & 0.709 & 0.684 & 0.916 & 0.667 & 0.823 & 0.671 & 2 \\\hline
& 10\% & 0.998 & 0.999 & 0.999 & 0.998 & 0.997 & 0.999 & 1 & 0.999 & 1 & 0.999 & 0.998 & 0.998 & 0.998 & 0.991 & 0.997 & 2 \\
& 20\% & 0.972 & 0.998 & 0.999 & 0.98 & 0.979 & 0.982 & 1 & 0.999 & 1 & 0.994 & 0.98 & 0.98 & 0.98 & 0.968 & 0.996 & 9 \\
& 30\% & 0.945 & 0.984 & 0.976 & 0.957 & 0.956 & 0.948 & 1 & 0.999 & 0.999 & 0.993 & 0.956 & 0.957 & 0.956 & 0.946 & 0.992 & 10 \\
& 40\% & 0.933 & 0.95 & 0.953 & 0.931 & 0.929 & 0.929 & 1 & 0.998 & 0.997 & 0.978 & 0.929 & 0.931 & 0.933 & 0.939 & 0.987 & 11 \\
CS\_PHD& 50\% & 0.94 & 0.938 & 0.947 & 0.94 & 0.938 & 0.928 & 1 & 0.998 & 0.994 & 0.977 & 0.938 & 0.94 & 0.942 & 0.906 & 0.98 & 11 \\
& 60\% & 0.925 & 0.932 & 0.946 & 0.941 & 0.939 & 0.941 & 1 & 0.998 & 0.986 & 0.979 & 0.939 & 0.941 & 0.939 & 0.911 & 0.976 & 10 \\
& 70\% & 0.928 & 0.953 & 0.937 & 0.928 & 0.925 & 0.925 & 1 & 1 & 0.966 & 0.975 & 0.928 & 0.928 & 0.928 & 0.89 & 0.967 & 11 \\
& 80\% & 0.911 & 0.948 & 0.939 & 0.93 & 0.925 & 0.93 & 1 & 0.991 & 0.897 & 0.972 & 0.93 & 0.93 & 0.948 & 0.883 & 0.965 & 11 \\
& 90\% & 0.935 & 0.953 & 0.925 & 0.916 & 0.916 & 0.916 & 1 & 0.953 & 0.897 & 0.944 & 0.916 & 0.916 & 0.944 & 0.832 & 0.958 & 13 
\end{tabular}
\end{tiny}
\end{table*}

\begin{table*}[h]
\centering
\caption{Classification error in active inference based on 'measure'-neighbour version of distinct selection strategy with LBP. Note: the last column represents how many of the non--random selection strategies outperformed the random case.\label{tab:lbp_neighbour_results}}
\begin{tiny}
\begin{tabular}{p{1.2cm}p{0.5cm}p{0.5cm}p{0.5cm}p{0.5cm}p{0.5cm}p{0.5cm}p{0.5cm}p{0.5cm}p{0.5cm}p{0.5cm}p{0.5cm}p{0.5cm}p{0.5cm}p{0.5cm}p{0.5cm}p{0.5cm}p{0.5cm}}
Dataset & Lab. nodes & Top indegree & Top outdegree & Top betweeness & Top clust. coeff. & Top hubness & Top authority & Top page rank & Down indegree & Down outdegree & Down betweeness & Down clust. coeff. & Down hubness & Down authority & Down page rank & Random & \# better\\\hline

& 10\% & 0.789 & 0.827 & 0.841 & 0.845 & 0.775 & 0.81 & 0.81 & 0.837 & 0.814 & 0.839 & 0.821 & 0.83 & 0.844 & 0.818 & 0.863 & 14 \\
& 20\% & 0.735 & 0.736 & 0.729 & 0.731 & 0.744 & 0.734 & 0.703 & 0.76 & 0.7 & 0.737 & 0.742 & 0.764 & 0.746 & 0.776 & 0.817 & 14 \\
& 30\% & 0.676 & 0.648 & 0.679 & 0.68 & 0.657 & 0.675 & 0.656 & 0.663 & 0.671 & 0.674 & 0.64 & 0.686 & 0.644 & 0.696 & 0.793 & 14 \\
& 40\% & 0.631 & 0.623 & 0.595 & 0.589 & 0.569 & 0.578 & 0.607 & 0.658 & 0.585 & 0.56 & 0.582 & 0.624 & 0.633 & 0.611 & 0.77 & 14 \\
AMD & 50\% & 0.557 & 0.549 & 0.541 & 0.608 & 0.566 & 0.532 & 0.603 & 0.579 & 0.564 & 0.58 & 0.528 & 0.549 & 0.492 & 0.545 & 0.729 & 14 \\
& 60\% & 0.503 & 0.492 & 0.522 & 0.528 & 0.455 & 0.454 & 0.528 & 0.536 & 0.536 & 0.542 & 0.5 & 0.495 & 0.508 & 0.462 & 0.714 & 14 \\
& 70\% & 0.432 & 0.439 & 0.413 & 0.447 & 0.443 & 0.442 & 0.457 & 0.454 & 0.494 & 0.523 & 0.484 & 0.457 & 0.464 & 0.455 & 0.692 & 14 \\
& 80\% & 0.418 & 0.403 & 0.386 & 0.475 & 0.436 & 0.401 & 0.447 & 0.424 & 0.411 & 0.458 & 0.361 & 0.394 & 0.399 & 0.473 & 0.681 & 14 \\
& 90\% & 0.402 & 0.328 & 0.299 & 0.393 & 0.372 & 0.323 & 0.394 & 0.412 & 0.38 & 0.373 & 0.295 & 0.314 & 0.413 & 0.345 & 0.651 & 14 \\\hline
& 10\% & 0.971 & 0.92 & 0.926 & 0.923 & 0.969 & 0.976 & 0.962 & 0.923 & 0.964 & 0.972 & 0.969 & 1 & 0.999 & 0.902 & 0.959 & 5 \\
& 20\% & 0.903 & 0.866 & 0.918 & 0.848 & 0.951 & 0.948 & 0.934 & 0.878 & 0.915 & 0.931 & 0.949 & 0.999 & 0.999 & 0.818 & 0.929 & 7 \\
& 30\% & 0.863 & 0.839 & 0.897 & 0.787 & 0.919 & 0.929 & 0.903 & 0.805 & 0.858 & 0.896 & 0.923 & 0.998 & 0.997 & 0.74 & 0.904 & 9 \\
& 40\% & 0.831 & 0.772 & 0.878 & 0.738 & 0.907 & 0.902 & 0.893 & 0.792 & 0.847 & 0.859 & 0.88 & 0.996 & 0.997 & 0.699 & 0.883 & 9 \\
NET\_SCIENCE& 50\% & 0.786 & 0.745 & 0.846 & 0.684 & 0.872 & 0.873 & 0.856 & 0.73 & 0.813 & 0.797 & 0.864 & 0.996 & 0.971 & 0.645 & 0.861 & 9 \\
& 60\% & 0.776 & 0.703 & 0.831 & 0.647 & 0.841 & 0.831 & 0.821 & 0.699 & 0.75 & 0.785 & 0.835 & 0.993 & 0.937 & 0.59 & 0.848 & 12 \\
& 70\% & 0.72 & 0.693 & 0.786 & 0.636 & 0.782 & 0.786 & 0.797 & 0.661 & 0.731 & 0.739 & 0.829 & 0.992 & 0.916 & 0.612 & 0.84 & 12 \\
& 80\% & 0.702 & 0.676 & 0.729 & 0.641 & 0.752 & 0.742 & 0.767 & 0.668 & 0.693 & 0.653 & 0.793 & 0.996 & 0.867 & 0.591 & 0.827 & 12 \\
& 90\% & 0.676 & 0.645 & 0.71 & 0.644 & 0.715 & 0.712 & 0.732 & 0.635 & 0.688 & 0.639 & 0.752 & 0.987 & 0.814 & 0.554 & 0.813 & 12 \\\hline
& 10\% & 0.916 & 0.913 & 0.892 & 0.902 & 0.915 & 0.888 & 0.89 & 0.876 & 1 & 0.916 & 0.927 & 0.906 & 0.922 & 0.864 & 0.908 & 7 \\
& 20\% & 0.877 & 0.839 & 0.84 & 0.909 & 0.858 & 0.895 & 0.906 & 0.861 & 0.911 & 0.873 & 0.954 & 0.917 & 0.913 & 0.797 & 0.86 & 4 \\
& 30\% & 0.854 & 0.835 & 0.791 & 0.902 & 0.809 & 0.898 & 0.864 & 0.792 & 0.859 & 0.882 & 0.943 & 0.907 & 0.915 & 0.821 & 0.844 & 5 \\
& 40\% & 0.797 & 0.834 & 0.812 & 0.91 & 0.75 & 0.892 & 0.798 & 0.795 & 0.858 & 0.851 & 0.954 & 0.917 & 0.916 & 0.757 & 0.823 & 6 \\
PAIRS\_FSG& 50\% & 0.239 & 0.199 & 0.23 & 0.227 & 0.225 & 0.234 & 0.236 & 0.201 & 0.22 & 0.188 & 0.207 & 0.203 & 0.332 & 0.18 & 0.245 & 13 \\
& 60\% & 0.222 & 0.188 & 0.216 & 0.214 & 0.214 & 0.231 & 0.231 & 0.173 & 0.2 & 0.182 & 0.183 & 0.193 & 0.171 & 0.178 & 0.234 & 14 \\
& 70\% & 0.214 & 0.191 & 0.203 & 0.195 & 0.2 & 0.219 & 0.232 & 0.163 & 0.195 & 0.164 & 0.177 & 0.192 & 0.161 & 0.161 & 0.223 & 13 \\
& 80\% & 0.205 & 0.169 & 0.206 & 0.175 & 0.183 & 0.216 & 0.209 & 0.14 & 0.183 & 0.15 & 0.169 & 0.173 & 0.151 & 0.154 & 0.213 & 13 \\
& 90\% & 0.204 & 0.172 & 0.21 & 0.169 & 0.171 & 0.204 & 0.2 & 0.135 & 0.168 & 0.137 & 0.168 & 0.158 & 0.134 & 0.133 & 0.208 & 13 \\\hline
& 10\% & 0.616 & 0.605 & 0.633 & 0.615 & 0.639 & 0.634 & 0.632 & 0.602 & 0.614 & 0.708 & 0.621 & 0.736 & 0.739 & 0.591 & 0.666 & 11 \\
& 20\% & 0.738 & 0.559 & 0.596 & 0.587 & 0.583 & 0.614 & 0.622 & 0.745 & 0.544 & 0.537 & 0.739 & 0.722 & 0.725 & 0.522 & 0.639 & 9 \\
& 30\% & 0.547 & 0.53 & 0.569 & 0.566 & 0.556 & 0.575 & 0.734 & 0.516 & 0.521 & 0.694 & 0.74 & 0.699 & 0.688 & 0.49 & 0.661 & 9 \\
& 40\% & 0.519 & 0.724 & 0.54 & 0.531 & 0.548 & 0.733 & 0.581 & 0.717 & 0.715 & 0.675 & 0.705 & 0.485 & 0.666 & 0.717 & 0.664 & 6 \\
PAIRS\_FSG\_SM& 50\% & 0.73 & 0.484 & 0.536 & 0.498 & 0.531 & 0.748 & 0.555 & 0.704 & 0.715 & 0.722 & 0.697 & 0.695 & 0.432 & 0.667 & 0.69 & 7 \\
& 60\% & 0.494 & 0.706 & 0.521 & 0.485 & 0.521 & 0.742 & 0.74 & 0.677 & 0.708 & 0.373 & 0.731 & 0.737 & 0.385 & 0.4 & 0.591 & 7 \\
& 70\% & 0.728 & 0.732 & 0.491 & 0.469 & 0.488 & 0.75 & 0.731 & 0.416 & 0.711 & 0.388 & 0.74 & 0.693 & 0.677 & 0.714 & 0.621 & 5 \\
& 80\% & 0.726 & 0.726 & 0.742 & 0.433 & 0.471 & 0.516 & 0.755 & 0.705 & 0.704 & 0.364 & 0.654 & 0.679 & 0.729 & 0.339 & 0.66 & 6 \\
& 90\% & 0.705 & 0.74 & 0.699 & 0.712 & 0.456 & 0.733 & 0.737 & 0.72 & 0.693 & 0.657 & 0.682 & 0.71 & 0.647 & 0.664 & 0.669 & 4 \\\hline
& 10\% & 0.775 & 0.758 & 0.777 & 0.817 & 0.77 & 0.822 & 0.817 & 0.781 & 1 & 0.826 & 0.873 & 0.865 & 0.841 & 0.833 & 0.863 & 11 \\
& 20\% & 0.7 & 0.745 & 0.699 & 0.714 & 0.713 & 0.728 & 0.74 & 0.72 & 1 & 0.837 & 0.854 & 0.883 & 0.856 & 0.815 & 0.803 & 8 \\
& 30\% & 0.807 & 0.639 & 0.631 & 0.656 & 0.68 & 0.856 & 0.671 & 0.744 & 1 & 0.761 & 0.748 & 0.869 & 0.697 & 0.751 & 0.775 & 10 \\
& 40\% & 0.614 & 0.577 & 0.564 & 0.646 & 0.604 & 0.634 & 0.815 & 0.745 & 0.946 & 0.812 & 0.819 & 0.764 & 0.809 & 0.666 & 0.743 & 7 \\
YEAST & 50\% & 0.758 & 0.535 & 0.55 & 0.619 & 0.554 & 0.611 & 0.627 & 0.646 & 0.815 & 0.77 & 0.774 & 0.718 & 0.77 & 0.668 & 0.696 & 8 \\
& 60\% & 0.689 & 0.511 & 0.562 & 0.758 & 0.753 & 0.563 & 0.587 & 0.632 & 0.69 & 0.662 & 0.747 & 0.755 & 0.607 & 0.594 & 0.74 & 10 \\
& 70\% & 0.678 & 0.701 & 0.664 & 0.57 & 0.522 & 0.777 & 0.552 & 0.572 & 0.729 & 0.606 & 0.595 & 0.593 & 0.576 & 0.544 & 0.678 & 11 \\
& 80\% & 0.685 & 0.703 & 0.519 & 0.74 & 0.528 & 0.688 & 0.53 & 0.674 & 0.697 & 0.579 & 0.734 & 0.684 & 0.496 & 0.531 & 0.663 & 6 \\
& 90\% & 0.702 & 0.496 & 0.526 & 0.567 & 0.714 & 0.515 & 0.694 & 0.633 & 0.519 & 0.701 & 0.509 & 0.721 & 0.476 & 0.496 & 0.671 & 9 \\\hline
& 10\% & 1 & 1 & 0.999 & 1 & 1 & 1 & 1 & 1 & 1 & 1 & 1 & 0.995 & 0.995 & 0.994 & 0.997 & 3 \\
& 20\% & 0.999 & 0.999 & 0.999 & 0.999 & 0.994 & 1 & 1 & 1 & 0.998 & 0.999 & 0.986 & 0.995 & 0.996 & 0.998 & 0.996 & 4 \\
& 30\% & 0.997 & 0.99 & 0.999 & 0.998 & 0.997 & 0.998 & 1 & 0.998 & 1 & 0.998 & 0.996 & 0.989 & 0.994 & 0.99 & 0.992 & 3 \\
& 40\% & 0.98 & 0.999 & 0.991 & 0.986 & 0.986 & 0.99 & 1 & 1 & 0.997 & 0.998 & 0.975 & 0.978 & 0.984 & 0.992 & 0.987 & 6 \\
CS\_PHD& 50\% & 0.996 & 0.993 & 0.982 & 0.978 & 0.981 & 0.983 & 0.998 & 0.999 & 0.997 & 0.999 & 0.982 & 0.99 & 0.983 & 0.988 & 0.98 & 1 \\
& 60\% & 0.981 & 0.987 & 0.995 & 0.993 & 0.991 & 0.984 & 0.997 & 0.994 & 0.986 & 0.994 & 0.98 & 0.977 & 0.975 & 0.99 & 0.976 & 1 \\
& 70\% & 0.987 & 0.982 & 0.954 & 0.992 & 0.997 & 0.973 & 0.999 & 0.999 & 0.995 & 0.992 & 0.968 & 0.973 & 0.976 & 0.979 & 0.967 & 1 \\
& 80\% & 0.992 & 0.986 & 0.979 & 0.992 & 0.978 & 0.988 & 1 & 1 & 0.99 & 0.997 & 0.962 & 0.978 & 0.974 & 0.972 & 0.965 & 1 \\
& 90\% & 0.988 & 0.988 & 0.986 & 0.994 & 0.98 & 0.981 & 0.994 & 0.981 & 0.979 & 0.98 & 0.982 & 0.98 & 0.984 & 0.967 & 0.958 & 0 
\end{tabular}
\end{tiny}
\end{table*}

\begin{table*}[t]
\caption{The comparison showing how often top or bottom of ranks for given measures outperformed each other; all datasets merged. The number indicated how many times results of a particular method (top or bottom) was better than another one. It happened in some cases that both methods were equal and provided the same error level. Overall, there were 72 comparisons.}

\label{tab:topbottom}
\begin{tabular}{|c|c|c|c|c|c|c|c|c|c|c|c|c|c|c|}
\hline
& \multicolumn{2}{c|}{\textbf{In-degree}} & \multicolumn{2}{c|}{\textbf{Out-degree}} & \multicolumn{2}{c|}{\textbf{Betweenness}} & \multicolumn{2}{c|}{\textbf{Clustering coeff.}} & \multicolumn{2}{c|}{\textbf{Hub centrality}} & \multicolumn{2}{c|}{\textbf{Authority}} & \multicolumn{2}{c|}{\textbf{PageRank}}\\\hline
& top & bottom & top & bottom & top & bottom & top & bottom & top & bottom & top & bottom & top & bottom\\\hline
\textbf{LBP} & 19 & \textbf{53} & \textbf{52} & 18 & 32 & \textbf{39} & \textbf{30} & 28 & \textbf{35} & 27 & \textbf{31} & 30 & 13 & \textbf{59} \\ \hline 
\textbf{LBP-neighbour} & 27 & \textbf{44} & \textbf{56} & 15 & \textbf{38} & 34 & \textbf{42} & 29 & \textbf{52} & 20 & \textbf{36} & \textbf{36} & 20 & \textbf{52} \\ \hline 
\textbf{ICA} & \textbf{48} & 24 & \textbf{37} & 33 & \textbf{40} & 31 & \textbf{36} & 31 & \textbf{37} & 33 & \textbf{48} & 23 & \textbf{52} & 17 \\ \hline 
\textbf{ICA-neighbour} & \textbf{38} & 34 & \textbf{40} & 32 & 33 & \textbf{39} & 35 & \textbf{37} & \textbf{41} & 31 & \textbf{41} & 31 & 28 & \textbf{44} \\ \hline 
\end{tabular}
\end{table*}

\section{Experimental results and discussion\label{sec:discussion}}

As the experiments were performed using large number of parameters, the obtained results can be analysed from many different perspectives. Overall, it can be noticed that accuracy  of the investigated approaches varies and the methods themselves cannot be compared in a straightforward way. This is clearly visible in Figure \ref{fig:best_results} depicting results obtained for the best combination of ICA, ICA neighbour,  LBP, and LBP neighbour settings and compared with random selection. All of the results are presented in Tables \ref{tab:ica_results}, \ref{tab:ica_neighbour_results}, \ref{tab:lbp_results} and \ref{tab:lbp_neighbour_results}. 

There are three basic factors that influence the output. The first one is the structural profile of the network, the second is the method of selecting nodes for the initial label acquisition (seed selection strategy), and the third one is the method of within--network collective classification. Also, the percentage of uncovered classes during the initial selection process contributes to the classification accuracy. 

In general, there is no single node selection method combined with inference concepts that would be best for every kind of network and every size of initial node set. However, some approaches and combinations of the reasoning algorithms with seed acquisition methods are better than others for particular network profiles, see Section \ref{sec:resultsForNetworkGroups}.
This would suggest that the results depends on the network profile. 

One of the observations that can be derived from the experimental results is the better performance of node selection methods based on 'measure'--neighbour approach described in Section \ref{sec:newUtilityScores} compared with the original rankings, especially for datasets in groups 1 and 2. This suggests that the bigger the clustering coefficient of the network, i.e. the higher probability that the clusters will exist in the network, the better the classification results. It is visible, if we juxtapose results for individual measures from Table \ref{tab:ica_results} with Table \ref{tab:ica_neighbour_results} and from Table \ref{tab:lbp_results} with Table \ref{tab:lbp_neighbour_results}. The comparison showing how often top or bottom of ranks for given measures outperformed each other is presented in Table~\ref{tab:topbottom}. As a result, 'measure'--neighbour methods more often surpass random selection than the measure based methods, see the last column in Table  \ref{tab:ica_neighbour_results} and \ref{tab:lbp_neighbour_results}. Moreover, while analysing the last column in Tables \ref{tab:ica_results},  \ref{tab:ica_neighbour_results},  \ref{tab:lbp_results}, and   \ref{tab:lbp_neighbour_results}, we can find out that there is always at least one ICA and at least one LBP node selection method outperforming the random approach. In some cases, e.g. for AMD and PAIRS\_FSG (Tables \ref{tab:ica_neighbour_results} and \ref{tab:lbp_neighbour_results}), all or almost all 'measure'--neighbour methods are better than random selection. The proof for existence of some methods better than random in any case is very important from practical point of view. It justifies that searching for more effective inference methods can always be successful. 

Experimental results also revealed that, regardless of the kind of nodes' selection method in active learning (degree, betweenness, etc.) and selection strategy (top or bottom of the ranking), none of the methods was able to satisfactory generalize networks that are very sparse and random by nature (especially for network CS\_PHD which belongs to group 3, see Section~\ref{sec_datasets}). For such problem, the results were quite similar to random seeding, see Tables \ref{tab:ica_results}, \ref{tab:ica_neighbour_results}, \ref{tab:lbp_results} and \ref{tab:lbp_neighbour_results}.

When analysing the active learning method giving the best results for ICA, in most cases, the inference results were not as much susceptible to the percentage of known nodes as for the LBP method. It means that the global network propagation of information applied by LBP is more dependent on the size of the training set than ICA method. In general, acquisition of labels by means of the best proposed methods (e.g. nodes with the greatest degree or neighbours of nodes with the lowest page rank) in conjunction with ICA and LBP algorithms, in most cases, outperformed random results. However, when using LBP the results might suffer from its basic property: if the selection method does not provide nodes from all connected components then the information about labels is not propagated to separated parts of the network.

In addition, as it was described in Section \ref{sec:Algorithms}, the Loopy Belief Propagation (LBP) method is heavily network dependent, because the underlying network structure and global objective function determine the propagation, while the ICA method utilizes only the local  structure of the network. The experimental results confirm that these differences heavily influenced the results of individual classification methods. For example compare results for AMD and NET\_SCIENCE (good performance of LBP for small--world like networks) with CS\_PHD (the poor LBP results for the loosely connected network with small node degree).        
\subsection{Influence of network characteristics on classification results\label{sec:resultsForNetworkGroups}}

To facilitate the analysis of the large number of experimental  results, for each group of networks presented in Section~\ref{sec_datasets}, the appropriate 'results profile' has been created (see Table~\ref{tab:datasets_groups}). Classification performed on the networks from group 1, which can be characterised as networks that exhibit small--world phenomenon, features good accuracy of 'measure'--neighbour methods with LBP neighbourhood (bottom page rank) outperforming other approaches. In addition, most of the 'measure'--neighbour methods outperform random case. Classification error, which for this group is high for small training sets, substantially decreases for the bigger training sets. Classification results for the second group of networks exhibit relatively small variance of error. In general, the more classes in the dataset, the worse classification accuracy. Moreover, for smaller modularity and density as well as greater clustering coefficient LBP neighbour approach outperforms others.

Finally, for the third group of networks, which are close to random networks with a very low connectivity probability, 'measure'--neighbour methods are worse than original and random approaches. The classification results are rather poor for all cases, but due to the fact that ICA method does not depend on connections within network for classification, it outperforms LBP-based methods.

\begin{figure}
               \centering
                \includegraphics[width=0.93\textwidth]{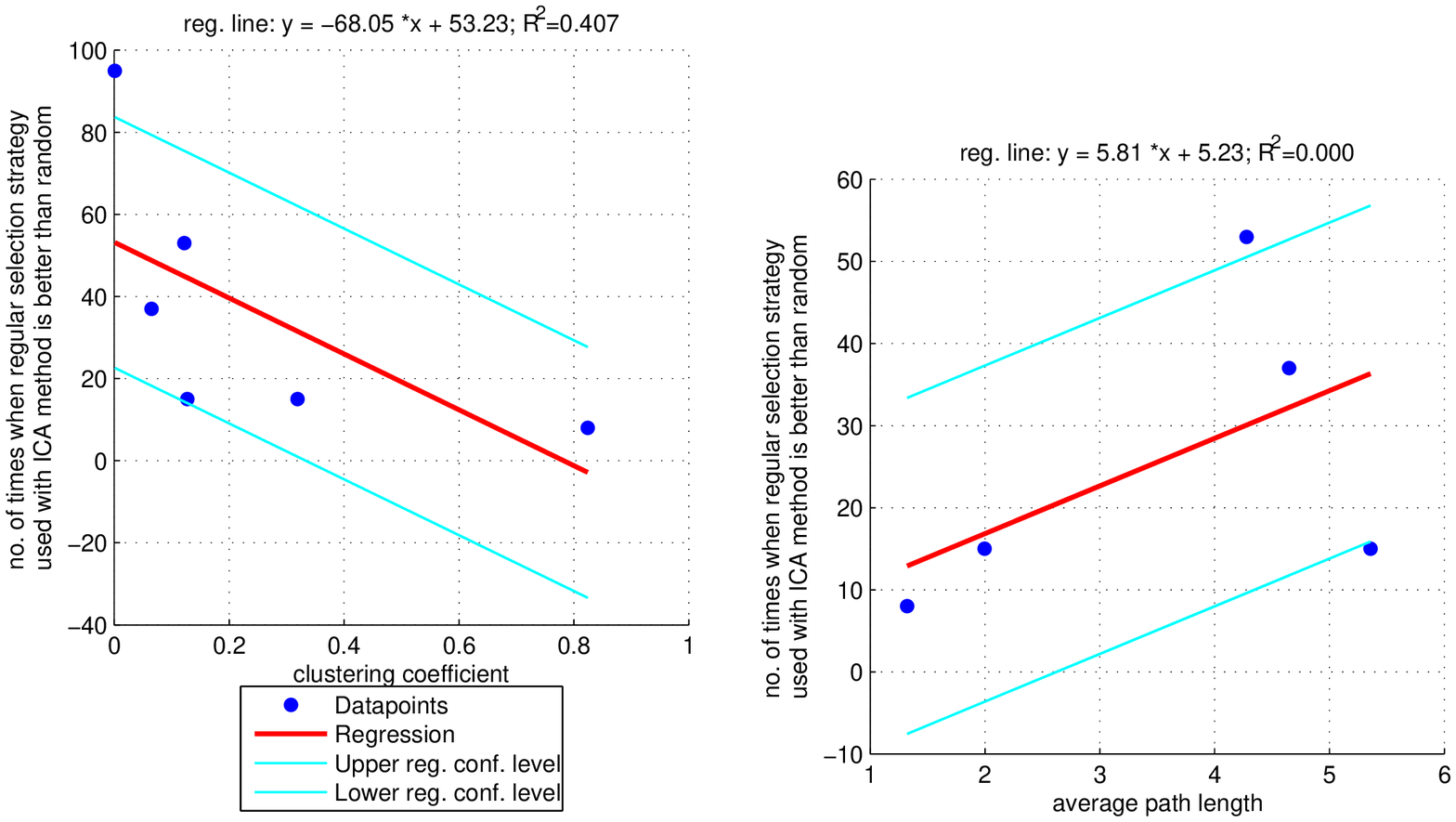}
                \caption{Robust fit regression between \textbf{clustering coefficient} (left plot) /\textbf{ average path length} (right plot) and the number of times when ICA method with regular selection strategy  is better than                 random one (sum of the last column from Table \ref{tab:ica_results} for each network). Plot on the right does not include CS\_PHD network as the network is not connected so average path length is not informative..\                 \label{fig:ICA_reg}}

\end{figure}

\begin{figure}
      \centering
      \includegraphics[width=0.93\textwidth]{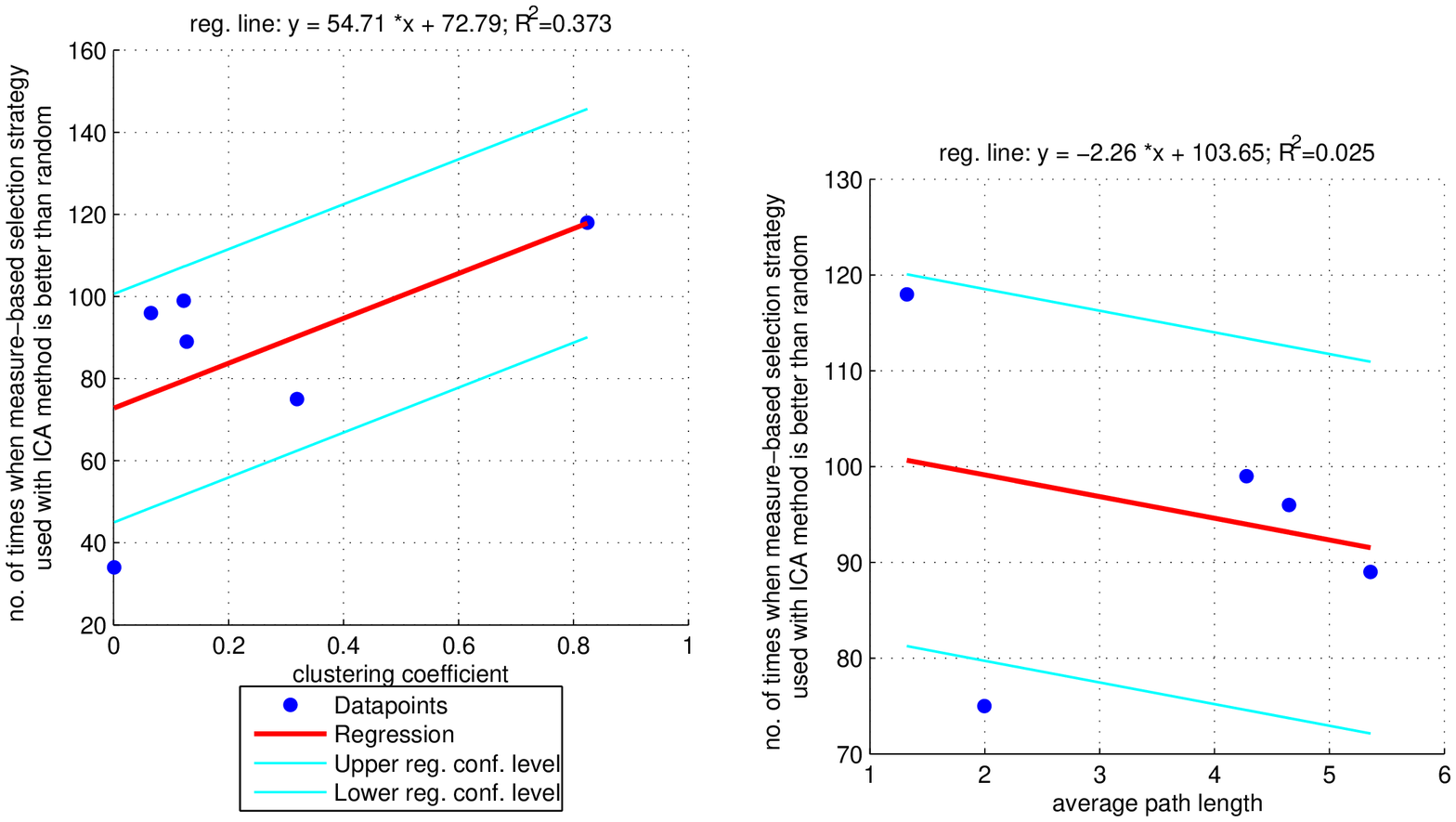}
      \caption{Regression between \textbf{clustering coefficient}                 (left plot) / \textbf{average path length} (right plot) and the number of times when ICA approach with 'measure'--neighbour selection strategies is better than                 random one (sum of the last column from Table \ref{tab:ica_neighbour_results} for each network). Right plot does not include CS\_PHD network as the network is not connected so average path length is not informative.\                 \label{fig:ICAn_reg}}
        
\end{figure}

\begin{figure}
               \centering
                \includegraphics[width=0.93\textwidth]{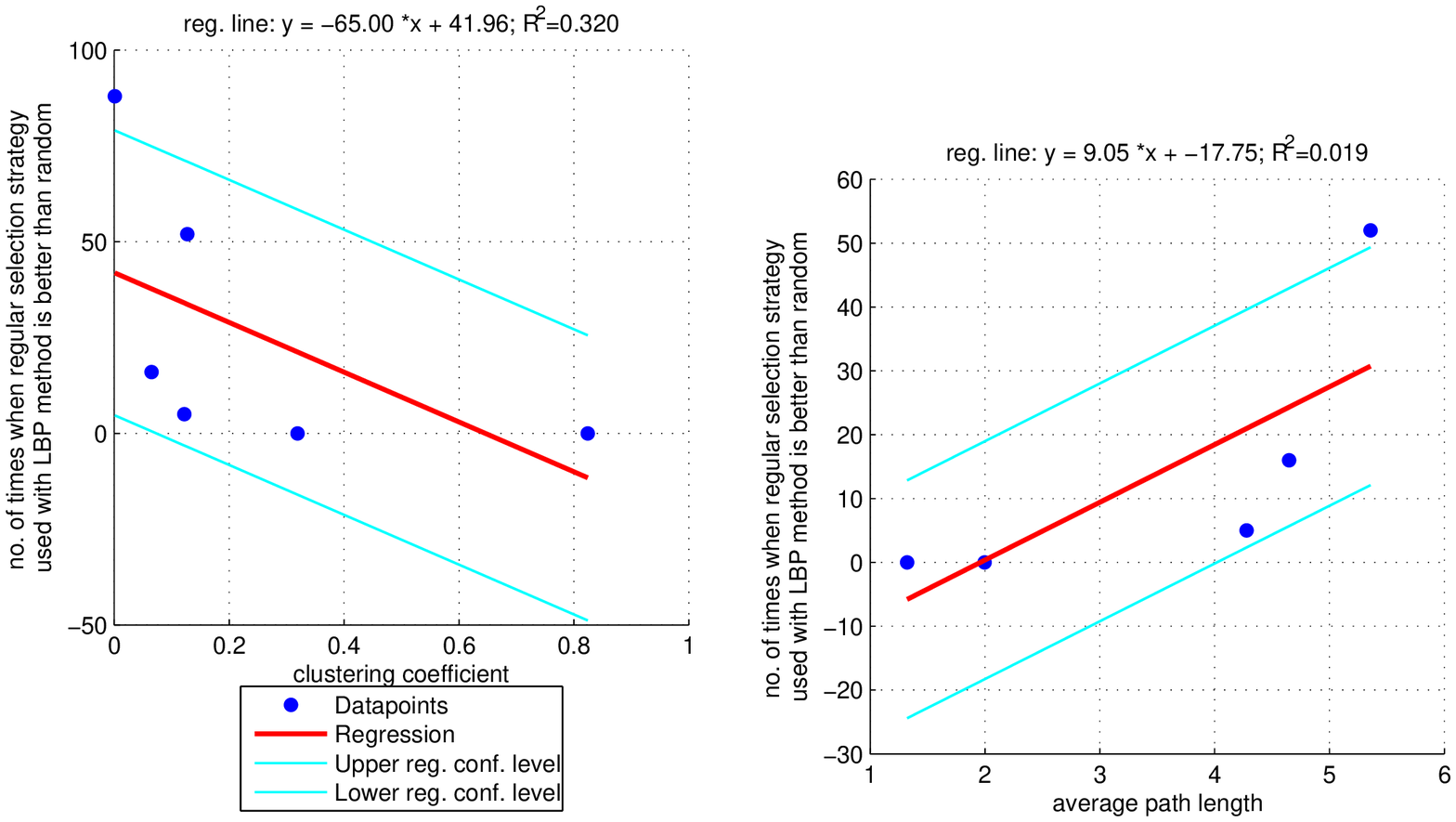}
                \caption{Regression between \textbf{clustering coefficient} (left plot) /\textbf{ average path length} (right plot) and the number of times when LBP method  with regular selection strategy is better than                 random one (sum of the last column from Table \ref{tab:lbp_results} for each network). Plot on the right does not include CS\_PHD network as the network is not connected so average path length is not informative..\                 \label{fig:LBP_reg}}

\end{figure}

\begin{figure}
      \centering
      \includegraphics[width=0.93\textwidth]{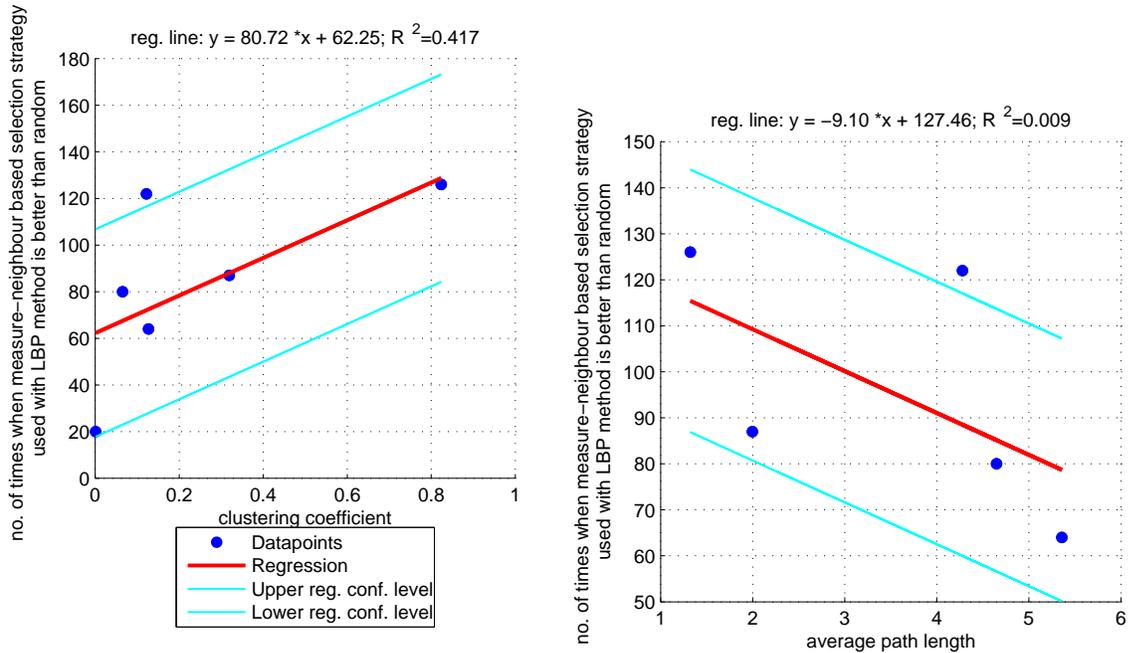}
      \caption{Regression between \textbf{clustering coefficient}                 (left plot) / \textbf{average path length} (right plot) and the number of times when LBP approach with 'measure'--neighbour selection strategies is better than                 random one (sum of the last column from Table \ref{tab:lbp_neighbour_results} for each network). Right plot  does not include CS\_PHD network as the network is not connected so average path length is not informative.\                 \label{fig:LBPn_reg}}
        
\end{figure}

Additionally, we performed a set of robust fit regressions in order to investigate the relation between structural properties and the results of the classifications (see Figures~\ref{fig:ICA_reg},~\ref{fig:ICAn_reg},~\ref{fig:LBP_reg}, and~\ref{fig:LBPn_reg}). For this part of the study by results we understand the number of times when a given approach (ICA measure based, ICA measure-neighbour based, LBP measure based, LBP measure-neighbour based) was better than random for each analysed network (sum of the last column from Tables~\ref{tab:ica_results},~\ref{tab:ica_neighbour_results},~\ref{tab:lbp_results},~\ref{tab:lbp_neighbour_results} for each network).  We took into account two metrics: (i) clustering coefficient and (ii) average path length as those are the measures that enable to classify networks as random, small--world or ordered ones.
Results show that for measure--based methods (for both ICA and LBP approaches) the smaller the clustering coefficient and the bigger the average path length the better the performance of the classification. Exactly opposite trend is visible for measure-neighbour approaches. However, we should rather neglect the results for average path length as in all cases $R^2$ is very close to 0. Concentrating on clustering coefficient metric and the obtained results  we can make a recommendation that for networks that are disconnected and very random in their nature (Group~3) we should use measure--based   selection strategies and for networks that are connected (Group 1 and 2) we should rather apply 'measure'--neighbour selection strategies.
\onecolumn
\twocolumn

\onecolumn
\twocolumn

\subsection{Representativeness of the selected training sets} 
 
While considering the relational classification results, it should be investigated to what extend the selected nodes used for training and propagation appropriately represent the whole network, especially in terms of class conditional distribution. 

\begin{figure}
\centering
\includegraphics[width=1\columnwidth]{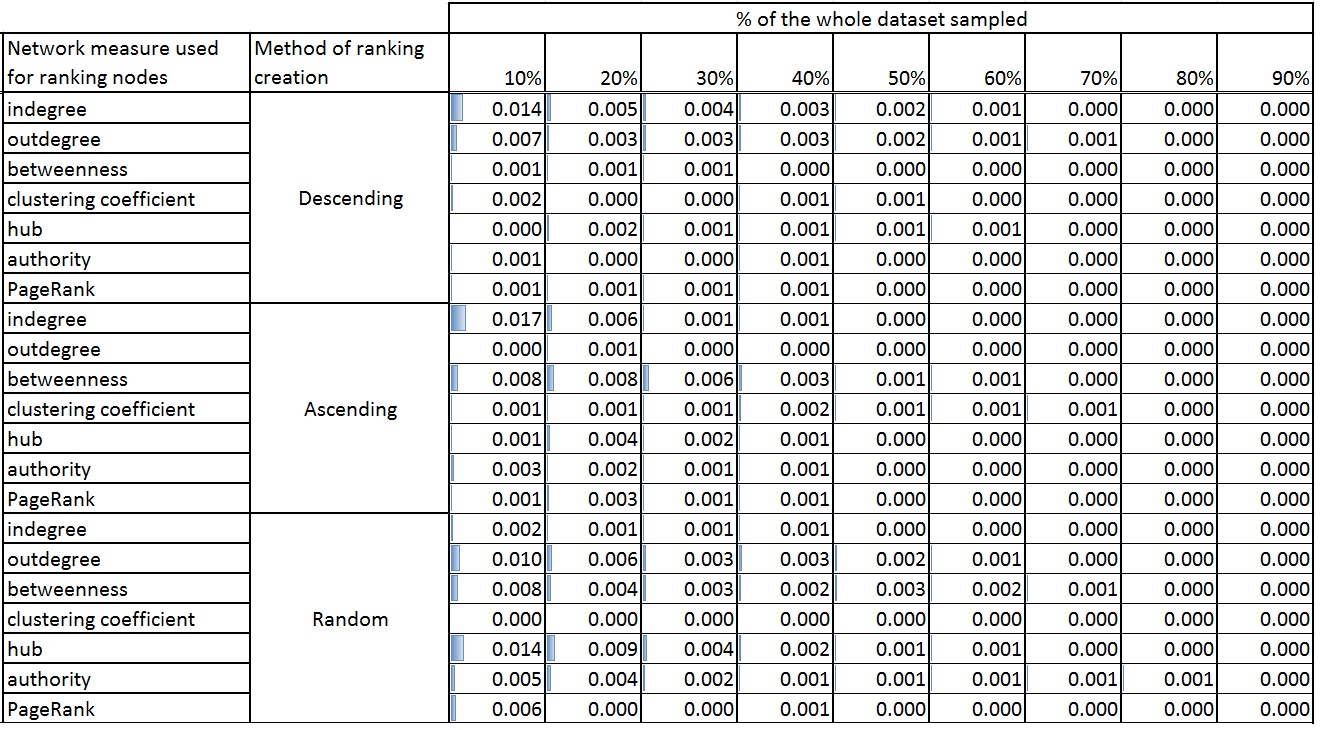}
\caption{Kullback--Leibler divergence for the PAIRS\_FSG network
\label{fig:KL_pairs}}
\end{figure}

\begin{figure}
\centering
\includegraphics[width=1\columnwidth]{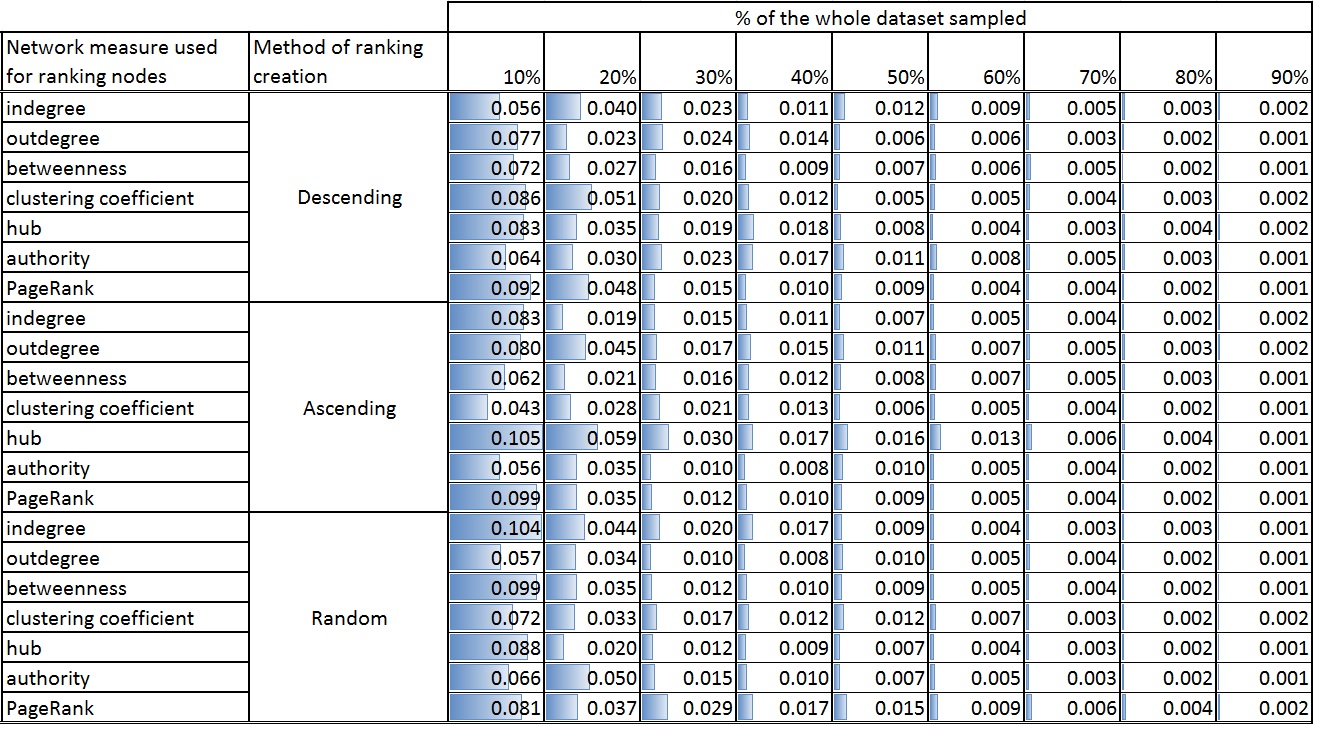}
\caption{Kullback--Leibler divergence for the NET\_SCIENCE network
\label{fig:KL_netscience}}
\end{figure}

In order to assess the representativeness of selected training set, the standard Kullback--Leibler divergence (a.k.a. relative entropy) was used  which is a measure of the difference between two probability distributions. It measures how much information is lost when one probability distribution (in our case it is a distribution of classes in a given sample -- 10\%, ..., 90\% of the whole dataset) is used to approximate another one; in here it is the probability distribution of classes in the whole dataset. The smaller the divergence, the smaller loss; 0 means that no information is lost and that both distributions are the same.
Results for all datasets are presented in the Supplement Materials in Section 3. Here, we present only the calculations for two networks: (i) PAIRS\_FSG with 3 classes  (Figure~\ref{fig:KL_pairs}) and (ii) NET\_SCIENCE with 26 classes (Figure~\ref{fig:KL_netscience}). The former has the smallest number of classes and the latter the largest number of classes out of all tested datasets.

Both Figures~\ref{fig:KL_pairs} and~\ref{fig:KL_netscience} show the Kullback--Leibler divergence for the selected networks and different structural measures used to create the rankings of nodes and for three methods of ranking ordering: descending, ascending and random. Although, the divergence is generally bigger for NET\_SCIENCE network than PAIRS, the absolute values are relatively small. The highest value is $0.105$ for NET\_SCIENCE, for the ranking created using hub measure and if only 10\% of nodes with the smallest value of hub measure was selected. Moreover, for all networks (please see supplement material) the divergence is the highest for small sample sizes (10\%, 20\% and 30\%). This is intuitive, but what is more important, the maximum value of divergence never exceeds 1 (see Supplement Material). Taking into account the fact that the limit of the measure is infinity, this value is acceptable from the perspective of data sampling. 

The fall of Kullback--Leibler divergence values with the increasing percentage of nodes used for learning is quite obvious and visible in Figures~\ref{fig:KL_pairs} and~\ref{fig:KL_netscience}. Even for not perfect distribution adjustment for smaller contribution of selected nodes we achieve very good representativeness (KL divergence value at the level of  0.01) already for 50\% of nodes, see Figure \ref{fig:KL_netscience}.

\begin{figure}
\centering
\includegraphics[width=0.7\textheight]{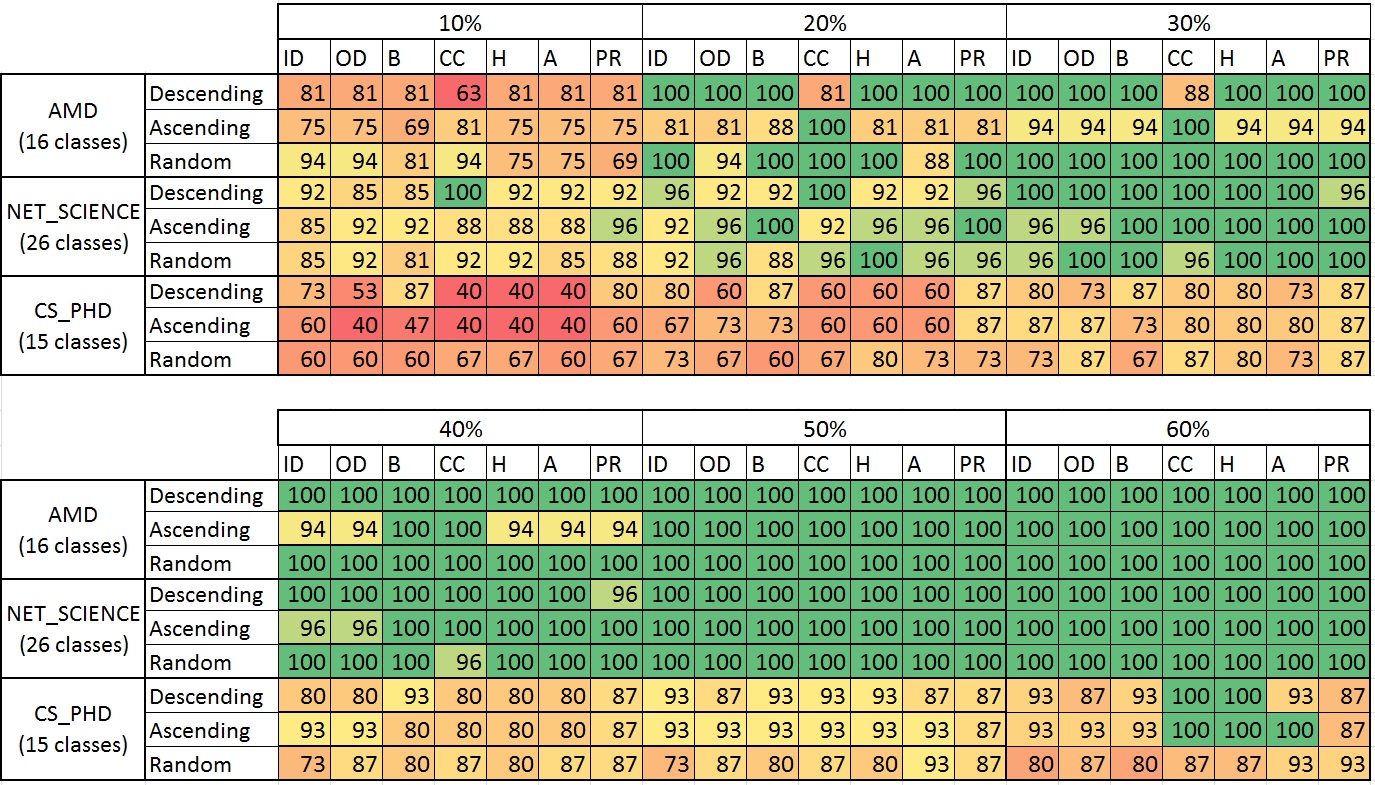}
\includegraphics[width=0.7\textheight]{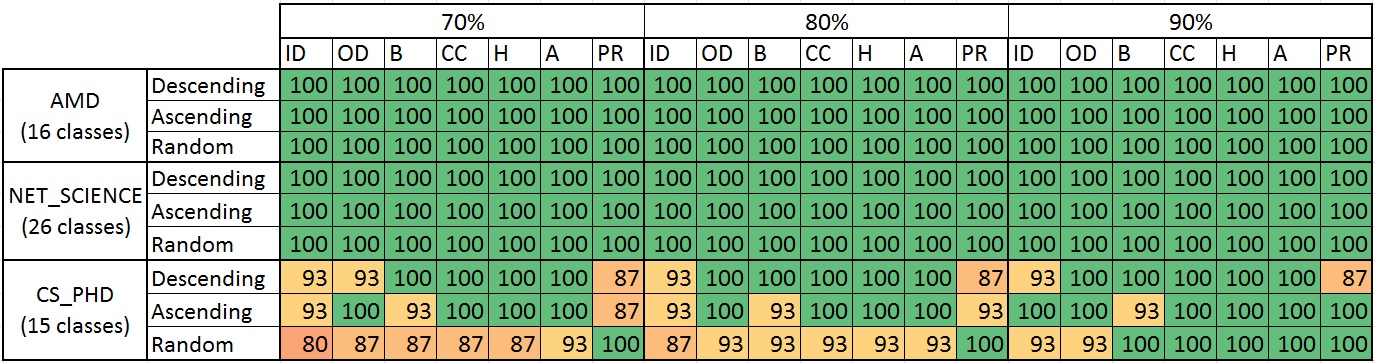}
\caption{Percentage of the classes uncovered in the initial node set used for learning; ID - indegree, OD - outdegree, B - betweeneess, CC- clustering coefficient, H - hubness, A - authority, PR - page rank. 
\label{fig:no_classes_uncovered}}
\end{figure}

One of the main challenges in active learning and inference is to acquire all classes that exist within a given dataset during the initial node selection. The sampling quality can be also measured by assessing what the percentage of uncovered classes in the process of sampling is. It is very important as if not all classes are discovered in the phase of uncovering initial labels then the method will not be able to generalize these classes during the classification process. The percentage of classes uncovered during each selection of initial nodes process is presented in Figure~\ref{fig:no_classes_uncovered}. Networks PAIRS\_FSG, PAIRS\_FSG\_SMALL and YEAST are neglected as no matter what method was used always all classes were uncovered in the selection process. Also the classification results for those networks are relatively good when comparing with the remaining datasets.

The smallest percentage of classes has been discovered for CD\_PHD network. In some cases even if 90\% of data was sampled, there were still some classes that stayed uncovered. Comparing this outcome with the classification results, it can be noticed that classification error for this data set is very high -- not smaller than 0.96 for LBP ('measure'-neighbour version) and 0.55 for ICA (Figure~\ref{fig:best_results}). This is also partially visible for the NET\_SCIENCE network: not all classes are being uncovered for 10\% or 20\% and the classification error for this percentages exceeded 0.8 (Figure~\ref{fig:best_results}). This mainly results from the profile of these networks. They are compounded of dozens of classes and not all of them can be found within 10\% or 20\% of selected nodes, see Table \ref{tab:datasets}.

\onecolumn 
\twocolumn

\onecolumn
\twocolumn
      
\subsection{Top vs. bottom selection from rankings}

Next step of the analysis is to determine how the results are influenced by using the nodes from top or bottom of particular rankings.

\begin{figure}
\centering
\includegraphics[width=1\columnwidth]{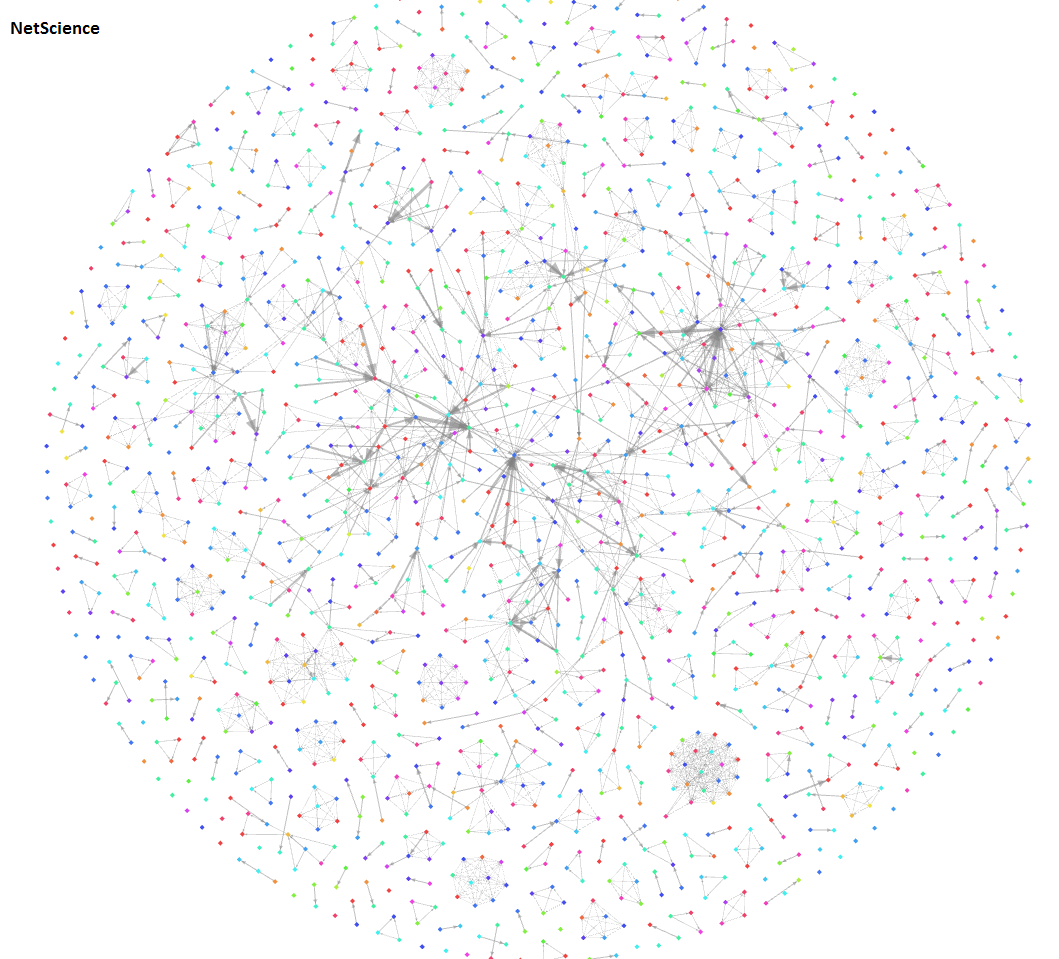}
\caption{The visualisation of the NET\_SCIENCE network; colours representv
 classes (labels). \label{fig:u_artificial}}
\end{figure}

The results revealed that in most cases the methods using top nodes from ranks were better (see Table~\ref{tab:topbottom}). However, some exceptions from this rule can be noticed. LBP was performing better for in--degree based ranks, if used nodes from the bottom of ranks. It was because nodes with low in--degree in some datasets had large out--degree, so they were able to propagate the label effectively and it was the same label within their direct neighbours, see e.g. Figure~\ref{fig:u_artificial}.

\begin{figure}
\centering
\includegraphics[width=1\columnwidth]{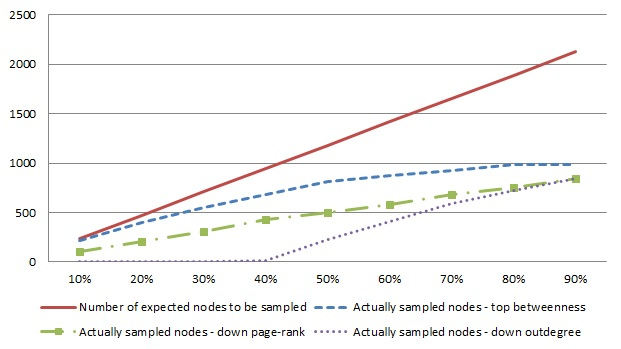}
\caption{The comparison of the number of nodes which theoretically should be sampled against actually sampled for the YEAST dataset and the 'measure'-neighbour method.
\label{fig:discovered}}
\end{figure}

Due to the fact that the selection of training set based on 'measure'--neighbour sampling is heavily dependent on the structure of the network, it happened that the number of neighbouring nodes utilized for learning and inference was smaller than the nominal number of nodes taken from the ranking. According to the nature of LBP, while using this algorithm regardless of the training set selection method (original and 'measure'--neighbour one), if a node taken from the ranking has no neighbour, the information about its label will not be propagated. On the other hand, the ICA method is able to overcome this problem and the label may be assigned to even disconnected nodes. This phenomenon can be observed e.g. for the YEAST dataset in Figure~\ref{fig:discovered}. In general, this LBP drawback did not influence the results so much, except one dataset -- CS\_PHD, where the network was highly disconnected and almost no nodes were labelled in the 'measure'-neighbour algorithm. In other cases, LBP--neighbour method outperformed typical LBP in both approaches -- top and bottom -- despite the fact that less nodes were used as an input for classification algorithm.

\section{Conclusions and future work\label{sec:conclusions}}

Active learning is an important problem that occurs when we need to specify what network sample should be taken to initially acquire its node labels (classes) in order to classify the rest of the network. In this paper various strategies of active learning for within--network classification were studied. 

In particular, two representative classification algorithms: locally--driven Iterative Classification Algorithm (ICA) and globally--based Loopy Belief Propagation (LBP) were tested. 

For each of them, seven main structure--based measures for node ranking were experimentally examined: (1) indegree, (2) outdegree, (3) betweenness, (4) clustering coefficient, (5) hubness, (6) authority and (7)~page rank. Additionally, a new 'measure'--neighbour set of methods was proposed in Section \ref{sec:newUtilityScores}. Its novel idea is to select for the initial acquisition not the nodes taken from a given ranking but their neighbours. Besides, for each ranking list either top or bottom nodes were considered for label discovery. In total, 29 selection methods were tested: 14 for original structural measures (7 measures with either 'top' or 'bottom' approach), 14 for 'measure'--neighbour selection and the random one. All of them were compared with each other. 

Experiments were carried out on six real--world datasets with different network profiles and diverse number of classes. 

The outcomes revealed that depending on both (1) network profiles and (2) complexity of class conditional probabilities the distinct settings of the methods perform differently. 
For example, inference applied within networks that exhibits small--world properties (with high clustering coefficient) give good performance for 'measure'--neighbour methods. However, this does not hold for random networks with very low connectivity  where 'measure'--neighbour approaches are outperformed by original and random methods (networks from Group 3).

Also, the results significantly depend on the distribution of individual classes among nodes with a given measure, e.g. top linked nodes (high degree) can belong to some classes more frequently than on average within the network (class imbalance). It is quite visible, if we compare results for selection methods with not close to zero values of Kullback--Leibler divergence between the selected set and the entire node set. The classification accuracy for such approaches is usually worse than in the cases when class distributions for the sample set and for the whole dataset  match closely  each other.

Overall, the new 'measure'--neighbour selection methods proposed in the paper performed better than their original approaches. They also more often surpassed the random selection in the final inference error level. It leads to one of the main findings of this paper: the relational inference is more effective if we learn on the labels of neighbours of the nodes with a given structural property (degree, page rank, etc.) rather than  on the labels of those nodes.

It should be also emphasized that none of the presented methods was able to generalize the datasets with many classes (\textgreater 10) at the satisfactory level, especially for smaller percentage of learning nodes.

In general, the experimental results presented in the paper have shown that the final classification quality depends on many factors like selection strategy, size of the learning set (percentage of all nodes), inference algorithm and network specific profile. However, in each case we can find many selection strategies that result in lower level of classification error than simple random approach. This is very important for real world applications especially when the total cost of wrong classification is high. The better and well adapted to a given environment selection mechanism, the lower misclassification level and lower costs. It plays an important role e.g. in marketing of frequently changing products or services where it is hardly possible to collect feedback from larger communities of potential customers. Also at high risk screening for very rare diseases but with very expensive diagnostic tests and fatal consequences, it is difficult to acquire an accurate group of patients that may suffer from such illness. Thus, more effective methods, including network--based, need to be applied. Better initial selection methodologies for complex networks may reduce costs in such cases.

The development of general adaptation rules that would enable to adjust node selection method to the network structural profile and class distributions is a new future research direction derived from the paper conclusions. Additionally, active learning can be seen as an iterative process with adaptive selection of  more and more nodes. It would complicate the learning even more.


\section*{Acknowledgments}
This work was partially supported by the European Union as part of the European Social Fund, the European Commission under the 7th Framework Programme, Coordination and Support Action, Grant Agreement Number 316097, ENGINE - European research centre of Network intelliGence for INnovation Enhancement (http://engine.pwr.wroc.pl/), and The National Science Centre, the decision no. DEC-2013/09/B/ST6/02317.


\bibliographystyle{abbrv}
\bibliography{activelearning}

\begin{thebibliography}{10}

\bibitem{angl88}
D.~Angluin.
\newblock Queries and concept learning.
\newblock {\em Machine Learning}, 2:319--342, 1988.

\bibitem{atpr11}
J.~Attenberg and F.~Provost.
\newblock Online active inference and learning.
\newblock In {\em Proceedings of the Seventeenth ACM SIGKDD International
  Conference on Knowledge Discovery and Data Mining}, 2011.

\bibitem{babe06}
M.~Balcan, A.~Beygelzimer, and J.~Langford.
\newblock Agnostic active learning.
\newblock In {\em Proceedings of the 23rd International Conference on Machine
  learning}, pages 65--72, 2006.

\bibitem{beda09}
A.~Beygelzimer, S.~Dasgupta, and J.~Langford.
\newblock Importance weighted active learning.
\newblock In {\em Proceedings of the 26-th International Conference on Machine
  Learning}, 2009.

\bibitem{bige08}
M.~Bilgic and L.~Getoor.
\newblock Effective label acquisition for collective classification.
\newblock In {\em Proceedings of the ACM SIGKDD International Conference on
  Knowledge Discovery and Data Mining}, pages 43--51, 2008.

\bibitem{bige10}
M.~Bilgic and L.~Getoor.
\newblock Active inference for collective classification.
\newblock In {\em Proceedings of Twenty-Fourth Conference on Artificial
  Intelligence AAAI'10}. AAAI Press, 2010.

\bibitem{brei96}
L.~Breiman.
\newblock Bagging predictors.
\newblock {\em Machine Learning}, 24(2):123--140, 1996.

\bibitem{breiman2001random}
L.~Breiman.
\newblock Random forests.
\newblock {\em Machine learning}, 45(1):5--32, 2001.

\bibitem{Sun:2003}
D.~Bu, Y.~Zhao, L.~Cai, H.~Xue, X.~Zhu, H.~Lu, J.~Zhang, S.~Sun, L.~Ling,
  N.~Zhang, G.~Li, and R.~Chen.
\newblock {Topological structure analysis of the protein–protein interaction
  network in budding yeast}.
\newblock {\em Nucleic Acids Research}, 31(9):2443--2450, 2003.

\bibitem{deka09}
C.~Desrosiers and G.~Karypis.
\newblock Within-network classification using local structure similarity.
\newblock {\em Lecture Notes in Computer Science}, 5781:260--275, 2009.

\bibitem{elne11}
H.~Eldardiry and J.~Neville.
\newblock Across-model collective ensemble classification.
\newblock {\em Association for the Advancement of Artificial Intelligence},
  2011.

\bibitem{elne12}
H.~Eldardiry and J.~Neville.
\newblock An analysis of how ensembles of collective classifiers improve
  predictions in graphs.
\newblock In {\em Proceedings of the 21st ACM International Conference on
  Information and Knowledge Management}, 2012.

\bibitem{faje08}
A.~Fast and D.~Jensen.
\newblock Why stacked models perform effective collective classification.
\newblock In {\em Proceedings of the 2008 Eighth IEEE International Conference
  on Data Mining.}, pages 785--790. IEEE, 2008.

\bibitem{gael08}
B.~Gallagher and T.~Eliassi-Rad.
\newblock Leveraging label-independent features for classification in sparsely
  labeled networks: An empirical study.
\newblock In {\em SNA-KDD'08}. ACM, 2008.

\bibitem{gege84}
S.~Geman and D.~Geman.
\newblock Stochastic relaxation, gibbs distributions and the bayesian
  restoration of images.
\newblock {\em IEEE Transactions on Pattern Analysis and Machine Intelligence},
  6:721--741, 1984.

\bibitem{kaka13}
T.~Kajdanowicz and P.~Kazienko.
\newblock Collective classification.
\newblock {\em Encyclopedia of Social Network Analysis and Mining, Springer},
  2013.

\bibitem{kaka12a}
T.~Kajdanowicz, P.~Kazienko, and M.~Janczak.
\newblock Collective classification techniques: an experimental study.
\newblock {\em New Trends in Databases and Information Systems}, 185:99--108,
  2012.

\bibitem{kaka12}
K.~Kazienko and T.~Kajdanowicz.
\newblock Label-dependent node classification in the network.
\newblock {\em Neurocomputing}, 75(1):199--209, 2012.

\bibitem{kamu11}
P.~Kazienko, K.~Musial, and T.~Kajdanowicz.
\newblock Multidimensional social network in the social recommender system.
\newblock {\em IEEE Transactions on Systems, Man and Cybernetics - Part A:
  Systems and Humans}, 41(4):746--759, 2011.

\bibitem{knde01}
A.~Knobbe, M.~de~Haas, and A.~Siebes.
\newblock Propositionalisation and aggregates.
\newblock In {\em Proceedings of Fifth European Conference on Principles of
  Data Mining and Knowledge Discovery}, pages 277--288, 2001.

\bibitem{kune11}
A.~Kuwadekar and J.~Neville.
\newblock Relational active learning for joint collective classification
  models.
\newblock In {\em Proceedings of the 28th International Conference on Machine
  Learning}, pages 385--392, 2011.

\bibitem{luge03}
Q.~Lu and L.~Getoor.
\newblock Link-based classification.
\newblock In {\em Proceedings of 20th International Conference on Machine
  Learning ICML}, pages 496--503, 2003.

\bibitem{muka13}
K.~Musial and P.~Kazienko.
\newblock Social networks on the internet.
\newblock {\em World Wide Web Journal}, 16(1):31--72, 2013.

\bibitem{Musial09}
K.~Musia\l, P.~Kazienko, and P.~Br\'{o}dka.
\newblock User position measures in social networks.
\newblock In {\em Proceedings of the 3rd Workshop on Social Network Mining and
  Analysis}, SNA-KDD '09, pages 6:1--6:9, New York, NY, USA, 2009. ACM.

\bibitem{nalo12}
G.~Namata, B.~London, L.~Getoor, and B.~Huang.
\newblock Query-driven active surveying for collective classification.
\newblock In {\em Workshop on Mining and Learning with Graphs, International
  Conference on Machine Learning ICML}, 2012.

\bibitem{Newman:2006}
M.~E.~J. Newman.
\newblock Finding community structure in networks using the eigenvectors of
  matrices.
\newblock {\em Physical Review E}, 74:036104+, 2006.

\bibitem{Nooy:2004}
W.~Nooy, A.~Mrvar, and V.~Batagelj.
\newblock {\em Exploratory Social Network Analysis with Pajek}, chapter~11.
\newblock Cambridge University Press, 2004.

\bibitem{pear88}
J.~Pearl.
\newblock {\em Probabilistic reasoning in intelligent systems.}
\newblock Morgan Kaufmann, 1988.

\bibitem{rama07}
M.~Rattigan, M.~Maier, and D.~Jensen.
\newblock Exploiting network structure for active inference in collective
  classification.
\newblock In {\em Proceedings of ICDM Workshop on Mining Graphs and Complex
  Structures}, 2007.

\bibitem{sena08}
P.~Sen, G.~Namata, M.~Bilgic, L.~Getoor, B.~Gallagher, and T.~Eliassi-Rad.
\newblock Collective classification in network data.
\newblock {\em Artificial Intelligence Magazine}, 29(3):93--106, 2008.

\bibitem{sett09}
B.~Settles.
\newblock Active learning literature survey.
\newblock {\em Computer Sciences Technical Report 1648, University of
  Wisconsin–Madison}, 1995.

\bibitem{sett11}
B.~Settles.
\newblock From theories to queries: Active learning in practice.
\newblock In {\em JMLR Workshop and Conference Proceedings}, pages 16:1--18,
  2011.

\bibitem{seop92}
H.~Seung, M.~Opper, and H.~Sompolinsky.
\newblock Query by committee.
\newblock In {\em Proceedings of the fifth annual Workshop on Computational
  Learning Theory}, pages 287--294, 1992.

\bibitem{sugi06}
M.~Sugiyama.
\newblock Active learning for misspecified models.
\newblock {\em Advances in neural information processing systems},
  18:1305--1312, 2006.

\bibitem{Taskar:2002}
B.~Taskar, P.~Abbeel, and D.~Koller.
\newblock Discriminative probabilistic models for relational data.
\newblock In {\em Proceedings of 18th Conference in Uncertainty in Artificial
  Intelligence}, San Francisco, 2002. Morgan Kaufmann, Publishers.

\end{thebibliography}

\onecolumn
\begin{center}\begin{Large}\textbf{Learning in Unlabelled Networks - An Active Learning and Inference Approach\\ }\end{Large}\end{center}

\begin{center}\textbf{\begin{Large}Supplement materials\end{Large}}\end{center}



\setcounter{section}{0}

\section{\begin{large}Characteristics of networks used in the experiments\end{large}}
Below characteristics of all used in the experiments datasets are presented in a form of boxplots. For each class within a given network such metrics as: indegree centrality, outdegree centrality, betweeness centrality, Page Rank, clustering coefficient, hub centrality, and authority are considered. 

Each boxplots shows: on each
box, the central mark is the median, the edges of the box are the 25th and 75th
percentiles, the whiskers extend to the most extreme data points not considered
outliers, and outliers are plotted individually.

\begin{figure}[ht]  
\centering
\includegraphics[width=1\textwidth]{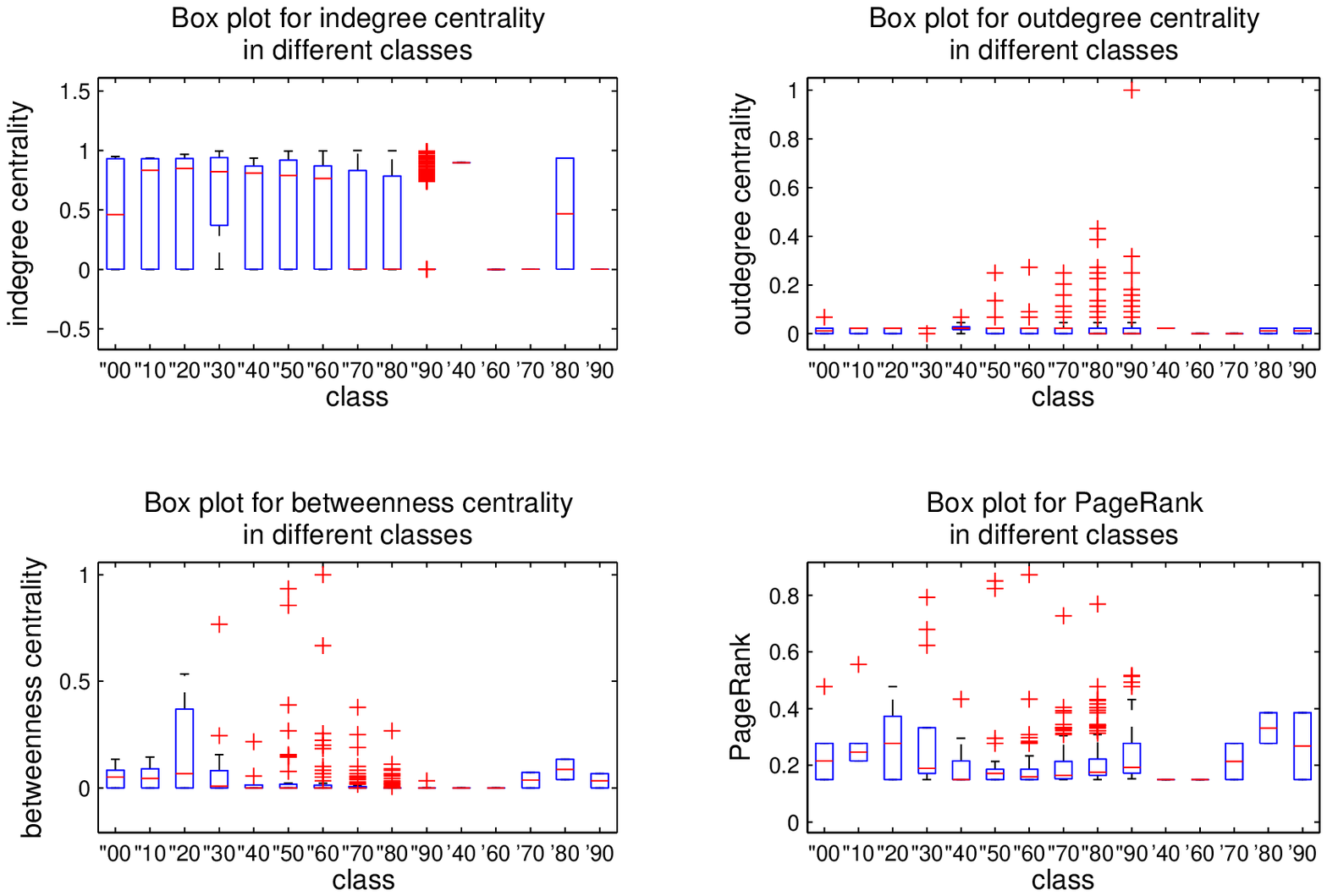}  
\includegraphics[width=1\textwidth]{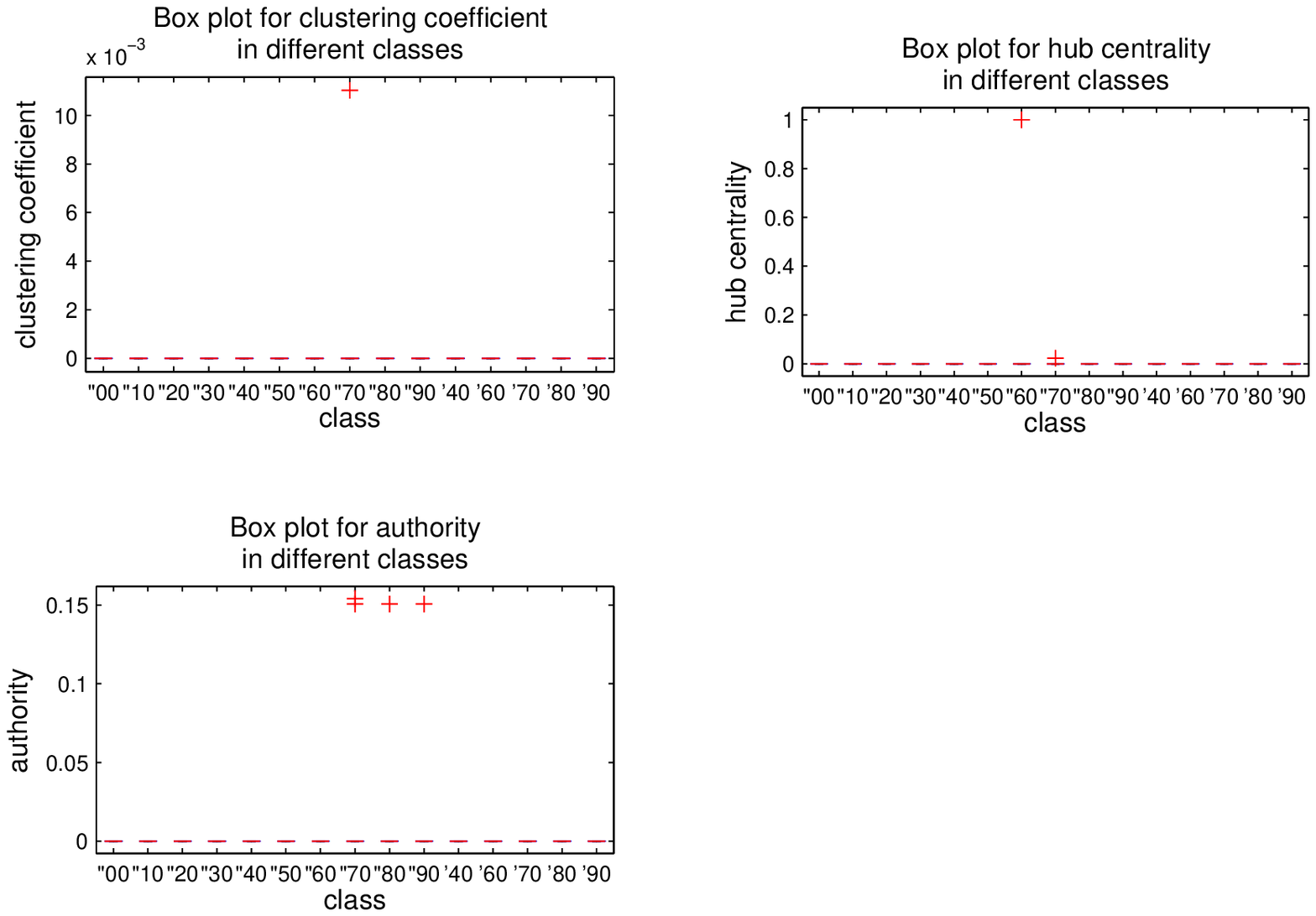}
\caption{\small \sl Characteristics of CSPhd network.
\label{fig:csphd}}  

\end{figure} 

\begin{figure}[ht]  
\centering
\includegraphics[width=1\textwidth]{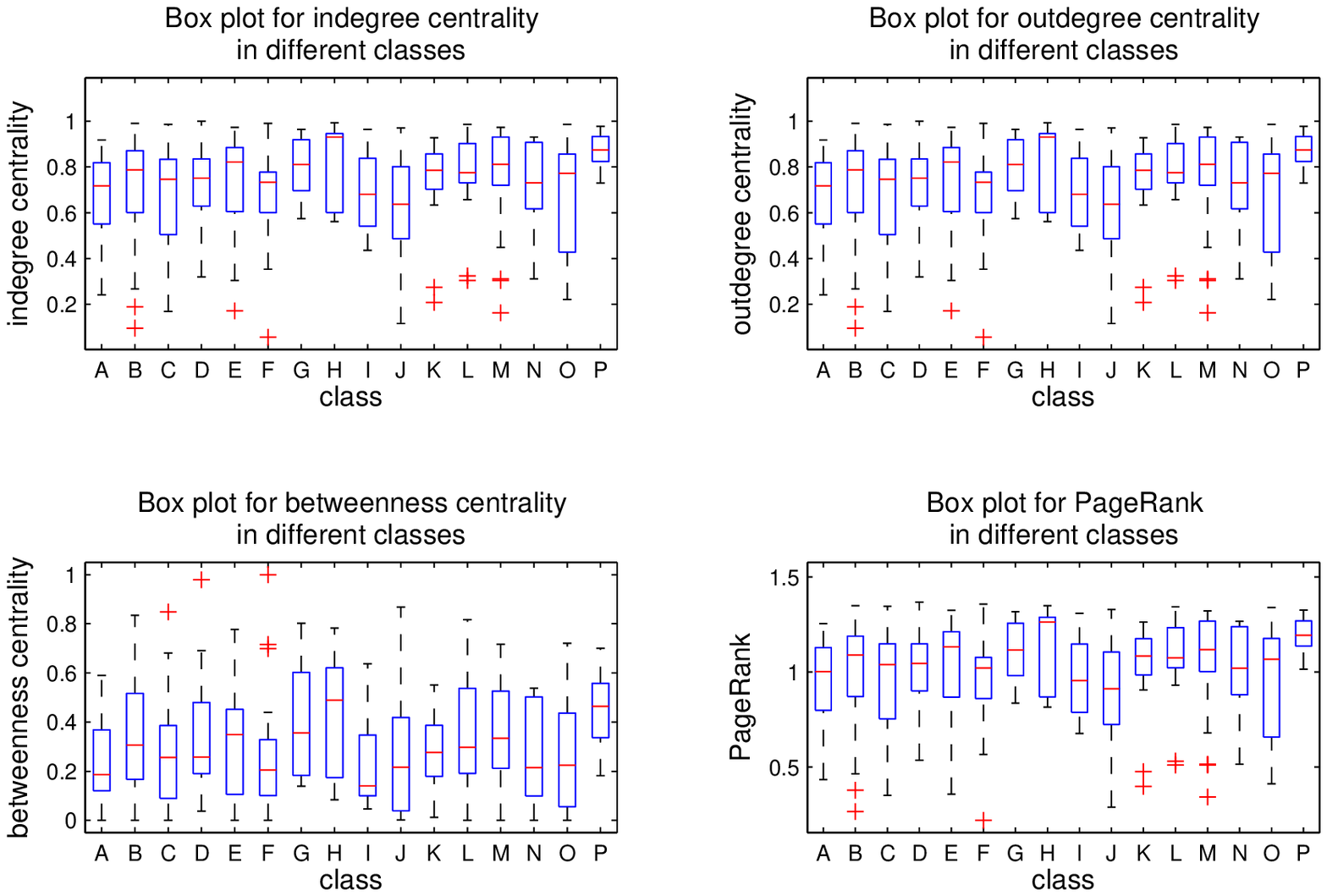}  
\includegraphics[width=1\textwidth]{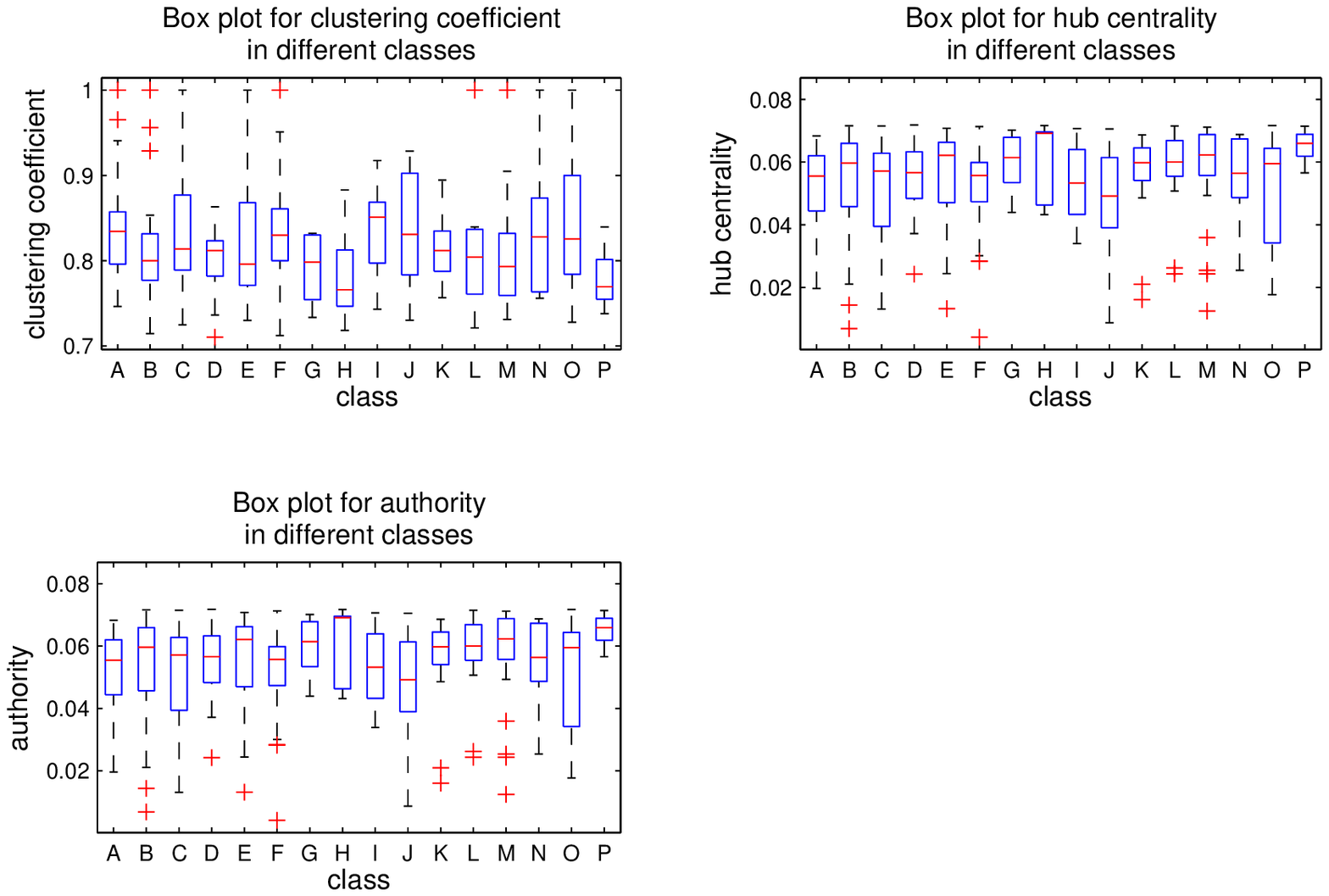}
\caption{\small \sl Characteristics of AMD network.
\label{fig:amd}} 
\end{figure} 

\begin{figure}[ht]  
\centering
\includegraphics[width=1\textwidth]{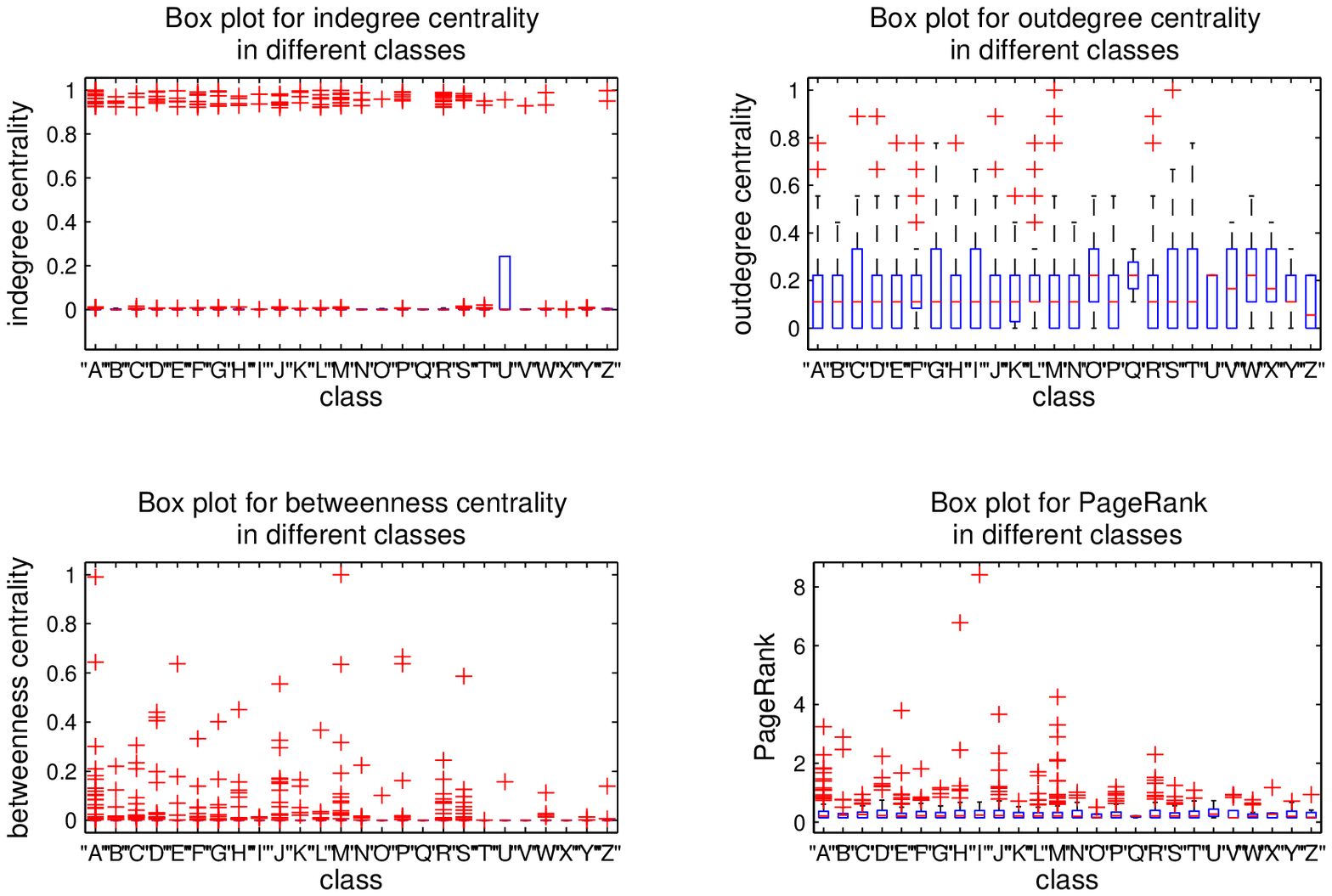}  
\includegraphics[width=1\textwidth]{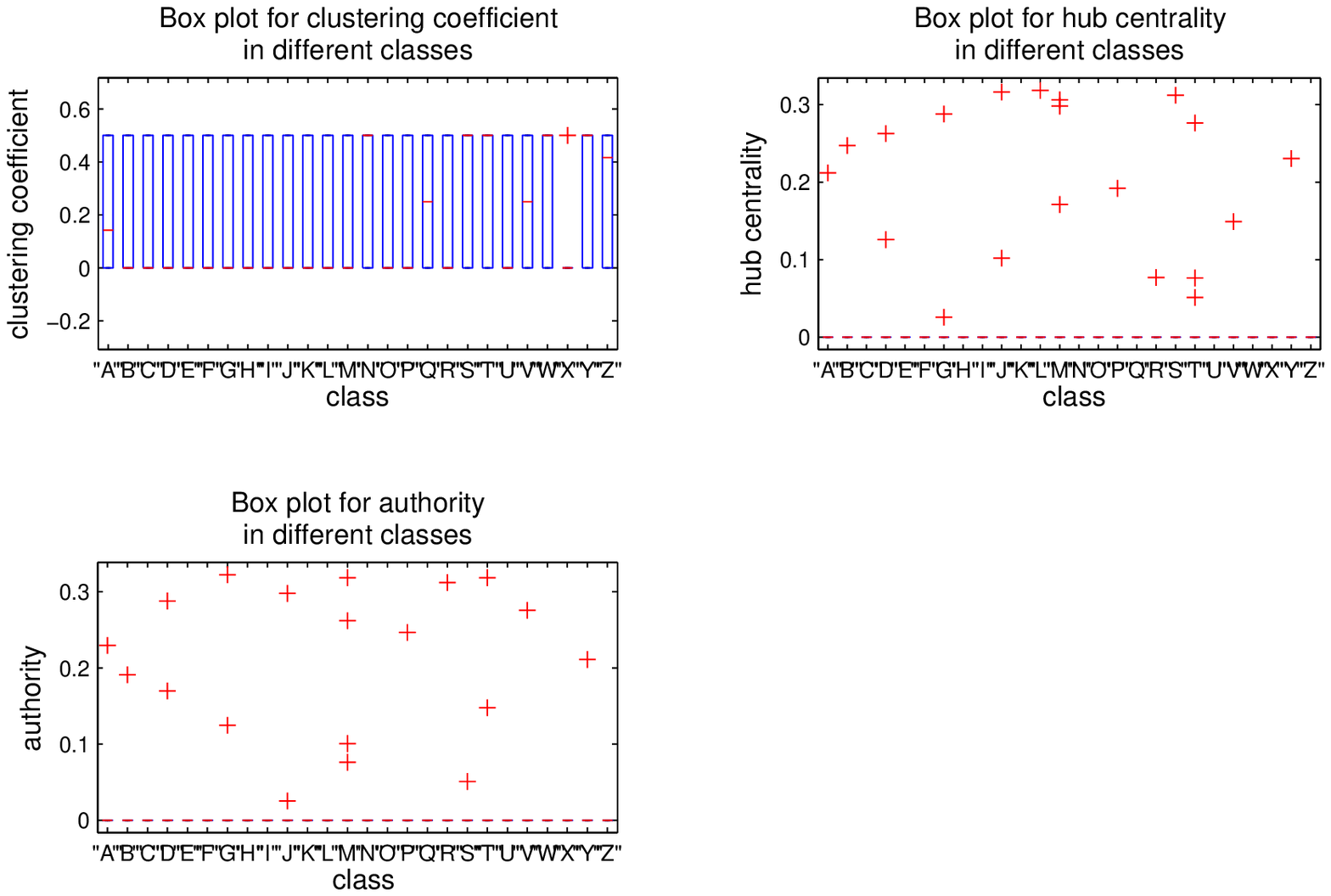}
\caption{\small \sl Characteristics of Net Science network.
\label{fig:net_science}}  
\end{figure}

\begin{figure}[ht]  
\centering  
\includegraphics[width=1\textwidth]{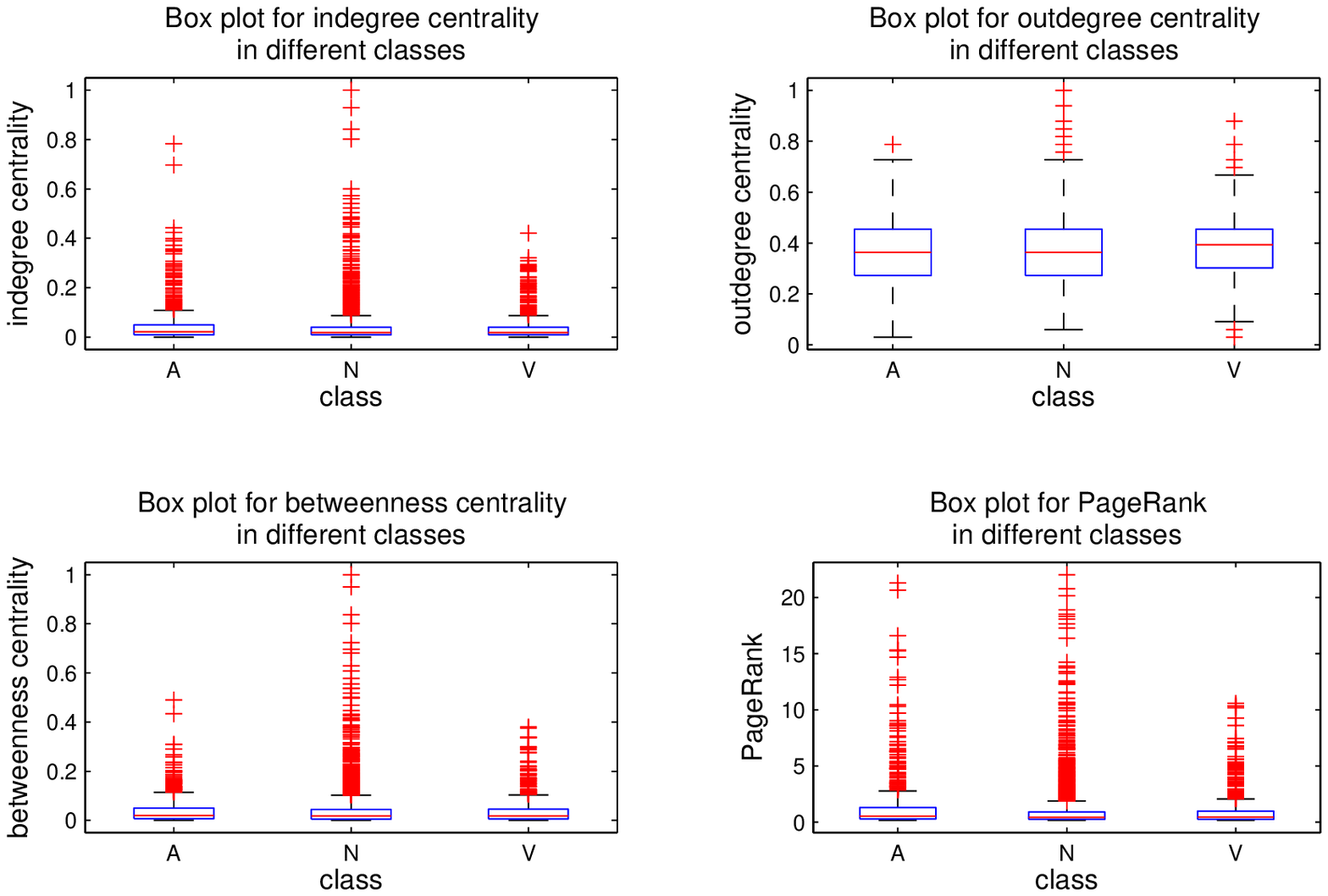}  
\includegraphics[width=1\textwidth]{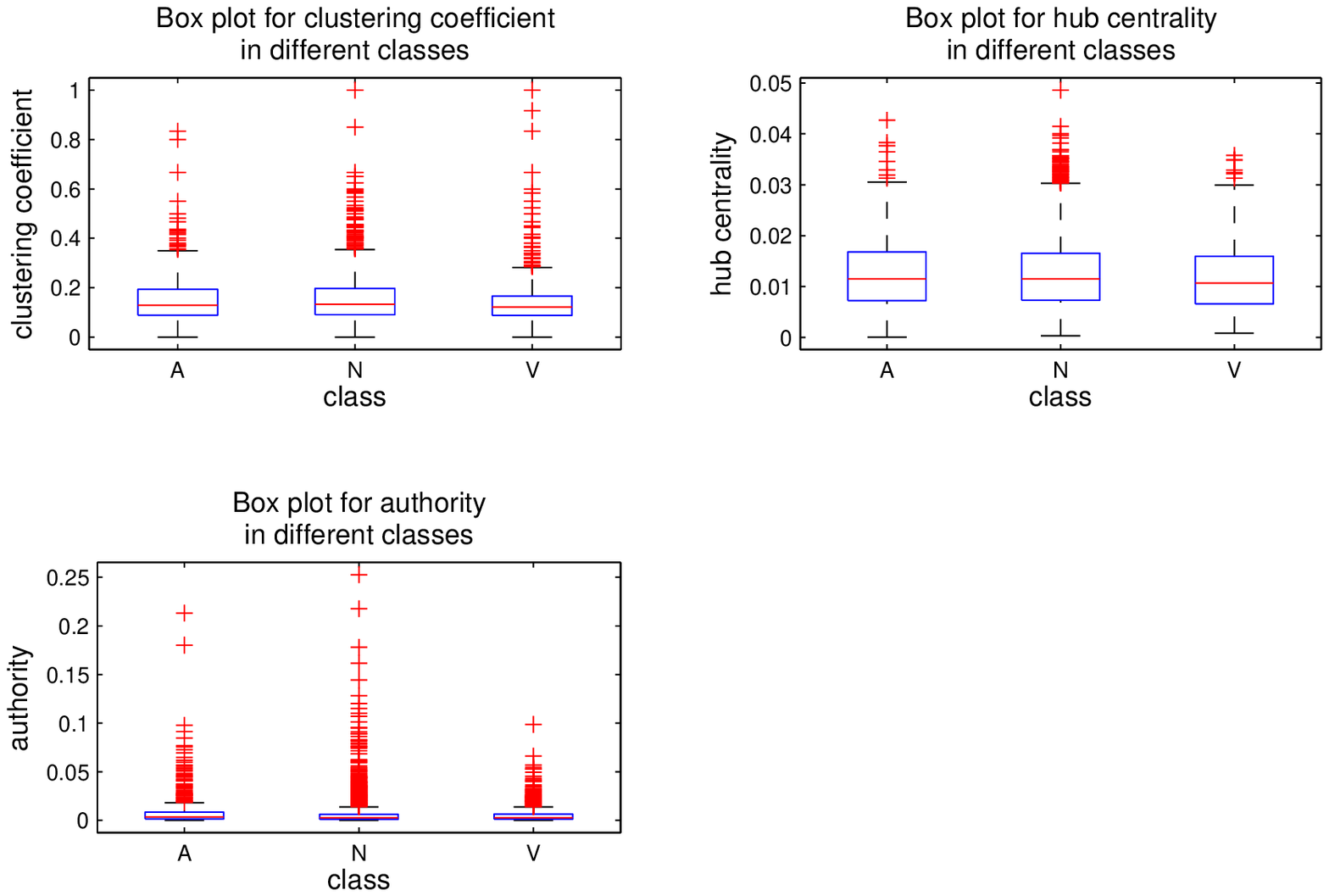}
\caption{\small \sl Characteristics of Pairs FSG network.
\label{fig:pairs}}  
\end{figure}

\begin{figure}[ht]  
\centering
\includegraphics[width=1\textwidth]{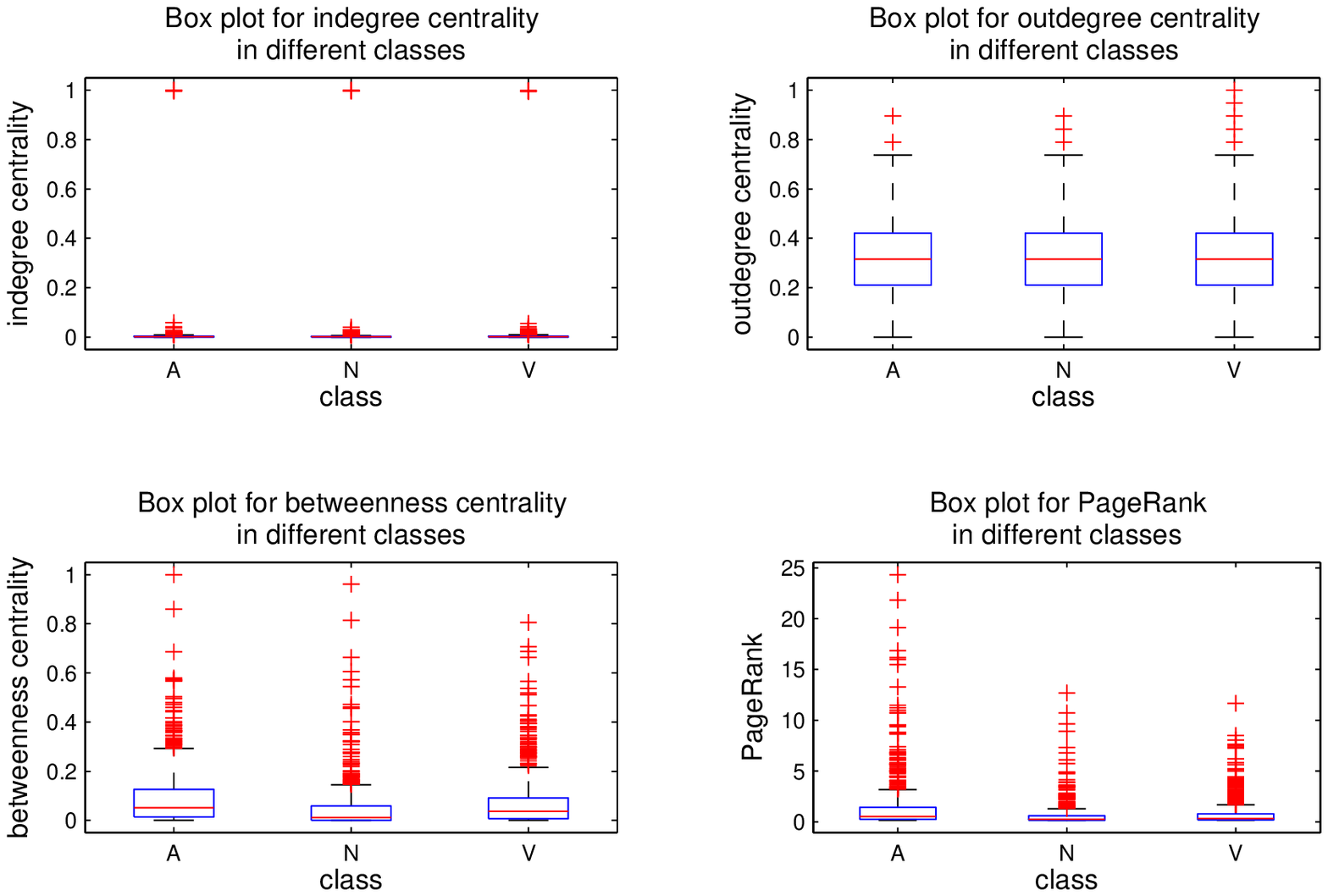}
\includegraphics[width=1\textwidth]{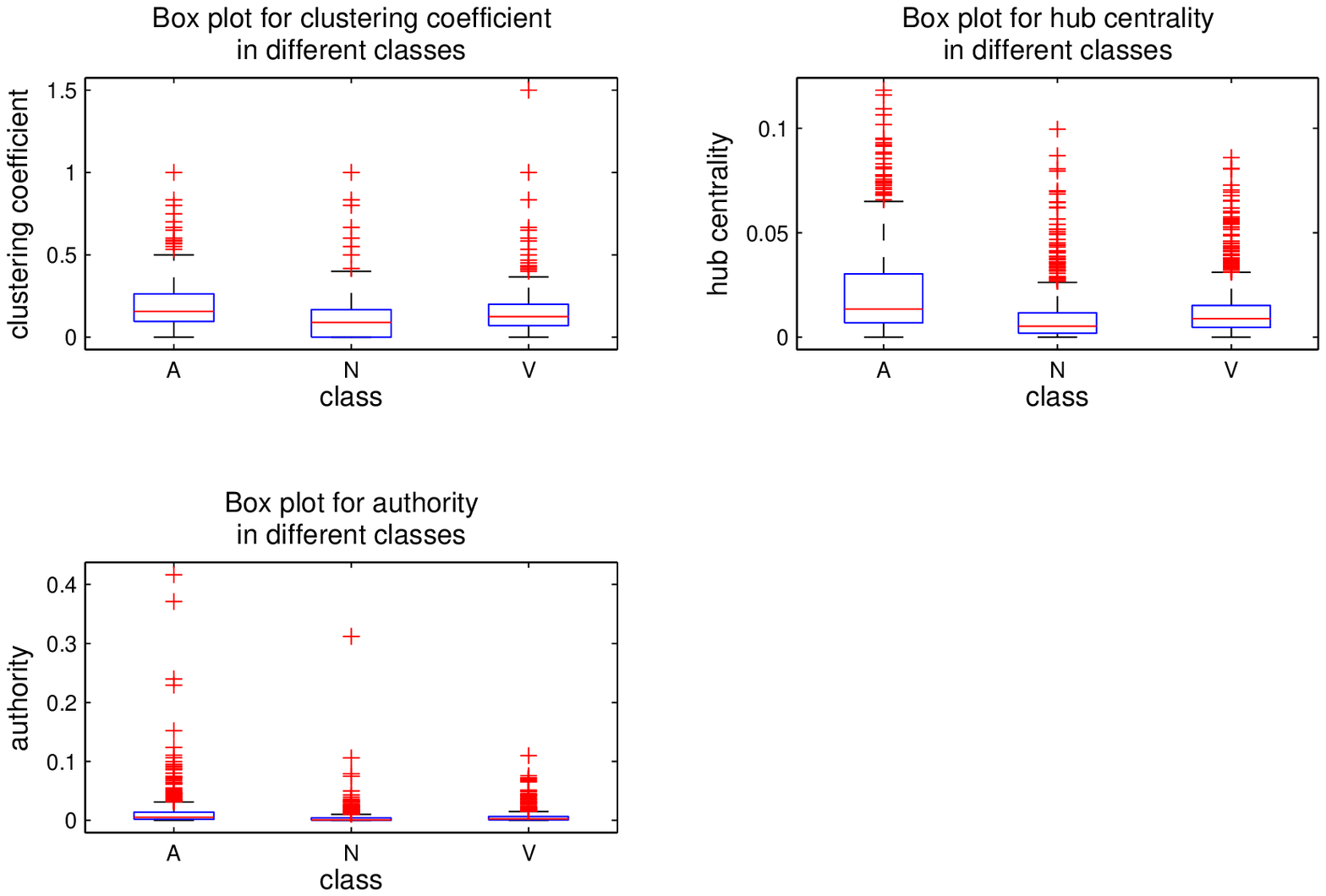}
\caption{\small \sl Characteristics of Pairs FSG small network.
\label{fig:pairs_small}}  
\end{figure}

\begin{figure}[ht]  
\centering
\includegraphics[width=1\textwidth]{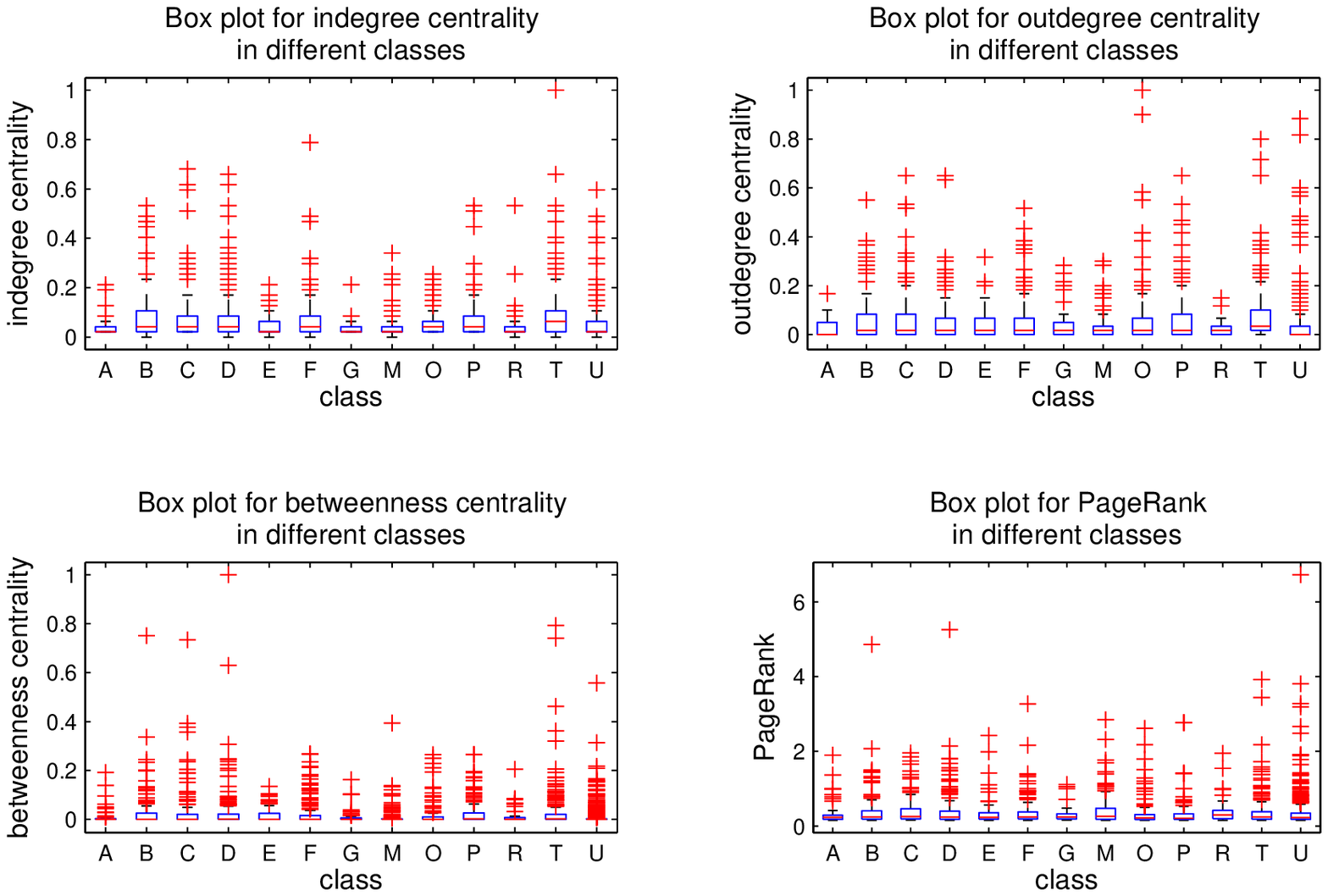}  
\includegraphics[width=1\textwidth]{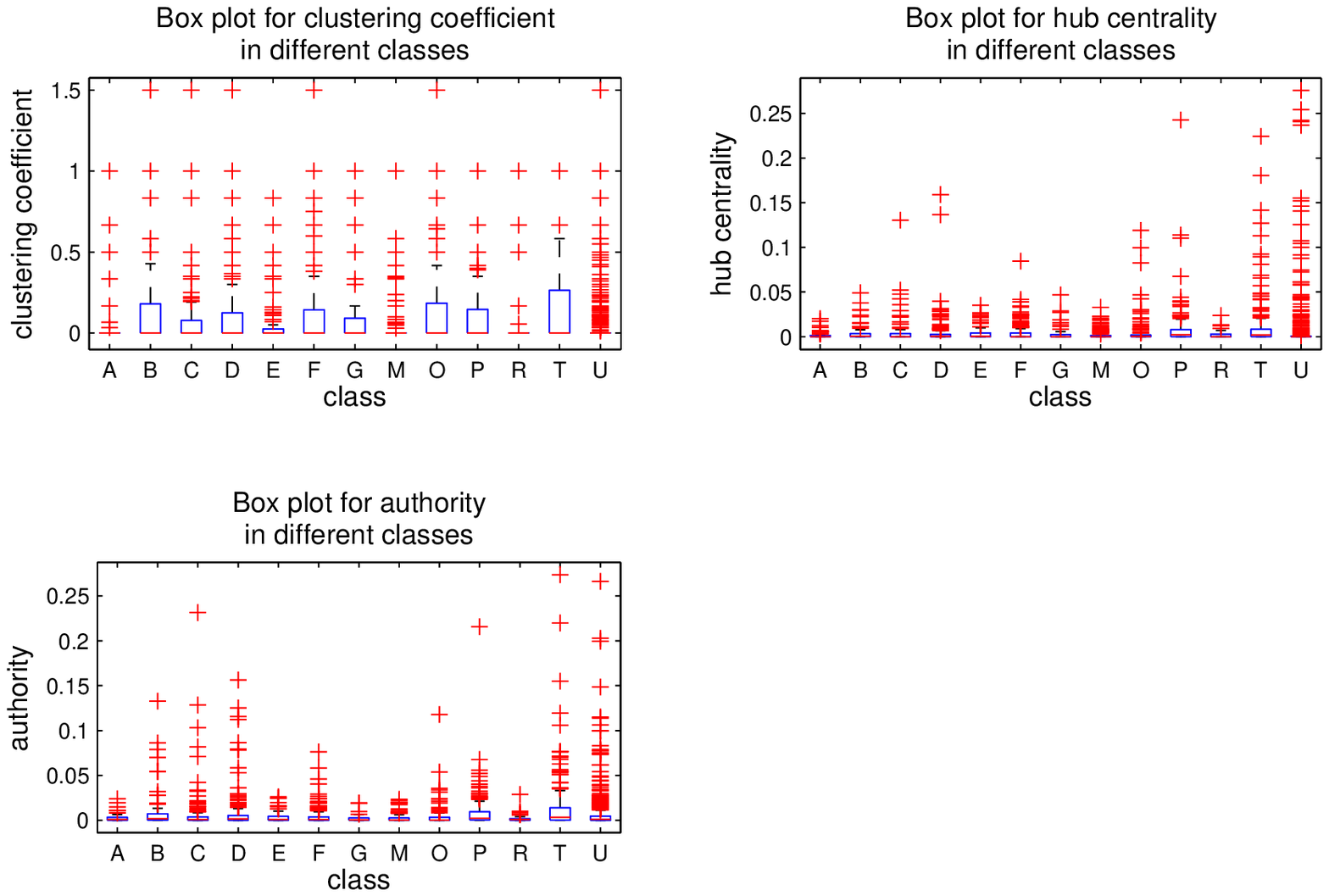}
\caption{\small \sl Characteristics of yeast network.
\label{fig:yeast}}  
\end{figure}

\clearpage

\newpage
\section{\begin{large}Distribution of classes in analysed networks\end{large}}

Below distribution of classes within each analysed network is presented.  

\begin{figure}[ht]
        \begin{subfigure}[b]{0.40\textwidth}
                \includegraphics[width=\textwidth]{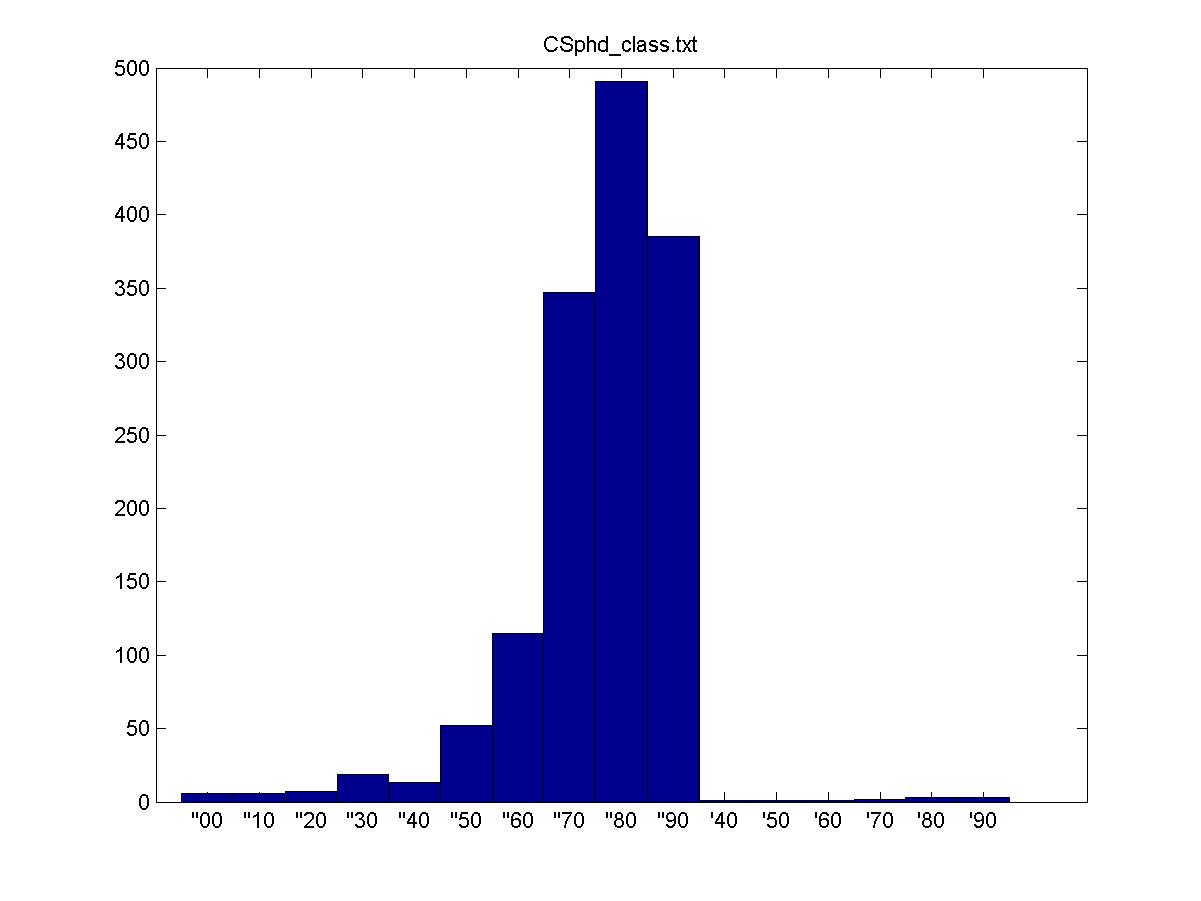}
                \caption{\small \sl CSPhD network}
                \label{fig:csphd_hist}
        \end{subfigure}
        \begin{subfigure}[b]{0.40\textwidth}
                \includegraphics[width=\textwidth]{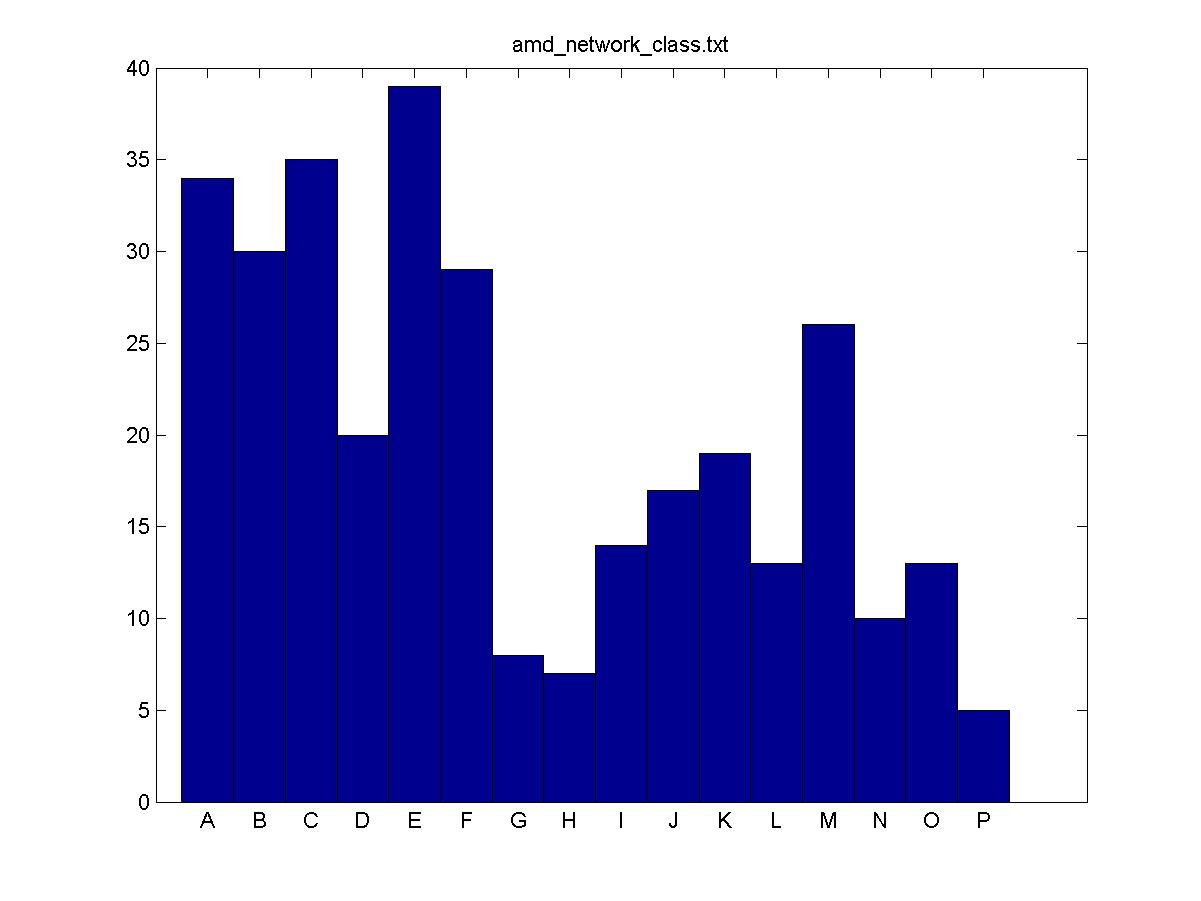}
                \caption{\small \sl AMD network}
                \label{fig:amd_hist}
        \end{subfigure}
        \begin{subfigure}[b]{0.40\textwidth}
                \includegraphics[width=\textwidth]{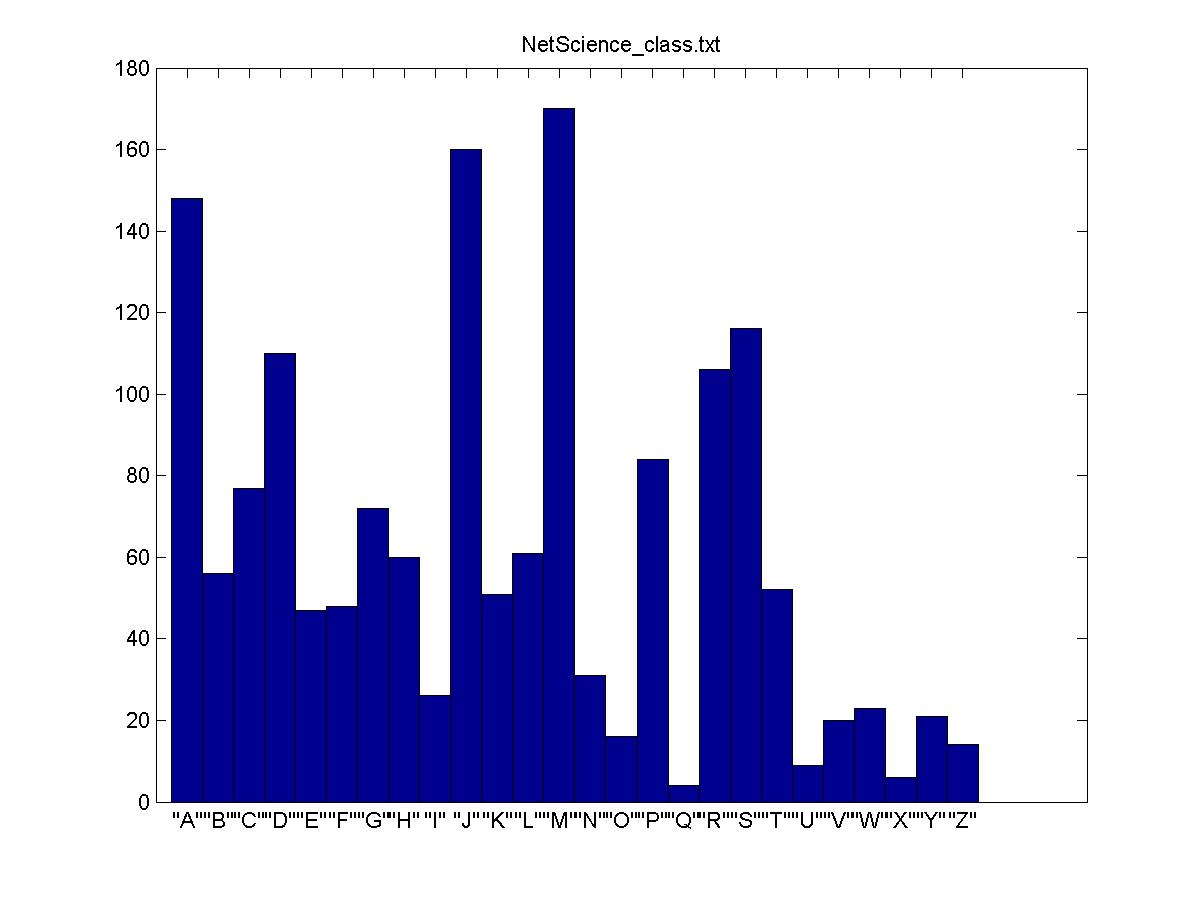}
                \caption{\small \sl NetScience network}
                \label{fig:netscience_hist}
        \end{subfigure}
        \begin{subfigure}[b]{0.40\textwidth}
                \includegraphics[width=\textwidth]{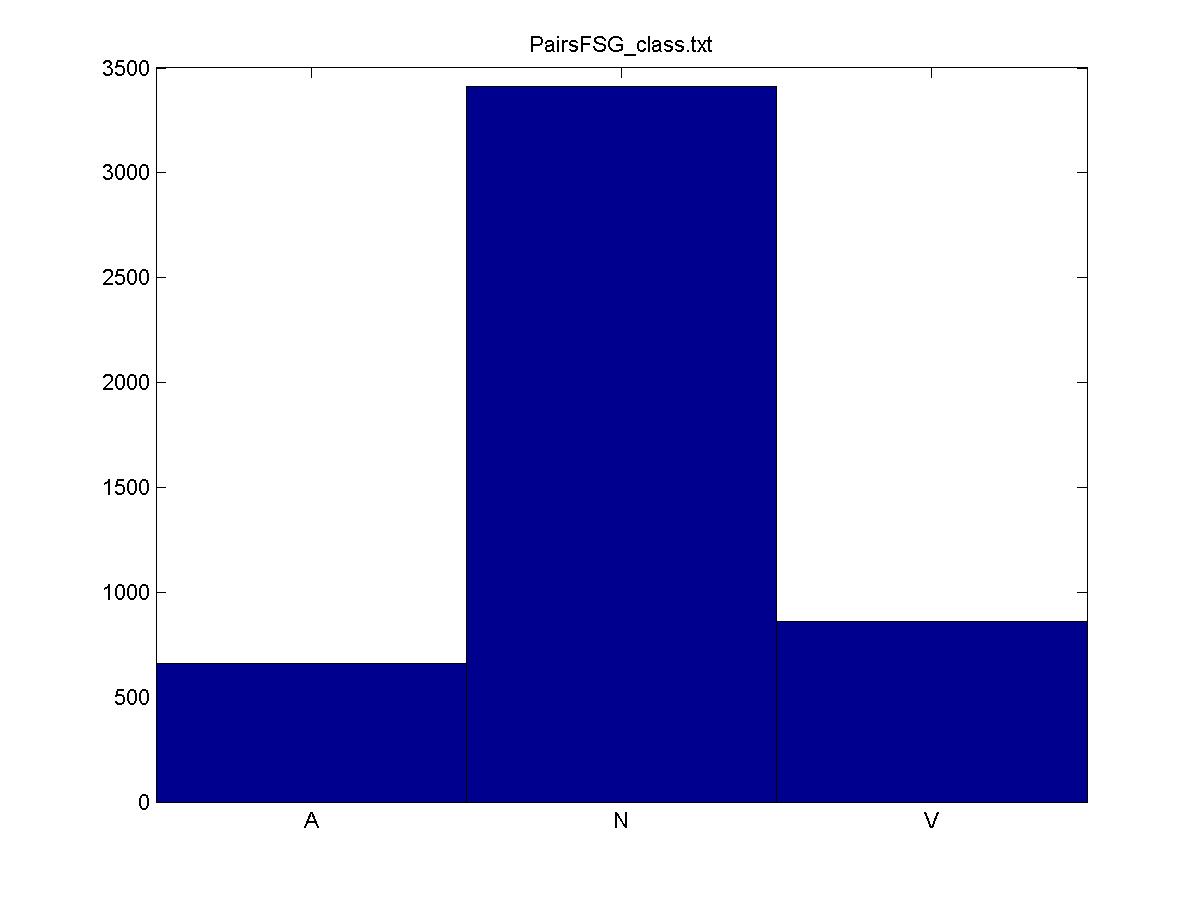}
                \caption{\small \sl PAIRS\_FSG network}
                \label{fig:pairs_hist}
        \end{subfigure}
        \begin{subfigure}[b]{0.40\textwidth}
                \includegraphics[width=\textwidth]{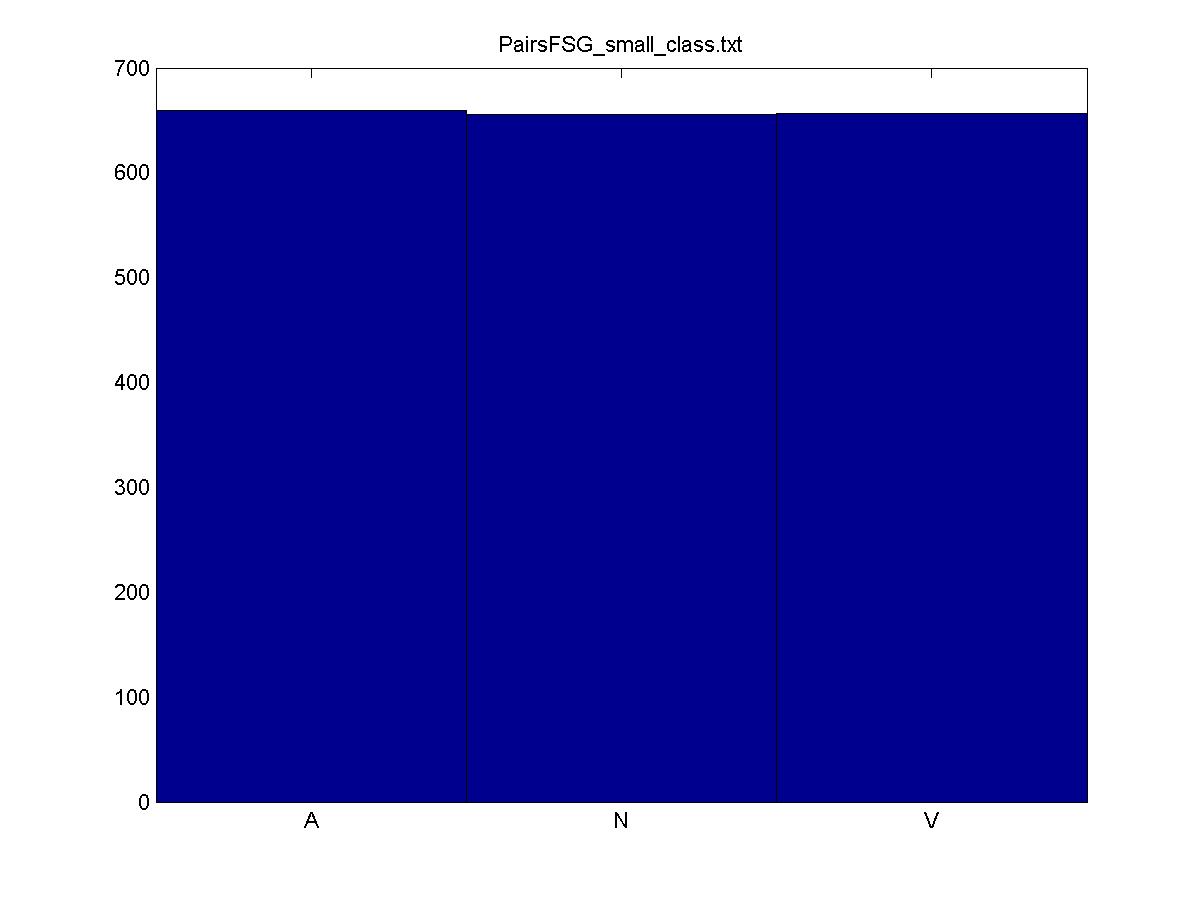}
                \caption{\small \sl PAIRS\_FSG\_small network}
                \label{fig:pairs_small_hist}
        \end{subfigure}
        \begin{subfigure}[b]{0.40\textwidth}
                \includegraphics[width=\textwidth]{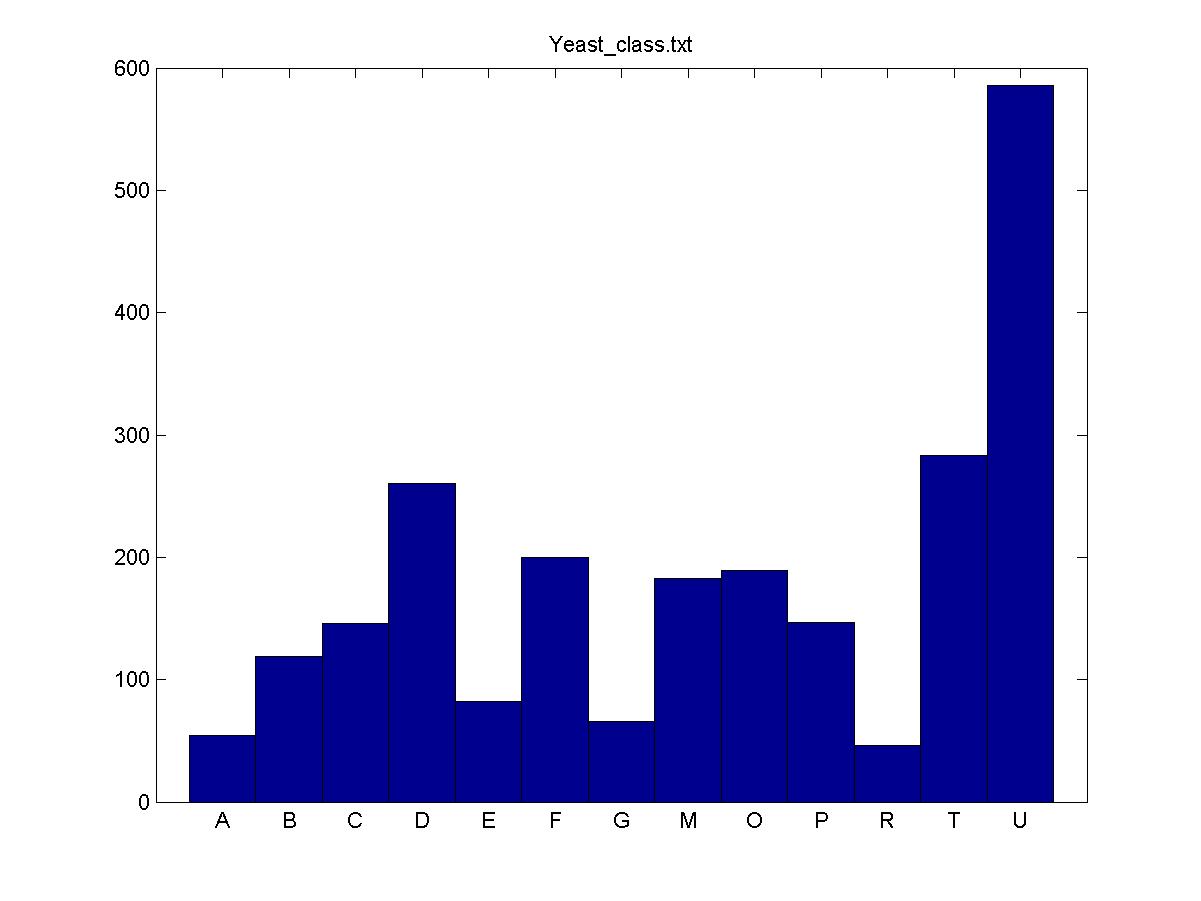}
                \caption{\small \sl YEAST network}
                \label{fig:yeast_hist}
        \end{subfigure}
        \caption{Histograms of classes for all evaluated networks}
\end{figure}

\clearpage

\newpage
\section{\begin{large}Representativeness of sampled data\end{large}}

The representativeness of a data sample is assessed using Kullback--Leibler divergence (a.k.a. relative entropy) which is a measure of the difference between two probability distributions. It measures how much information is lost when one probability distribution (in our case it is a distribution of classes in a given sample -- 10\%, ..., 90\% of the whole dataset) is used to approximate another one (in this paper it is the probability distribution
of classes in the whole dataset).
The smaller the divergence the smaller loss; 0 means that no information is lost.

Below the Kullback--Leibler divergence for each analysed network is presented.\begin{figure}[ht]  
\begin{center}  
\includegraphics[width=1\textwidth]{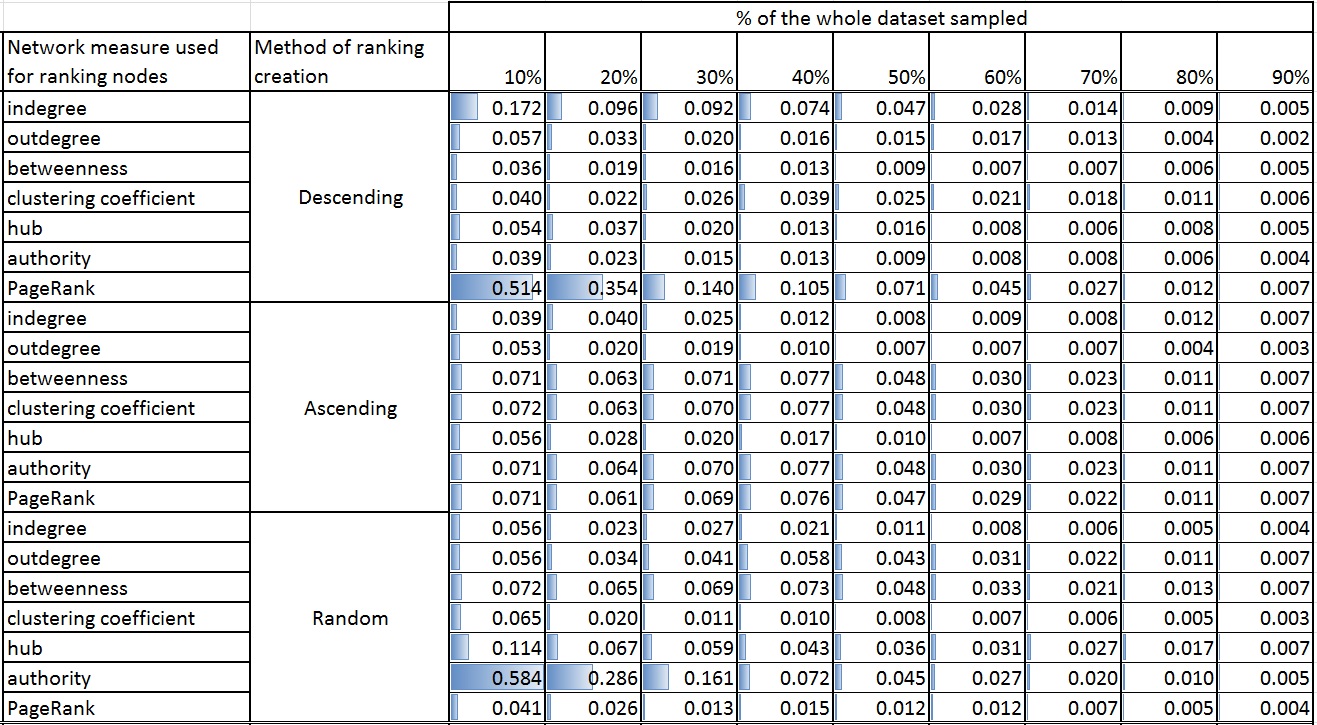} 
\caption{\small \sl Kullback--Leibler divergence for CSPhd network.
\label{fig:kl_div_cs_phd}}  
\end{center}  
\end{figure}

\begin{figure}[ht]  
\begin{center}  
\includegraphics[width=1\textwidth]{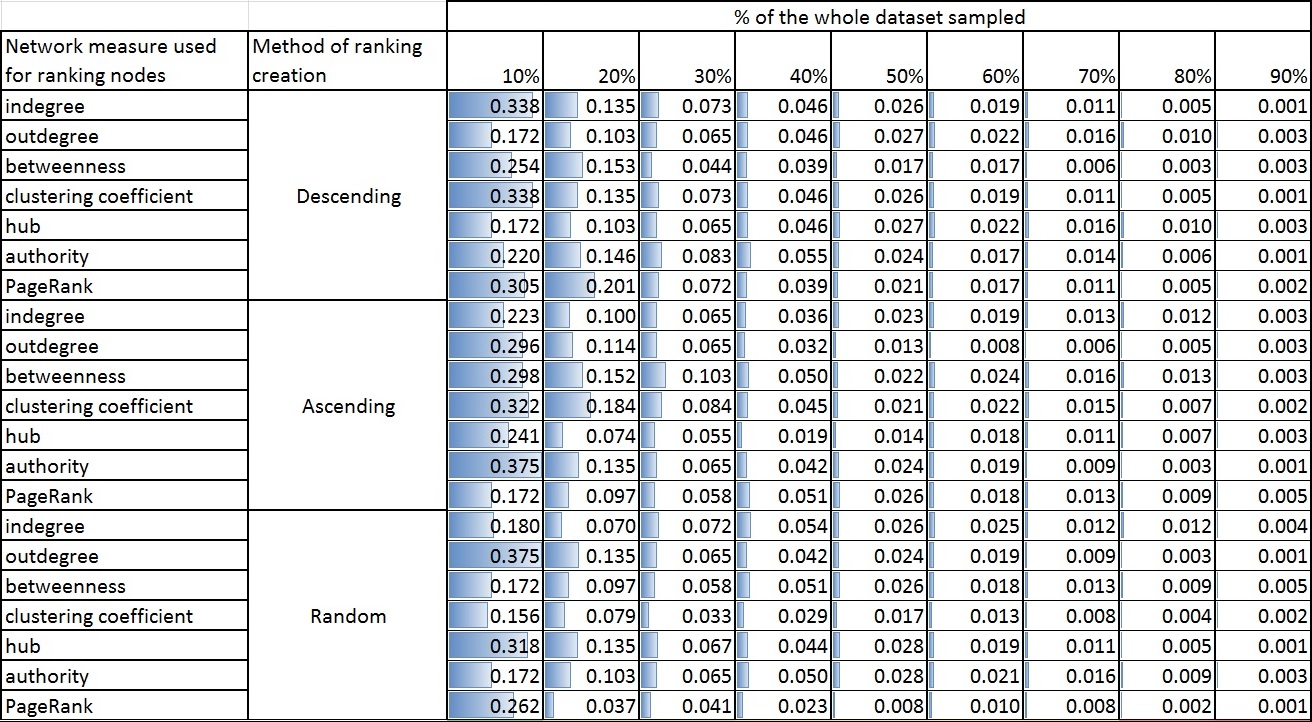} \caption{\small \sl Kullback--Leibler divergence for AMD network.
\label{fig:kl_div_amd}}  
\end{center}  
\end{figure}

\begin{figure}[ht]  
\begin{center}  
\includegraphics[width=1\textwidth]{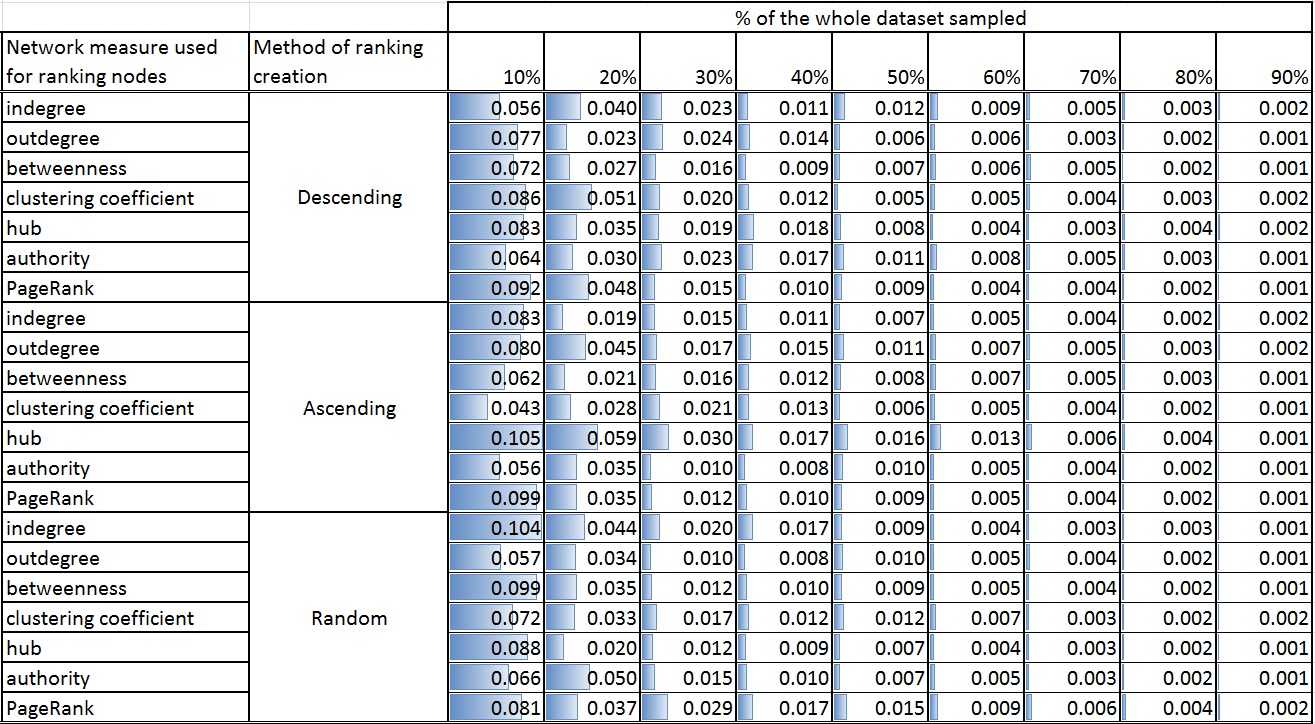} \caption{\small \sl Kullback--Leibler divergence for Net Science network.
\label{fig:kl_div_netsci}}  
\end{center}  
\end{figure} 

\begin{figure}[ht]  
\begin{center}  
\includegraphics[width=1\textwidth]{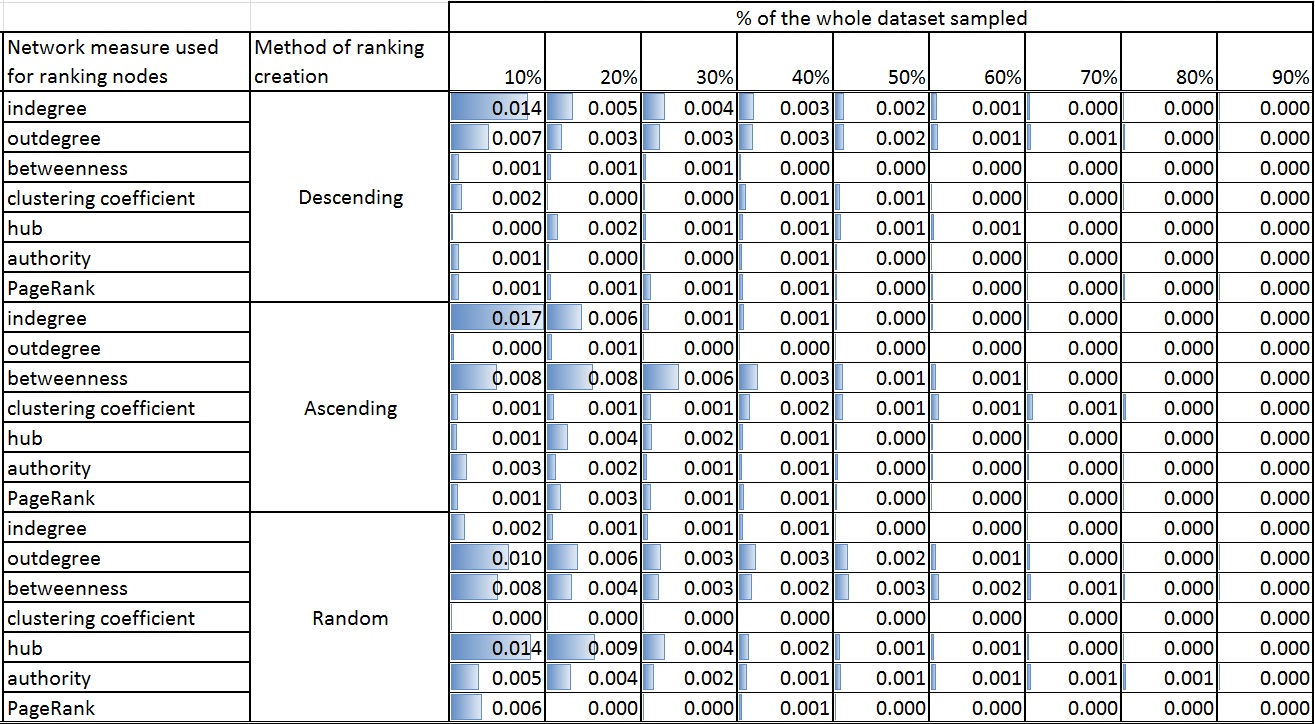} \caption{\small \sl Kullback--Leibler divergence for Pairs FSG network.
\label{fig:kl_div_pairs}}  
\end{center}  
\end{figure} 

\begin{figure}[ht]  
\begin{center}  
\includegraphics[width=1\textwidth]{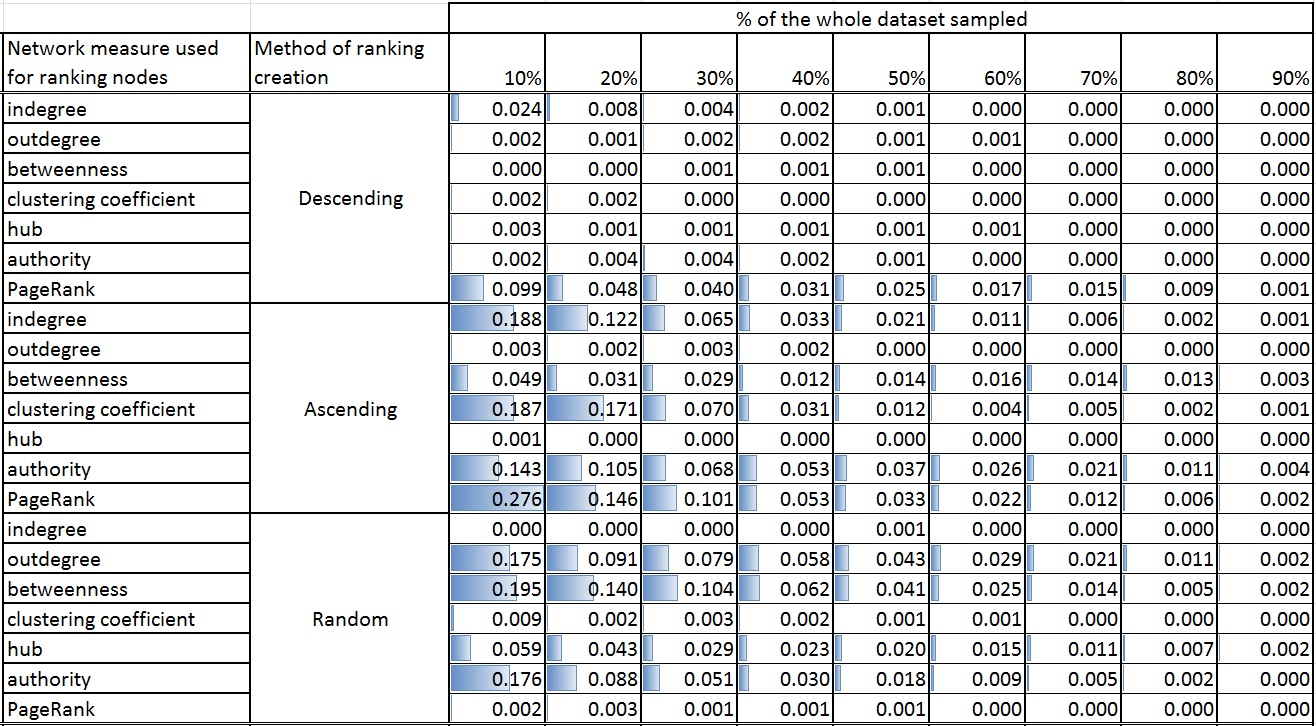} \caption{\small \sl Kullback--Leibler divergence for Pairs small FSG network.
\label{fig:kl_div_pairs_small}}  
\end{center}  
\end{figure} 

\begin{figure}[ht]  
\begin{center}  
\includegraphics[width=1\textwidth]{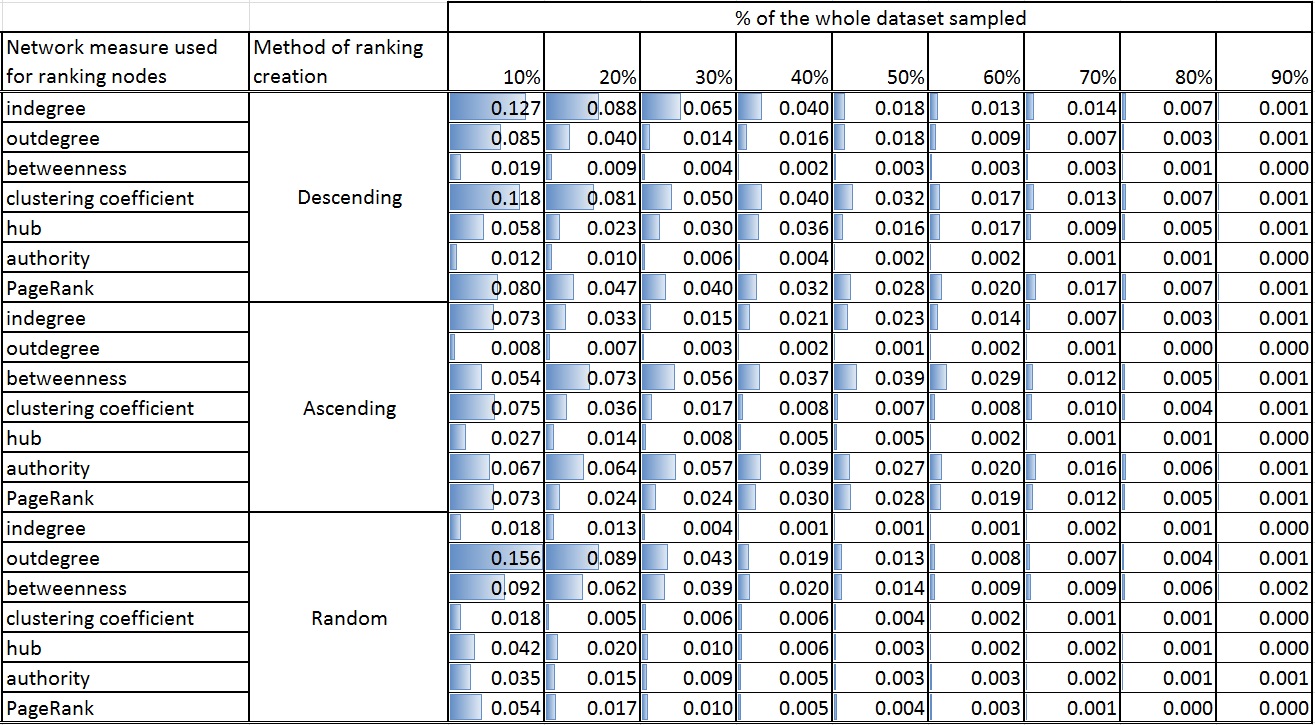} 
\caption{\small \sl Kullback--Leibler divergence for yeast network.
\label{fig:kl_div_yeast}}  
\end{center}  
\end{figure} 
\clearpage

\end{document}